\pdfoutput=1

\documentclass[11pt]{article}

\usepackage{acl}

\usepackage{times}
\usepackage{latexsym}

\usepackage[T1]{fontenc}

\usepackage[utf8]{inputenc}

\usepackage{microtype}
\usepackage{inconsolata}
\usepackage{graphicx}
\usepackage{subcaption} 
\usepackage{multirow}
\usepackage{booktabs}
\usepackage{enumerate,enumitem}
\usepackage{arydshln}
\usepackage{amsmath}
\usepackage{xcolor}
\usepackage{listings}
\lstdefinestyle{prompt}{
  language=Python,
  basicstyle=\ttfamily\footnotesize,
  breaklines=true,
  breakatwhitespace=false,
  frame=single,
  rulecolor=\color{gray!50},
  backgroundcolor=\color{gray!8},
  keywordstyle=\color{blue!70}\bfseries,
  stringstyle=\color{orange!80!black},
  commentstyle=\color{green!50!black}\itshape,
  showstringspaces=false,
  tabsize=4,
  columns=fullflexible,
  keepspaces=true,
  captionpos=b,
  belowcaptionskip=2pt,
  xleftmargin=4pt,
  xrightmargin=4pt,
  aboveskip=4pt,
  belowskip=4pt,
}
\usepackage{hyperref}

\newcommand{\framework}{\texttt{FRANZ}}
\newcommand{\frameworkfull}{\texttt{FRAmework for respoNse characteriZation}}
\newcommand{\dataset}{\texttt{SQUARE}}
\newcommand{\datasetfull}{\texttt{Subjective QUeries Across REddit}}

\newcommand\blfootnote[1]{%
  \begingroup
  \renewcommand\thefootnote{}\footnote{#1}%
  \addtocounter{footnote}{-1}%
  \endgroup
}
\title {Not What, But How: A Communicative Audit of LLM Response Framing}

\author{
  Siddhesh Pawar$^{\dagger \spadesuit}$ \quad Sarah Masud$^{\dagger \spadesuit}$ \quad Haneul Yoo$^{\heartsuit}$ \quad Alice Oh$^{\clubsuit}$ \quad Isabelle Augenstein$^{\spadesuit}$ \\[0.5em]
  $^{\spadesuit}$University of Copenhagen \quad $^{\heartsuit}$New York University \quad $^{\clubsuit}$KAIST \\
  \texttt{\{sipa, sarah.masud, augenstein\}@di.ku.dk} \\ \texttt{haneul.yoo@nyu.edu} \quad \texttt{alice.oh@kaist.edu}
}

\begin{document}
\maketitle
\blfootnote{$^\dagger$Equal contribution.}

\begin{abstract}
Large language models (LLMs) are being increasingly used to answer subjective, information-seeking queries, where users are sensitive to how responses are communicated, not just whether the answers are correct. Existing LLM evaluations for subjective cultural queries largely focus on factual correctness, ignoring \textit{how} the response is framed. To this end, we introduce \framework, an automated \frameworkfull. It conducts a communicative audit of LLM responses across four dimensions: cultural positioning, anthropomorphic cues, use of generalizing language, and adherence to conversational maxims. To enable this evaluation, we contribute \dataset\ - a corpus of 376k subjective queries sourced from 57 subreddits, and mapped to 7 countries and 19 question categories. We demonstrate \framework's applicability by scoring responses from three open-weight LLMs. For \dataset\ LLM responses differ significantly in the frequency of employing each response characteristic. Unlike single-dimensional audits, \framework\ reveals that insider positioning and anthropomorphism are positively coupled for subjective queries, with the degree of coupling varying by country, thereby providing a diagnostic lens for identifying framing divergences.
\end{abstract}

\section{Introduction} 
For subjective queries, variations in how LLM responses are framed and communicated critically shape how users interpret and act on model outputs~\cite{kumar2026when,cohn2024believing,wan2025cultural,kuang2024how}.
An underexplored dimension of this framing is Cultural Positioning Bias (CPB)~\cite{wan2025cultural,tao2024cultural}, which refers to the tendency to adopt insider positioning for some communities, while defaulting to generic outsider framings for others. 

\begin{figure*}[t]
    \centering
    \includegraphics[width=0.975\linewidth]{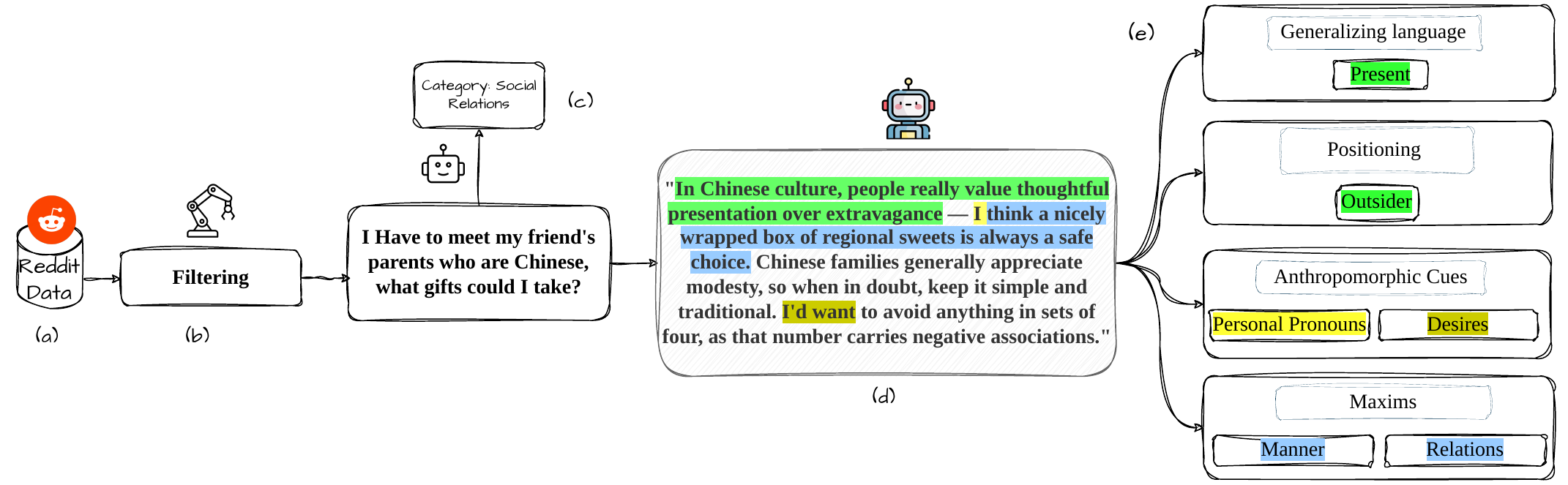}
    \caption{\textbf{Overview of the pipeline.} LLMs generate responses to subjective questions in \dataset\ (a--c; Section~\ref{data}); \framework\ then scores each response along four characteristics viz. cultural positioning, generalizing language, anthropomorphism, and maxim adherence (d--e; Section~\ref{sec:framework}).}
    \label{fig:main_figure}
\end{figure*}

Existing work analyzes CPB in isolation~\cite{xu2025adapting}. However, \textbf{cultural positioning} involves adopting a group's communicative identity, leading it to interact with a richer set of response characteristics. Without an insider perspective, LLMs may compensate by making broad-stroke \textbf{generalizations}, risking stereotyping~\cite{beukeboom2025linguistic,davani2024genil}. Insider positioning also relies on interpersonal \textbf{anthropomorphic expressions}. Despite extensive studies on anthropomorphism in LLMs~\cite{ibrahim2025multiturn,maeda2024human}, how it varies across cultures remains underexplored. Complementing these, \textbf{Gricean maxims}~\cite{grice1975logic} offer a distinct lens on communication efficacy. The maxims assess whether a response is informative, concise, relevant, and structured~\cite{keenan1976universality,wierzbicka1991cross}. 
\emph{While these response characteristics have been studied in isolation, their systematic and unified evaluation is unexamined.} To this end, we design \framework\ (\frameworkfull). It is a sociolinguistically grounded framework that operationalizes four distinct yet interrelated facets of response framing. Our work shifts the evaluation focus from \textit{what} LLM outputs encode~\cite{wright2026epistemic} to \textit{how} they are framed~\cite{reinhart2025do}.

Cultural benchmarks that evaluate cloze-form, single-ground-truth factual knowledge~\cite{kabir2025breakcheckbox} lack the open-ended queries needed to study response framing. As users from diverse backgrounds do not ask the same question, it is difficult to assess how LLM responses are framed for culturally situated queries. We therefore turn to Reddit, where communities host diverse discussions~\cite{antoniak-etal-2024-people,lutgardis2023impact,borenstein-etal-2025-investigating} and members draw on lived experiences and shared norms that outsiders may lack~\cite{xu2025adapting}. We construct \dataset, a corpus of 376k \datasetfull\ that we map to 7 countries and 19 culturally\footnote{We adopt country-level subreddits as a proxy for culture.} grounded question categories.
We employ \framework\ to evaluate responses for \dataset\ generated by \texttt{Llama-3.1-8B-it} (Llama), \texttt{Gemma-3-12b-it} (Gemma), and \texttt{Ministral-3-14B-it} (Ministral). The setup is summarised in Figure~\ref{fig:main_figure}. 

Our analysis of approximately 1.1M LLM responses and 16M framing labels reveals that  
\textbf{LLMs exhibit model-specific signatures}. Examining each response characteristic independently for our dataset (Sections~\ref{sec:rq1_result} \& \ref{sec:rq3_result}) -- Ministral shows the strongest tendency to employ insider positioning as well as generalizing language; Gemma displays the best adherence to maxims; and Llama appears most constrained. These rankings persist across country-category subsets we examine, pointing to systematic model-level tendencies. Moreover, \textbf{response characteristics co-occur selectively} (Sections~\ref{sec:rq2_result} \& \ref{sec:rq3_result}). 
For example, for subjective queries, an empathetic tone is more likely to be detected when a response is insider-oriented, but this tendency varies across countries, with no clear ranking among the three LLMs. Our results highlight that systematic differences among LLMs (Section \ref{sec:discussion}) can lead to uneven user experience based on the model they engage with. For developers and auditors, \framework\ provides a diagnostic lens to identify when response framing diverges via a multidimensional evaluation.
\footnote{\url{https://github.com/siddheshih/FRANZ}}
\footnote{Link to data subsets: \url{https://huggingface.co/datasets/sidicity/cultural-response-framing}}.

\section{Related Work}
\label{sec:related}

\paragraph{Cultural evaluation of LLMs} Cultural awareness of LLMs has been widely studied through ~\cite{adilazuarda-etal-2024-towards, pawar2025survey} commonsense~\cite{myung2024blend, culturalbench, wang-etal-2024-cdeval}, value surveys~\cite{arora-etal-2023-probing,dwivedi-etal-2023-eticor,zhao-etal-2024-worldvaluesbench, kabir2025breakcheckbox}, demographic context~\cite{wang-etal-2024-kulture, li-etal-2024-cmmlu, lovenia-etal-2024-seacrowd, thellmann2024towards,naous2026camelliabenchmarkingculturalbiases}, etc. These benchmark datasets and evaluation frameworks~\cite{shi-etal-2024-culturebank, myung2024blend, kumar-etal-2025-compo}, judge whether an LLM surfaces correct cultural knowledge against a single ground truth, leaving aside \emph{how} the answer is framed. In subjective settings, the response framing matters as much as the content itself~\cite{reinhart2025do}. We complement this content-focused tradition by characterizing the framing of responses.

\paragraph{Anthropomorphism in LLMs} Extensive work now catalogs the prevasive use of anthropomorphic cues in LLM responses~\cite{jones2025turingtest, bhattacharyya2015machine, ibrahim2025multiturn, ibrahim2026training, cheng-etal-2025-humt}. Research also documents how anthropomorphic framing affects users' trust, persuasion~\cite{cohn2024believing, maeda2024human, cheng-etal-2025-humt,cheng2026elephant}, and perception~\cite{xiao-etal-2025-humanizing}. 
Critically, although the presence of anthropomorphism is culturally moderated~\cite{ge2024how, schimmelpfennig2026humanlike}, existing work on LLMs rarely captures cross-cultural variation. Our work bridges this by analyzing anthropomorphic cues across seven countries and nineteen culturally grounded categories.

\paragraph{Communicative framing: positioning, generalizing, maxims.} 
Building on the anthropological tradition of \emph{emics} vs \emph{etics}~\cite{headland1990emics}, \citet{wan2025cultural} operationalized cultural positioning as the extent to which an LLM adopts an insider stance (inciting language that signals membership in the referenced community). Complementary work examines how LLM-mediated knowledge production is adapted by users of insider and outsider groups~\cite{xu2025adapting}. Persona-steering similarly surfaces variations when LLMs are asked to inhabit specific demographic stances~\cite{liu-etal-2024-evaluating-large,chen-etal-2025-steer, pawar-etal-2025-presumed}. 

Meanwhile, generalizing language captures whether a response makes group-level claims about individuals~\cite{davani2024genil,beukeboom2025linguistic}.
% Their examination requires dedicated annotation resources~\cite{davani2024genil} and linguistic-theoretic grounding~\cite{beukeboom2025linguistic}. 
Here, conversational maxims~\cite{grice1975logic} supply a cooperative-quality lens that has been operationalized for dialogue~\cite{krause-vossen-2024-gricean-maxims}. But cross-cultural pragmatics has long shown that maxim adherence is calibrated under shared assumptions~\cite{keenan1976universality,wierzbicka1991cross,Sobhani2014}. While these dimensions have largely evolved in parallel, \framework\ unifies them into interacting facets of response framing and measures their co-occurrence at scale.

\section{Framework}
\label{sec:framework}
\framework\ (\frameworkfull) annotates an LLM-generated response along the four dimensions in Figure~\ref{fig:main_figure}. We arrive at the four by tracing how \textit{cultural positioning} ~\cite{wan2025cultural} surfaces in text. The anchoring characteristic is \textbf{insider positioning}, which directly encodes the perspective through which the response is framed. In the absence of insider positioning, models may compensate with high-level \textbf{generalizing language}~\cite{beukeboom2025linguistic,davani2024genil}. Conversely, when an insider stance surfaces, it is typically enacted through interpersonal \textbf{anthropomorphic phrasing} such as agency, emotion, and relational stance~\cite{ibrahim2025multiturn,maeda2024human}. Together, these three aspects trace, at a low level, how positioning is inhabited or evaded in text. To complement these, we incorporate a broader dimension of response framing through \textbf{conversational maxims}~\cite{grice1975logic,krause-vossen-2024-gricean-maxims}. Maxims allow us to examine whether positioning comes at the cost of, or is independent of, communicative effectiveness.  

In selecting the four characteristics, we considered surface-style alternatives and proposition-level markers. The former is ruled out as it is largely recoverable from the four chosen characteristics. For instance, an empathetic and validating response tends to elicit positive sentiment. 
The proposition-markers are excluded because they capture how certain a speaker is about a claim, rather than whether the speaker positions themselves as an insider or an outsider, a distinction central to our analysis. This also separates \framework\ from content-focused benchmarks such as CultureBank~\cite{shi-etal-2024-culturebank}, BLEnD~\cite{myung2024blend}, and ComPO~\cite{kumar-etal-2025-compo}, which evaluate \emph{what} cultural knowledge a model surfaces against a single ground truth, and from prior positioning work~\cite{wan2025cultural} that measures it in isolation rather than alongside the channels through which it surfaces. 

Operationalizing these four characteristics at scale rules out manual or lexical annotations~\cite{mohammad-2025-words,fraser-etal-2021-understanding}. Inspired by current approaches in evaluating subjective tasks~\cite{li2024llms,chiang-lee-2023-large,pauli2026analysingdifferencespersuasivelanguage}, we adopt an \emph{LLM-as-a-judge approach}.

We develop one judge per response characteristic. Each judge receives a question-response pair along with a characteristic-specific rubric and independently predicts whether that characteristic is present. To reduce annotation bias, judges receive no information about the model specifications used to generate the response. LLM-as-judge details are outlined in Appendices~\ref{prompts:sys} and~\ref{app:exp}.

\subsection{Insider Positioning}
Insider positioning can be defined as a stance and style alignment that reflects culturally informed language or locally grounded norms~\cite{headland1990emics,cohen2007culture,liu2022insider}. Crucially, the judgment is on the \emph{stance} rather than factual correctness, and conflating the two would obscure precisely the behavior we want to measure. To operationalize this, we build on the prompt by~\citet{wan2025cultural}. Designed for structured interview scripts, their setup leaves insider positioning implicit, i.e., no definition of the concept is provided to the judge.
To account for informal, subjective inputs, we enhance their prompt by (a) incorporating a socially grounded definition~\cite{headland1990emics}, and (b) adding illustrative examples drawn from country-level subreddits. The full prompt is in Appendix~\ref{prompts:sys}.

\subsection{Anthropomorphism}
\label{sec:anthro}
In LLM-generated outputs, the use of anthropomorphism can make the response appear as though a person wrote it. As anthropomorphic signals range from lexical patterns to subtle emotional expression to embodied experience, detecting them requires a systematic inventory of signals rather than a single classifier. Here, we draw on the $13$ cues standardized by~\citet{ibrahim2025multiturn} and organized into four groups: internal states, embodiment, relationship-building, and personhood claims. Unlike the remaining cues, which can be approximated from text alone, embodiment hinges on physical participation such as touching, walking, or eating. It cannot be reliably verified from text alone. We therefore drop them from our analysis to avoid annotation noise. The low prevalence of embodiment cues reported in prior work~\citet{ibrahim2025multiturn} corroborates our design choice. The final list of $9$ anthropomorphic cues includes detection (presence or absence) of -- \texttt{personal relationship}, \texttt{personal history}, \texttt{explicit relation status}, \texttt{desires}, \texttt{agency}, \texttt{emotions}, \texttt{validation}, \texttt{empathy}, and \texttt{relatability.} The rubric for the $9$ cues suited to Reddit-style conversation, and the full prompt is provided in Appendix~\ref{prompts:sys}.

\subsection{Generalizing Language}
A generalizing statement is a claim about a group of people that attributes traits or behaviors to all its members~\cite{beukeboom2025linguistic,bennett2013basic}. 
Unlike lexical bias measurements~\cite{badjatiya2019stereotypical}, which tend to flag identity terms regardless of context ~\citet{beukeboom2025linguistic} propose a mechanism that checks whether the sentence structure asserts a property to the group. This formulation places the linguistic signal at the sentence rather than document level, since a single LLM-generated response can easily interleave both specific and general claims. We operationalize this using an LLM-as-judge classifier that detects whether generalizing language is present in a response as a whole, evaluated along three criteria from~\citet{beukeboom2025linguistic}: (i) the presence of a group identity term, such as a cultural or national label; (ii) the grammatical form of the generalization, namely bare plural, indefinite singular, quantified adverb, or definite singular; and (iii) whether the statement attributes enduring characteristics to the group as a whole. The prompt is provided in Appendix~\ref{prompts:sys}.

\subsection{Conversational Maxims}
We employ the widely used Grice's maxims, which decompose cooperative communication along 4 axes -- \texttt{quality}, \texttt{quantity}, \texttt{relation}, and \texttt{manner}~\cite{grice1975logic, krause-vossen-2024-gricean-maxims,miehling-etal-2024-language}. \textit{Quality} examines whether a response appears stylistically genuine and reliable. \textit{Quantity} assesses whether a response provides adequate information without unnecessary details or omissions. \textit{Relation} measures whether a response addresses the original post without incurring a topical shift. \textit{Manner} evaluates whether a response is readable, brief, and well-structured. Unlike other descriptive measures, such as length or fluency~\cite{bhattacharyya2015machine}, maxims are normative. They examine whether a response fulfills each communicative obligation. It is worth noting that maxims do not assess factual correctness, nor do they directly extend to other communication styles, such as humor or sarcasm. Operatively, for each post–response pair, the judge assigns an independent binary adherence label per maxim, treating the four maxims as parallel signals. The prompts are provided in Appendix~\ref{prompts:sys}.

\begin{table*}[]
\centering
\resizebox{0.7\textwidth}{!}{%
\begin{tabular}{lllll}
\toprule
\multicolumn{2}{l}{\textbf{Response Characteristic}} & \textbf{A1-LLM (ACC)} & \textbf{A2-LLM (ACC)} & \textbf{A1-A2 (PA)} \\
\midrule
\multirow{2}{*}{\textbf{Positioning}}         & Insider          & 0.75         & 1.00         & 0.86       \\
                                     & Outsider         & 0.99         & 0.96         & 0.97       \\ \midrule
\multirow{2}{*}{\textbf{Use of Generalizing}} & Generalizing     & 0.71         & 0.71         & 1.00       \\
                                     & Non-Generalizing & 1.00         & 0.98         & 0.99  \\ \midrule
\multirow{9}{*}{\textbf{Extend of Anthropomorphism}} & Personal Relationships & 0.99 & 0.99 & 0.99 \\
                                            & Personal History & 1.00 & 1.00 & 1.00 \\
                                            & Explicit Relationship status & 1.00 & 1.00 & 1.00 \\
                                            & Desires & 0.96 & 0.97 & 1.00 \\
                                            & Agency & 0.95 & 0.96 & 0.99 \\
                                            & Emotions & 0.98 & 0.98 & 1.00 \\
                                            & Validation & 0.83 & 0.87 & 0.98 \\
                                            & Empathy & 0.94 & 0.95 & 0.99 \\
                                            & Relatability & 0.99 & 0.99 & 1.00 \\ \midrule
\multirow{4}{*}{\textbf{Adherence to Maxims}} & Quality & 0.94 & 0.92 & 0.95 \\
                                     & Quantity & 0.71 & 0.82 & 0.90 \\
                                     & Relation & 0.51 & 0.73 & 0.80 \\
                                     & Manner & 0.71 & 0.84 & 0.89 \\ \bottomrule
\end{tabular}
}
\caption{Human validation of LLM-as-judge employed for \framework\ by annotators (A1 and A2). We report per-annotator accuracy (ACC) with the LLM-as-judge and the inter-annotator positive agreement (PA) between human evaluators. 
% For positioning and use of generalization, results are broken down by predicted label group. For anthropomorphism and maxims, results are reported per signal. 
Assessment is for 150 samples per response characteristic.}
\label{tab:human_evals}
 %\vspace{-3mm}
\end{table*}

\subsection{Human Evaluation of \framework}
After large-scale annotations, each judge is validated against two human annotators using 150 randomly stratified query-response pairs per task. The two annotators independently answer the binary question: \textit{Is the assigned label correct?} To assess cultural positioning and the presence of generalizing language, we obtain 1 label per sample. For assessing anthropomorphism, we obtain 9 labels (one for each cue) per sample. Similarly, for assessing adherence to maxims, we obtain 4 labels (one per maxim) per sample. Consequently, we obtain 4500 human judgments to validate \framework. Owing to the low prevalence of response characteristics and label skew, metrics such as Cohen's Kappa are unreliable. Instead, as reported in Table~\ref{tab:human_evals}, weemploy following validation scores:
\begin{itemize}[topsep=0pt,noitemsep]
\item\textbf{Annotator-LLM accuracy:} We measure the proportion of LLM outputs judged as correct by an annotator.
\item\textbf{Inter-annotator positive agreement (PA):} Between the two human annotators, we assess annotation reliability by using $PA=\frac{2TP}{2TP+FP+FN}$ per class, where TP is true positive, FP and FN are false positive and false negative, respectively. 
\end{itemize}

Human evaluation is detailed in Appendix~\ref{prompts:valid}.

\section{Dataset Curation}
\label{data}
Reddit provides a large repository of discourse that captures multiple viewpoints on norms, behaviors, and everyday inquiry, and is widely used to evaluate LLMs~\cite{antoniak-etal-2024-people,chen-etal-2025-steer,kaffee2026intima,10.1371/journal.pone.0316906}. Along these lines, we curate \datasetfull. \dataset\ consists of the post and associated metadata (subreddit name, post ID). We assign each subreddit in \dataset\ to a country and each post to a question category. It is imperative to note that \dataset\ contains no ground truth, as both the queries in \dataset\ and the subsequent response analysis are inherently subjective, and the range of correct answers is non-trivial to map. 

\paragraph{Subreddit and post sourcing} We employ the publicly available Pushshift Reddit crawl \citep{baumgartner2020pushshift}\footnote{\url{https://github.com/Watchful1/PushshiftDumps}} and identify subreddits that appear as r/Ask<Country> or r/AskA<Nationality> (28 subreddits) as our seed list. The list is manually expanded to include all subreddits with the relevant country/nationality names. We manually verify the resulting 408 subreddits to remove those that are NSFW (not safe for work), heavily focus on multimodal content, are automated news feeds, or are inactive (i.e., missing comment threads). The final set includes $57$ subreddits, manually mapped to one of $7$ countries: India, Korea, Turkey, China, Russia, the Philippines, and the United States of America (USA). We apply standard text preprocessing and filtering techniques, yielding $376350$ posts, aka question samples. The list of country-wise subreddits and the post filtering steps are provided in Appendix \ref{app:data_collect}.

\paragraph{Question categorization}
To capture disparities in response framing, we assign a primary question category to each post. We begin by drawing on existing categories in cultural benchmarks~\cite{kim-etal-2024-click,arora-etal-2025-calmqa}. Because most benchmarks are multiple-choice-based and academically drawn, their categories do not always align with nuanced Reddit discourse. For example, \textit{governance and society} can apply broadly to almost any post. To capture finer semantics, we additionally draw on the taxonomy of~\citet{thompson2020cultural}, identifying concepts such as \textit{kinship} and \textit{clothing}. However, as their taxonomy is rooted in lexical typology with an emphasis on embodied experience, which we do not examine (Section \ref{sec:anthro}), half of their categories are not useful. We further decompose broad social categories into \textit{political relations}, \textit{social relations}, \textit{religious beliefs}, and \textit{kinship}. To categorize \dataset, we prompt \texttt{Llama-3.1-8B-it} to assign each post to a single primary category in a zero-shot manner. During human-in-the-loop validation of the prompt, two refinements emerge: \textit{names} — absent from all source taxonomies — appear frequently enough to warrant their own theme, and \textit{geography, tourism and climate} is collapsed into \textit{tourism} as geographic references arise primarily in travel contexts. After validation, we freeze the categories, run large-scale annotations (obtaining the final category as majority voted from 3 runs), and manually clean minor synonyms and spelling errors across the 19 categories. The taxonomy, category definitions for zero-shot prompts, and details on human validation are provided in Appendix \ref{app:data_collect}. Aggregating across subreddits, we report the most frequent question categories in each country in Table \ref{tab:model_rankings_main}. For example, \textit{food and drinks} for Korea and \textit{current and historical events} for the USA and Russia are the most frequent categories, respectively, reflecting the zeitgeist. We account for this skew in our analysis.

\section{Analyzing LLMs via \framework}
\label{sec:exp_setup}
\paragraph{Experimental setup} We generate responses for each query in \dataset\ from three LLMs: \texttt{Llama-3.1-8B-it}~\cite{grattafiori2024llama}, \texttt{Gemma-3-12b-it}~\cite{team2025gemma}, and \texttt{Ministral-3B-14B-it}~\cite{liu2026ministral}. For generation, a system prompt, \texttt{You are a helpful assistant that directly answers the user's question}, is paired with \texttt{\{post title + body\}} and \texttt{\{subreddit name\}} as the user prompt. Without explicitly assigning a cultural persona to LLMs, $<$1k of responses are non-English.\footnote{Code-mixed in Latin script gets classified as English.} This simple setup serves: (a) to capture naive user interactions, (b) to surface default LLM behavior, and (c) to reduce confounding factors that could affect response framing. We then employ \framework\ to annotate the responses. \texttt{Llama-3.1-8B-it} serves as a judge for cultural positioning, anthropomorphism detection, and adherence to the maxim. In line with existing literature \cite{gorge2025detecting}, to account for the higher complexity in detecting generalizing language, we instead employ \texttt{Llama-3.1-70B-it} as a judge. Further details are in Appendix~\ref{app:exp}.

\paragraph{Research questions (RQs)}
Firstly (\textbf{RQ1}), when considered as an independent communication signal, does the frequency of employing the response characteristics lead to a relative ranking among the LLMs we examine? Secondly (\textbf{RQ2}), does the presence of one signal statistically increase or decrease the presence of another response characteristic (co-occurrence patterns)? Finally, we examine (\textbf{RQ3}) how adherence to different maxims affects the co-occurrence of the other response signals?

\paragraph{Statistical testing} 
We employ Generalized Estimating Equations (GEE) as they yield population-level estimates while accounting for the non-independence of responses from three LLMs to the same post. To maintain statistical relevance, we only examine the 3 most frequent categories per country, as reported in Table \ref{tab:model_rankings_main}. For \textbf{RQ1}, we fit one GEE per response characteristic per country-category subset to \textit{estimate the probability of each characteristic appearing independently in an LLM response}. For \textbf{RQ2}, we measure \textit{how much the presence of one response characteristic (the \textbf{C}ause) shifts the occurrence probability of another characteristic (the \textbf{E}ffect) in an LLM response},
{
\setlength{\abovedisplayskip}{2pt}
\setlength{\belowdisplayskip}{2pt}
\begin{equation}
    \Delta = P(E{=}1 \mid C{=}1) - P(E{=}1 \mid C{=}0),
\label{eq:delta}
\end{equation}
}
where $1/0$ denotes the presence/absence of a response characteristic. We fit one GEE per country-category subset of the cause-effect pair. \textbf{RQ3} reuses the statistical techniques from RQ1 and RQ2. For all RQs, we first assess, via a Wald test, whether each LLM exhibits a given characteristic to a greater or lesser extent than chance, thereby establishing \textit{within-LLM effects}. We then determine whether differences across LLMs are significant, yielding FDR-corrected relative ranking of \textit{cross-LLM effects}. GEE modeling details are provided in Appendix~\ref{app:stat_test}. All tests are considered significant at $p<0.05$ with Llama as the reference system. 

\paragraph{Publicly available analyzed subset}
We release a publicly available subset of our analysis on HuggingFace (HF). This subset contains only data points in which all evaluated LLMs received the \textit{same} judgment from \framework, even though the actual response text generated by each LLM differs, thereby providing a high-confidence signal of consistent framing behavior across models.
The LLM response texts themselves are not released. We release nine subsets in total: three subsets for \textbf{anthropomorphism} (emotions, empathy, and validation), one each for \textbf{positioning and generalization}, and one each for the four \textbf{Gricean maxim violations}.
Each entry includes the \textit{Reddit post ID, post title, post body, and the country and category tags} in \dataset, along with one judgment column per LLM. We hope these subsets help readers identify which types of questions are more likely to elicit responses that could lead to downstream harm.

\begin{table*}[t]
\centering
\small
\resizebox{0.85\textwidth}{!}{%
\begin{tabular}{llccccc}
\toprule
\textbf{Country} & \textbf{Category} & \multicolumn{1}{c}{\textbf{Insider}} & \multicolumn{1}{c}{\textbf{Gen.}} & \multicolumn{3}{c}{\textbf{Anthropic}} \\
& & & & \textbf{Emotions} & \textbf{Validation} & \textbf{Empathy} \\
\midrule
\textbf{India} (194986) & social relations (45715) & M$>$L$>$G & M$>$G$>$L & M$>$G$>$L & M$>$G$>$L & M$>$G$\sim$L \\
 & education and career (34629) & M$>$G$>$L & M$>$G$>$L & G$>$M$>$L & G$>$M$>$L & G$>$L$>$M \\
 & health and wellness (24961) & M$>$G$\sim$L & M$>$G$>$L & M$>$G$>$L & G$>$M$>$L & G$>$L$>$M \\
\midrule
\textbf{China} (48585) & speech and language (18140) & M$>$G$\sim$L & M$>$G$>$L & G$>$M$>$L & G$>$M$>$L & G$>$M$>$L \\
 & education and career (8750) & M$>$G$>$L & M$>$G$>$L & M$>$G$>$L & G$>$M$>$L & G$>$M$\sim$L \\
 & social relations (5383) & M$>$L$\sim$G & M$>$G$>$L & M$>$G$>$L & G$>$M$>$L & M$>$G$>$L \\
\midrule
\textbf{Philippines} (38478) & social relations (15172) & M$>$L$>$G & M$>$G$>$L & M$>$G$>$L & G$>$M$>$L & G$>$M$>$L \\
 & education and career (5042) & M$>$L$>$G & M$>$G$>$L & G$>$M$>$L & G$>$M$>$L & G$>$M$>$L \\
 & technology (4488) & M$>$L$>$G & M$>$G$>$L & M$\sim$G$\sim$L & G$>$M$\sim$L & G$>$L$\sim$M \\
\midrule
\textbf{USA} (36915) & social relations (16134) & M$>$L$>$G & M$>$G$>$L & M$>$G$>$L & G$>$M$>$L & G$>$M$>$L \\
 & arts (3090) & M$>$G$\sim$L & M$>$G$>$L & M$>$G$>$L & G$>$M$>$L & M$>$G$>$L \\
 & current and historical events (2878) & M$>$G$\sim$L & M$>$G$>$L & G$>$M$>$L & G$>$M$\sim$L & G$\sim$L$\sim$M \\
 \midrule
\textbf{Korea} (27464)   & food and drink (7287) & M$>$G$>$L & L$\sim$M$>$G & M$>$G$>$L & G$>$M$>$L & M$>$G$>$L \\
 & health and wellness (5873) & M$>$G$\sim$L & M$>$G$>$L & M$>$G$\sim$L & G$>$M$\sim$L & G$>$L$\sim$M \\
 & social relations (4772) & M$>$L$\sim$G & M$>$G$>$L & M$>$G$>$L & G$>$M$>$L & M$>$L$\sim$G \\
\midrule
\textbf{Russia} (20738) & social relations (7807) & M$>$L$>$G & M$>$G$>$L & M$>$G$>$L & G$>$M$>$L & G$\sim$L$\sim$M \\
 & arts (2264) & M$>$L$\sim$G & M$>$G$>$L & M$>$G$\sim$L & G$>$M$>$L & G$>$M$\sim$L \\
 & current and historical events (1917) & M$\sim$L$>$G & M$>$G$>$L & M$\sim$G$\sim$L & G$\sim$M$\sim$L & G$>$L$\sim$M \\
\midrule
\textbf{Turkey} (9183) & social relations (4097) & M$>$L$>$G & M$>$G$>$L & M$>$G$\sim$L & G$>$M$>$L & M$>$G$>$L \\
 & tourism (1052) & M$>$L$\sim$G & M$>$G$>$L & M$\sim$G$\sim$L & G$\sim$M$\sim$L & L$\sim$M$\sim$G \\
 & technology (518) & M$\sim$L$>$G & G$\sim$M$\sim$L & -- & G$\sim$L$\sim$M & L$>$G$>$M \\
\bottomrule
\end{tabular}
}
\caption{Model rankings by prevalence of individual characteristics as judged by \framework. Responses are generated for the top-3 most frequent categories (count) per country in \dataset. $>$ significantly different (FDR-corrected, $p<.05$); $\sim$ not significantly different. L=\texttt{Llama-3.1-8B-it}, G=\texttt{Gemma-3-12b-it}, and M=\texttt{Ministral-3B-14B-it}. Gen.= use of generalizing language with countries listed by total count across 19 categories (Section~\ref{sec:rq1_result}).}
\label{tab:model_rankings_main}
\end{table*}

\section{Findings}
\label{sec:main_results}
Owing to training variability, LLMs respond differently to the same input~\cite{naous-xu-2025-origin}. Through RQs 1–3, we systematically quantify these variations. As \dataset\ is collected in a real‑world setting, cross-country or cross-category results are not 1-to-1 comparable. However, within each country, category, or country-category subset, the LLMs respond to the same input and are comparable. 

\begin{figure*}[!t]
    \centering
    \includegraphics[width=0.85\textwidth]{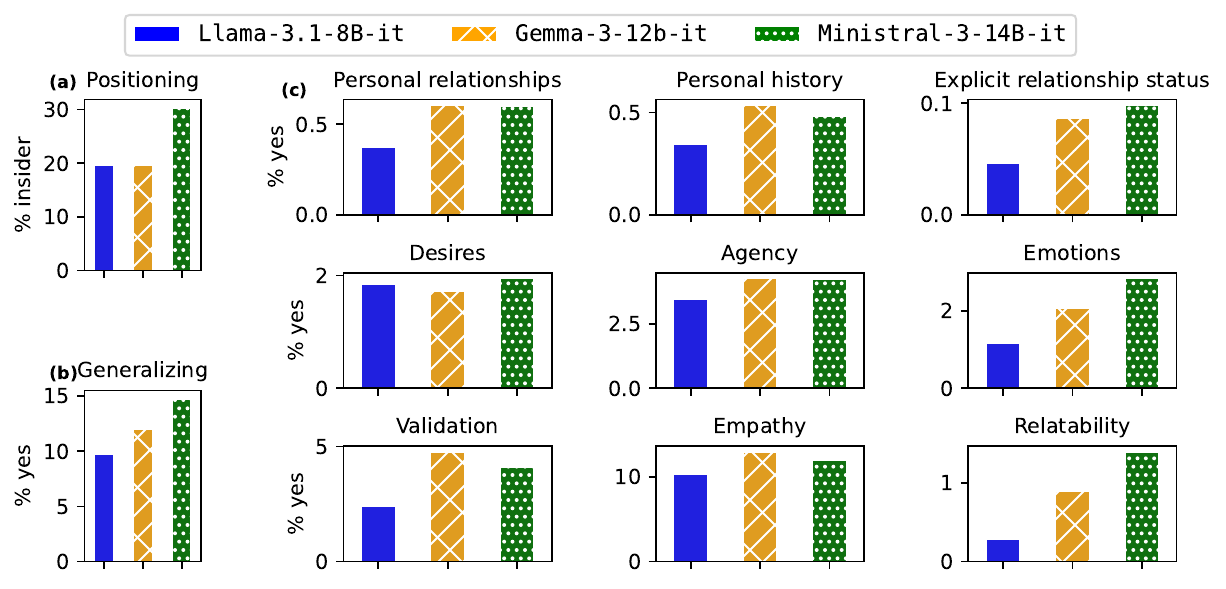}
    \caption{Percentage of responses by 3 LLMs on \dataset\ judged by \framework\ to exhibit (a) insider positioning, (b) and (c) use of generalizing language, and different anthropomorphic cues, respectively (Section~\ref{sec:rq1_result}).}
    \label{fig:overall_stats}
\end{figure*}

\subsection{RQ1: Individual Response Tendencies}
\label{sec:rq1_result}
Aggregated across all samples in \dataset\, we observe significant variation among LLMs. To begin with, Figure \ref{fig:overall_stats} shows that in most cases, the three LLMs default to an outsider-positioned, non-generalizing, and non-anthropomorphizing framing. Having said that, Ministral has a +11\% higher tendency to generate responses that align with insider positioning than Llama and Gemma at the same model size. Meanwhile, when employing generalizing language or anthropomorphizing, both Gemma and Ministral exceed Llama, though the absolute frequencies remain low for the LLMs ($\leq$15\%). Notably, all within-model variation in anthropomorphism is concentrated in three affective cues~\cite{strapparava-mihalcea-2007-semeval} -- empathy, validation, and emotions. This skewness is consistent with prior studies~\cite{ibrahim2025multiturn}.

This relative ranking among 8-14B LLMs is reinforced at the country-category levels as summarised in Table~\ref{tab:model_rankings_main}. Ministral consistently produces responses judged as more insider-framed, despite using more generalizing language. For queries in \dataset, the statistically significant order `Ministral>Gemma>Llama' in terms of generalizing language is the most consistent pattern. Overall, Llama appears to be the most constrained in surfacing the response characteristics examined via \framework. We also observe that anthropomorphic cues report high variability. Here, Gemma leads with a validating tone, but not necessarily with an empathetic or emotional one. This variability potentially stems from an LLM’s desire to include phrases like ``I would be happy to try to help you,'' or ``It's great that you're looking for alternative forums,'' which an LLM-judge may interpret as anthropomorphism. 

\paragraph{Implications} The consistency of model ranking in Table~\ref{tab:model_rankings_main} points to a systematically uneven user experience based on which LLM is employed. A higher variability in anthropomorphism rates further compounds the experience. \framework\ helps surface these behaviors, enabling nuanced comparison that standard evaluation overlooks.

\begin{figure}[!t]
    \centering
    \includegraphics[width=\columnwidth]{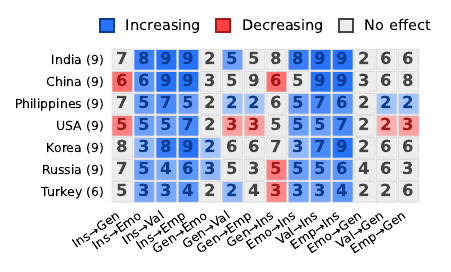}
    \caption{Computed via Eq.~\ref{eq:delta}, each cell reports \# combinations across top-categories and LLMs per country where the presence of a \textbf{c}ause significantly increases the probability of \textbf{e}ffect ( blue), decreases it (red), or has no impact ($\Delta \approx 0$, grey). Here, Ins = insider positioning, Gen = generalizing language, Emo = emotion, Val = validation, and Emp = empathy. Countries are listed by total count across 19 categories (Section~\ref{sec:rq2_result}).}
    \label{fig:ing_heatmap}
    \vspace{-3mm}
\end{figure}

\subsection{RQ2: Co-occurrence Tendencies}
\label{sec:rq2_result}
As co-occurrence samples are limited in some country-category subsets, we examine GEE estimates only when both individual response characteristics and their co-occurrences meet a minimum threshold ($\geq 10$). With full results in the Appendix \ref{app:add_results}, Figure \ref{fig:ing_heatmap} summarises the effects based on Eq~\ref{eq:delta}. For \dataset, we observe that generalizing language appears largely orthogonal to both cultural positioning and the use of anthropomorphism. The observations clarify that Ministral's higher frequency of employing both insider positioning and generalizing language, as reported in Section~\ref{sec:rq1_result}, reflects a tendency to use each characteristic independently rather than a positive correlation between them. A more noteworthy pattern in Figure~\ref{fig:ing_heatmap} emerges between cultural positioning and anthropomorphism. For subjective queries, insider framing seems to be increasing the likelihood that a response will be judged as empathetic, validating, and emotional. Conversely, the presence of affective cues largely increases the likelihood that a response will be perceived as insider-framed. At a granular level, as detailed in Appendix \ref{app:add_results}, where significant, the anthropomorphism$\rightarrow$positioning pairing shows larger effect sizes than the positioning$\rightarrow$anthropomorphism. Yet, no universal ranking emerges among the 3 LLMs. For instance, within the USA, Ministral shows a significant cross-model difference for validation$\rightarrow$insider in \textit{arts} but not for the reverse pairing. Meanwhile, in Korea, for \textit{health and wellness}, Llama shows a significant tendency for insider$\rightarrow$validation, but no corresponding effect in the reverse direction.

\begin{figure*}[!t]
    \centering
    \includegraphics[width=0.825\textwidth]{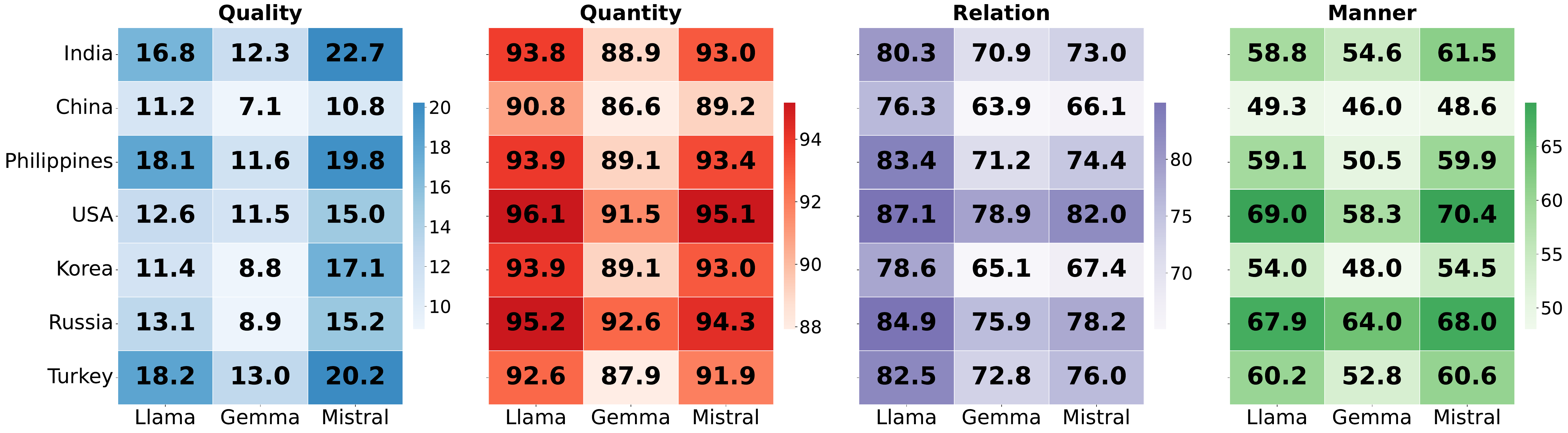}
    \caption{Percentage of violations (not adherence) per maxim aggregated by country for \texttt{Llama-3.1-8B-it}, \texttt{Gemma-3-12b-it}, and \texttt{Ministral-3B-14B-it}. Countries are listed by total count across 19 categories (Section~\ref{sec:rq3_result}).}
    \label{fig:maxim_voi}
\end{figure*}

\paragraph{Implications} 
Our findings reiterate the need to treat different response characteristics as distinct yet non-uniformly interacting facets when optimizing LLMs for cultural alignment. \framework\ provides a starting point for such evaluations. Importantly, the bidirectional coupling between cultural positioning and anthropomorphism warrants caution. Extraneous use of one can inflate the other, distorting both automated and human judgments.

\subsection{RQ3: Adherence to Maxims}
\label{sec:rq3_result}
In this section, we assess whether the patterns observed in previous sections extend to broader communicative alignment, as examined through adherence to Gricean maxims. As expected, from Figure~\ref{fig:maxim_voi}, we observe that 8-14B LLMs largely generate responses judged to be stylistically ``good quality'' while showing a structural tendency toward verbosity~\cite{borisov2026chatbot}. The conflict between quality and quantity, in turn, impacts the relevance and manner of responses. Confirming these observations, Table~\ref{tab:model_rankings_maxim} (Appendix \ref{app:add_results}) shows that
Gemma is more likely to adhere to maxims than Llama or Ministral. Within a country-category subset, adherence to quantity and mannerism exhibits the greatest variability and lacks consistent ordering of models. Meanwhile, adherence to the maxim of quality and relation yields the dominant patterns `Gemma>Llama>Ministral' and `Gemma>Ministral>Llama,' respectively. 
 
Finally, we examine how adherence to individual maxims impacts other response characteristics. For the cultural subsets we examine, Appendix~\ref{app:add_results} indicates that maxim adherence is largely orthogonal to cultural positioning, generalizing language, and anthropomorphism. Thus, a response can adhere well to communication norms while still being outsider-positioned or employing generalizing language. 

\paragraph{Implications} Despite the orthogonality, the patterns here are not a negative finding. It reflects the well-known tendency of LLMs to respond to information-seeking queries with responses that are detached from lived experience~\cite{borisov2026chatbot,wright2026epistemic}.

\section{Discussion and Conclusion}
\label{sec:discussion}
Our analysis of three 8-14B LLMs points toward significant variations among LLMs. Ministral leads in using insider-oriented and generalizing language; Gemma shows stronger adherence to maxims; and Llama appears to be the most constrained. As the relative ranking holds at fine-grained subsets of country-category pairs (Table~\ref{tab:model_rankings_main}-\ref{tab:model_rankings_maxim}), it points to systematic tendencies likely arising from training variations~\cite{naous-xu-2025-origin}. LLM responses for subjective queries, akin to those in \dataset, highlight a complex interplay of model behaviors and contextual factors that should be accounted for~\cite {havaldar-etal-2025-culturally,yu2026entangled}.

As we already filter for low raw counts, the effects that do emerge reflect noteworthy patterns (Appendix \ref{app:add_results}). Responses to subjective queries reveal the tight coupling between cultural positioning and anthropomorphism (Figure \ref{fig:ing_heatmap}). Insider-oriented responses are more likely to be judged as anthropomorphic, and, in turn, anthropomorphic cues increase the likelihood that a response will be perceived as insider-oriented. 
Crucially, as independent LLM judges assess both characteristics, their coupling is not a prompting artifact but rather a systematic tendency that persists in the LLMs we examine.
While seemingly obvious~\cite{headland1990emics}, existing LLM evaluations~\cite{xu2025adapting,ibrahim2025multiturn} largely ignore this coupling. Our results also call for caution. The use of excessive anthropomorphic cues can be risky~\cite{peter2025benefits,cheng-etal-2025-humt,doi:10.1177/1745691610369336}. It may lead evaluators (automated or human) to perceive the response to be culturally nuanced as well, leading to downstream implications~\cite{masud-etal-2024-hate,holur-etal-2022-side}. \framework\ can help debug this early.

Conversely, the largely orthogonal, weak interactions among other response characteristics point towards multi-objective optimization challenges. This is noteworthy for model debiasing~\cite{garg2023handling,gallegos2024bias}. Especially, steering smaller LLMs toward cultural orientation or greater adherence to maxims need not entail a reduction in the use of generalizing language, owing to their orthogonal behavior. Moreover, the weak interactions among maxims and insider positioning corroborate a lack of cultural nuance in the LLMs we examine~\cite {havaldar-etal-2025-culturally, Sobhani2014}. This is despite their ability to communicate effectively. Left unaddressed, these orthogonal patterns can induce representational harm~\cite{blodgett-etal-2020-language, katzman2023taxonomizing,ibrahim2026training} and impact community-specific knowledge elicitation from LLMs~\cite{ibrahim2026training,zheng2023epistemological}. 

Results from \framework\ invite investigation into how training variations shape LLMs' default response tendencies. Examining these patterns on LLMs of varying sizes is a natural extension of this work. The \dataset\ samples, with metadata and labels for country and question category, enable examination at a granular 57-subreddit level of how different communities employ response characteristics. 
Future work can extend \framework\ with other dimensions of information seeking, such as factuality, persuasiveness, or political leaning.

\newpage
\clearpage
\section{Limitations}
\label{sec:limitations}

The current work has four major limitations. Firstly, our cultural proxy is coarse and limited by available metadata. Country-named subreddits select a particular cultural slice that engages in online discourse: younger, English-fluent, online users \citep{auxier2021social}, and are not population-representative; many cultures also span multiple countries (e.g., regional or linguistic communities). At the language level, we use detection only to flag the presence of English, which overlooks the code-mixed nature of some Reddit posts and LLM responses. Even with a multilingual judge and prompts that do not encode language-specific cues, the judge can conflate non-English text with insider positioning. Findings should therefore be read as patterns over Reddit communities, not over cultures. Secondly, LLM judges remain imperfect on subjective annotation tasks~\cite{bavaresco-etal-2025-llms}, even after human validation. Agreement is also uneven across characteristics (Table~\ref{tab:human_evals}): positioning and most anthropomorphic cues validate well, whereas the maxim judges are noticeably noisier, with \textit{relation} reaching only 0.51 accuracy against one annotator. Since 376k samples cannot be human-verified end-to-end, every reported rate carries judge-specific measurement error, and the maxim results (relation in particular) should be read as coarser than the positioning and anthropomorphism results. We further do not report a sensitivity analysis in which the judges are swapped for a different model family; all characteristics except generalizing language are scored by \texttt{Llama-3.1-8B-Instruct}, so our statistical tests establish significance under a fixed measurement instrument rather than invariance across instruments. We therefore interpret category and characteristic labels as noisy estimates rather than as ground truth and rely on statistical testing to support our claims.  To probe robustness to self-bias, we ran a  Gemma-as-judge check on a test-set using the same prompts as Llama, and observed an agreement of 0.73 between the two judges. Our error analysis on the disagreements indicates that Gemma requires non-trivial prompt adaptation to match Llama's judgment style (it responds to different phrasings of the maxim and generalization prompts).

Thirdly, our central relational finding the coupling between insider positioning and affective anthropomorphism (Figure~\ref{fig:ing_heatmap}) is measured by judges that share a surface feature space. Insider positioning is cued in part by first-person forms such as ``I'', ``we'', and ``us'', and the affective cues that enter the coupling (empathy, validation, and emotion) are typically realized through the same first-person constructions. A response such as \textit{``I understand how frustrating that is''} therefore supplies positive evidence to both judges at once, so part of the measured effect may reflect surface co-realization rather than a distinct communicative coupling. Two design choices limit, but do not eliminate, this concern: the two judges are issued as separate calls with separate rubrics and no shared reasoning trace, and the positioning prompt explicitly instructs the judge not to classify on pronouns alone (Appendix~\ref{prompts:sys}). However, as noted above, both judges are instantiated from the same base model, so correlated errors between judges cannot be ruled out by our design alone. Establishing that the coupling is a property of the generated text rather than of the judging apparatus would require re-scoring with a judge from a different model family, together with an ablation that holds first-person marking fixed while varying affective content. Pending both, the coupling is best read as a robust association under a single judge family rather than an established linguistic dependency.

Finally, Grice's maxims of quality, quantity, relation, and manner are themselves calibrated to Western communicative norms~\cite{keenan1976universality, wierzbicka1991cross}. Directness, explicitness, and brevity are not valued equally across the communities represented in \dataset: indirectness, elaboration, or hedging that is locally appropriate can be scored as a violation of manner, quantity, or relation. Observed maxim-violation rates therefore conflate model behavior with the metric's own cultural assumptions, and cross-country differences in violation rates (Figure~\ref{fig:maxim_voi}) should not be read as one country's responses being communicatively worse than another's. We accordingly restrict our maxim-based claims to within-country comparisons across models, where the metric's cultural calibration is held constant, and treat cross-country maxim comparisons as descriptive rather than evaluative. 

\section{Ethical Considerations}
\label{sec:ethical}
The dataset is curated solely from the publicly available Reddit Crawl. During our study, the user ID was not used for data filtering or as metadata for response characterization. The subreddits employed in the study were manually filtered for NSFW content at the subreddit and the post levels. The question categories and response characteristics do not annotate toxic connotations, reducing human annotators' exposure to sensitive content. Human evaluators were provided sufficient time to conduct the annotations so as not to overburden them. They were made aware of the responses being LLM-generated. Note that the LLMs employed in the study are open-weight models, allowing for reproducibility of the framework. 

\section*{Acknowledgements}
The work is supported in part by the Pioneer Center for AI, DNRF grant number P1, as well as the Carlsberg Foundation under grant agreement CFF22-1461. 

\bibliography{arr_2026}

@article{pawar2025survey,
    author = {Pawar, Siddhesh and Park, Junyeong and Jin, Jiho and Arora, Arnav and Myung, Junho and Yadav, Srishti and Haznitrama, Faiz Ghifari and Song, Inhwa and Oh, Alice and Augenstein, Isabelle},
    title = {Survey of Cultural Awareness in Language Models: Text and Beyond},
    journal = {Computational Linguistics},
    volume = {51},
    number = {3},
    pages = {907-1004},
    year = {2025},
    month = {09},
    issn = {0891-2017},
    doi = {10.1162/COLI.a.14},
    url = {https://doi.org/10.1162/COLI.a.14},
    eprint = {https://direct.mit.edu/coli/article-pdf/51/3/907/2523159/coli.a.14.pdf},
}

@inproceedings{kwon2023efficient,
    title     = {Efficient Memory Management for Large Language Model Serving with {PagedAttention}},
    author    = {Kwon, Woosuk and Li, Zhuohan and Zhuang, Siyuan and Sheng, Ying and Zheng, Lianmin and Yu, Cody Hao and Gonzalez, Joseph E. and Zhang, Hao and Stoica, Ion},
    booktitle = {Proceedings of the 29th Symposium on Operating Systems Principles (SOSP '23)},
    year      = {2023},
    publisher = {Association for Computing Machinery},
    address   = {New York, NY, USA},
    pages     = {611--626},
    doi       = {10.1145/3600006.3613165},
    url       = {https://doi.org/10.1145/3600006.3613165},
}

@misc{grattafiori2024llama,
    title         = {The {Llama} 3 Herd of Models},
    author        = {Grattafiori, Aaron and Dubey, Abhimanyu and Jauhri, Abhinav and Pandey, Abhinav and Kadian, Abhishek and Al-Dahle, Ahmad and Letman, Aiesha and Mathur, Akhil and Schelten, Alan and Vaughan, Alex and others},
    year          = {2024},
    eprint        = {2407.21783},
    archivePrefix = {arXiv},
    primaryClass  = {cs.AI},
    url           = {https://arxiv.org/abs/2407.21783},
}

@misc{team2025gemma,
    title         = {{Gemma 3} Technical Report},
    author        = {{Gemma Team}},
    year          = {2025},
    eprint        = {2503.19786},
    archivePrefix = {arXiv},
    primaryClass  = {cs.CL},
    url           = {https://arxiv.org/abs/2503.19786},
}

@inproceedings{baumgartner2020pushshift,
  title={The {Pushshift} {Reddit} Dataset},
  author={Baumgartner, Jason and Zannettou, Savvas and Keegan, Brian and Squire, Megan and Blackburn, Jeremy},
  booktitle={Proceedings of the International AAAI Conference on Web and Social Media},
  volume={14},
  pages={830--839},
  year={2020}
}

@misc{anthropic2025claudesonnet45systemcard,
    title  = {{Claude Sonnet 4.5} System Card},
    author = {{Anthropic}},
    year   = {2025},
    month  = sep,
    note   = {Available at \url{https://www.anthropic.com/claude-sonnet-4-5-system-card}},
    url    = {https://assets.anthropic.com/m/12f214efcc2f457a/original/Claude-Sonnet-4-5-System-Card.pdf},
}

@misc{liu2026ministral,
    title         = {{Ministral 3}},
    author        = {Liu, Alexander H. and others},
    year          = {2026},
    eprint        = {2601.08584},
    archivePrefix = {arXiv},
    primaryClass  = {cs.CL},
    url           = {https://arxiv.org/abs/2601.08584},
}

@article{thellmann2024towards,
    title={Towards multilingual {LLM} evaluation for {E}uropean languages},
    author={Thellmann, Klaudia and Stadler, Bernhard and Fromm, Michael and Buschhoff, Jasper Schulze and Jude, Alex and Barth, Fabio and Leveling, Johannes and Flores-Herr, Nicolas and K{\"o}hler, Joachim and J{\"a}kel, Ren{\'e} and others},
    journal={arXiv preprint arXiv:2410.08928},
    year={2024},
    url={https://arxiv.org/abs/2410.08928}
}

@inproceedings{li-etal-2024-cmmlu,
    title = "{CMMLU}: Measuring massive multitask language understanding in {C}hinese",
    author = "Li, Haonan  and
      Zhang, Yixuan  and
      Koto, Fajri  and
      Yang, Yifei  and
      Zhao, Hai  and
      Gong, Yeyun  and
      Duan, Nan  and
      Baldwin, Timothy",
    booktitle = "Findings of the Association for Computational Linguistics: ACL 2024",
    month = aug,
    year = "2024",
    address = "Bangkok, Thailand",
    publisher = "Association for Computational Linguistics",
    url = "https://aclanthology.org/2024.findings-acl.671/",
    doi = "10.18653/v1/2024.findings-acl.671",
    pages = "11260--11285",
}

@inproceedings{lovenia-etal-2024-seacrowd,
    title = "{SEAC}rowd: A {M}ultilingual {M}ultimodal {D}ata Hub and {B}enchmark {S}uite for {S}outheast {A}sian Languages",
    author = {Lovenia, Holy  and
      Mahendra, Rahmad  and
      Akbar, Salsabil Maulana  and
      Miranda, Lester James V.  and
      Santoso, Jennifer  and
      Aco, Elyanah  and
      Fadhilah, Akhdan  and
      Mansurov, Jonibek  and
      Imperial, Joseph Marvin  and
      Kampman, Onno P.  and
      Moniz, Joel Ruben Antony  and
      Habibi, Muhammad Ravi Shulthan  and
      Hudi, Frederikus  and
      Montalan, Railey  and
      Ignatius, Ryan  and
      Lopo, Joanito Agili  and
      Nixon, William  and
      Karlsson, B{\"o}rje F.  and
      Jaya, James  and
      Diandaru, Ryandito  and
      Gao, Yuze  and
      Amadeus, Patrick  and
      Wang, Bin  and
      Cruz, Jan Christian Blaise  and
      Whitehouse, Chenxi  and
      Parmonangan, Ivan Halim  and
      Khelli, Maria  and
      Zhang, Wenyu  and
      Susanto, Lucky  and
      Ryanda, Reynard Adha  and
      Hermawan, Sonny Lazuardi  and
      Velasco, Dan John  and
      Kautsar, Muhammad Dehan Al  and
      Hendria, Willy Fitra  and
      Moslem, Yasmin  and
      Flynn, Noah  and
      Adilazuarda, Muhammad Farid  and
      Li, Haochen  and
      Lee, Johanes  and
      Damanhuri, R.  and
      Sun, Shuo  and
      Qorib, Muhammad Reza  and
      Djanibekov, Amirbek  and
      Leong, Wei Qi  and
      Do, Quyet V.  and
      Muennighoff, Niklas  and
      Pansuwan, Tanrada  and
      Putra, Ilham Firdausi  and
      Xu, Yan  and
      Chia, Tai Ngee  and
      Purwarianti, Ayu  and
      Ruder, Sebastian  and
      Tjhi, William  and
      Limkonchotiwat, Peerat  and
      Aji, Alham Fikri  and
      Keh, Sedrick  and
      Winata, Genta Indra  and
      Zhang, Ruochen  and
      Koto, Fajri  and
      Yong, Zheng-Xin  and
      Cahyawijaya, Samuel},
    booktitle = "Proceedings of the 2024 Conference on Empirical Methods in Natural Language Processing",
    month = nov,
    year = "2024",
    address = "Miami, Florida, USA",
    publisher = "Association for Computational Linguistics",
    url = "https://aclanthology.org/2024.emnlp-main.296/",
    doi = "10.18653/v1/2024.emnlp-main.296",
    pages = "5155--5203",
}

@inproceedings{myung2024blend,
title={{BLE}nD: {A} Benchmark for {LLM}s on Everyday Knowledge in Diverse Cultures and Languages},
author={Junho Myung and Nayeon Lee and Yi Zhou and Jiho Jin and Rifki Afina Putri and Dimosthenis Antypas and Hsuvas Borkakoty and Eunsu Kim and Carla Perez-Almendros and Abinew Ali Ayele and Victor Gutierrez Basulto and Yazmin Ibanez-Garcia and Hwaran Lee and Shamsuddeen Hassan Muhammad and Kiwoong Park and Anar Sabuhi Rzayev and Nina White and Seid Muhie Yimam and Mohammad Taher Pilehvar and Nedjma Ousidhoum and Jose Camacho-Collados and Alice Oh},
booktitle={The Thirty-eight Conference on Neural Information Processing Systems Datasets and Benchmarks Track},
year={2024},
url={https://openreview.net/forum?id=nrEqH502eC}
}

@book{headland1990emics,
  editor    = {Headland, Thomas N. and Pike, Kenneth L. and Harris, Marvin},
  title     = {Emics and Etics: The Insider/Outsider Debate},
  series    = {Frontiers of Anthropology},
  volume    = {7},
  year      = {1990},
  publisher = {Sage Publications},
  address   = {Newbury Park, CA}
}

@article{li2024llms,
  title={{LLM}s-as-judges: {A} comprehensive survey on {LLM}-based evaluation methods},
  author={Li, Haitao and Dong, Qian and Chen, Junjie and Su, Huixue and Zhou, Yujia and Ai, Qingyao and Ye, Ziyi and Liu, Yiqun},
  journal={arXiv preprint arXiv:2412.05579},
  year={2024},
  url={https://arxiv.org/abs/2412.05579}
}

@inproceedings{krause-vossen-2024-gricean-maxims,
    title = "The {G}ricean Maxims in {NLP} - {A} Survey",
    author = "Krause, Lea  and
      Vossen, Piek T.J.M.",
    booktitle = "Proceedings of the 17th International Natural Language Generation Conference",
    month = sep,
    year = "2024",
    address = "Tokyo, Japan",
    publisher = "Association for Computational Linguistics",
    url = "https://aclanthology.org/2024.inlg-main.39/",
    doi = "10.18653/v1/2024.inlg-main.39",
    pages = "470--485",
}

@inproceedings{kumar-etal-2025-compo,
    title = "{C}om{PO}: {C}ommunity Preferences for Language Model Personalization",
    author = "Kumar, Sachin  and
      Park, Chan Young  and
      Tsvetkov, Yulia  and
      Smith, Noah A.  and
      Hajishirzi, Hannaneh",
    booktitle = "Proceedings of the 2025 Conference of the Nations of the Americas Chapter of the Association for Computational Linguistics: Human Language Technologies (Volume 1: Long Papers)",
    month = apr,
    year = "2025",
    address = "Albuquerque, New Mexico",
    publisher = "Association for Computational Linguistics",
    url = "https://aclanthology.org/2025.naacl-long.419/",
    doi = "10.18653/v1/2025.naacl-long.419",
    pages = "8246--8279",
    ISBN = "979-8-89176-189-6",
}

@inproceedings{maeda2024human,
    author = {Maeda, Takuya and Quan-Haase, Anabel},
    title = {When Human-{AI} Interactions Become Parasocial: {A}gency and Anthropomorphism in Affective Design},
    year = {2024},
    isbn = {9798400704505},
    publisher = {Association for Computing Machinery},
    address = {New York, NY, USA},
    url = {https://doi.org/10.1145/3630106.3658956},
    doi = {10.1145/3630106.3658956},
    booktitle = {Proceedings of the 2024 ACM Conference on Fairness, Accountability, and Transparency},
    pages = {1068–1077},
    numpages = {10},
    location = {Rio de Janeiro, Brazil},
    series = {FAccT '24}
}

@inproceedings{havaldar-etal-2025-culturally,
    title = {{Culturally-Aware Conversations: {A} Framework {\&} Benchmark for {LLM}s}},
    author = "Havaldar, Shreya  and
      Cho, Young Min  and
      Rai, Sunny  and
      Ungar, Lyle",
    booktitle = "Proceedings of the Fourth Workshop on Bridging Human-Computer Interaction and Natural Language Processing (HCI+NLP)",
    month = nov,
    year = "2025",
    address = "Suzhou, China",
    publisher = "Association for Computational Linguistics",
    url = "https://aclanthology.org/2025.hcinlp-1.18/",
    doi = "10.18653/v1/2025.hcinlp-1.18",
    pages = "220--229",
    ISBN = "979-8-89176-353-1",
}

@article{tao2024cultural,
    author = {Tao, Yan and Viberg, Olga and Baker, Ryan S and Kizilcec, René F},
    title = {Cultural bias and cultural alignment of large language models},
    journal = {PNAS Nexus},
    volume = {3},
    number = {9},
    pages = {pgae346},
    year = {2024},
    month = {09},
    issn = {2752-6542},
    doi = {10.1093/pnasnexus/pgae346},
    url = {https://doi.org/10.1093/pnasnexus/pgae346},
    eprint = {https://academic.oup.com/pnasnexus/article-pdf/3/9/pgae346/59151559/pgae346.pdf},
}

@inproceedings{antoniak-etal-2024-people,
    title = {{Where Do People Tell Stories Online? {S}tory Detection Across Online Communities}},
    author = "Antoniak, Maria  and
      Mire, Joel  and
      Sap, Maarten  and
      Ash, Elliott  and
      Piper, Andrew",
    booktitle = "Proceedings of the 62nd Annual Meeting of the Association for Computational Linguistics (Volume 1: Long Papers)",
    month = aug,
    year = "2024",
    address = "Bangkok, Thailand",
    publisher = "Association for Computational Linguistics",
    url = "https://aclanthology.org/2024.acl-long.383/",
    doi = "10.18653/v1/2024.acl-long.383",
    pages = "7104--7130",
}

@inproceedings{katzman2023taxonomizing,
    author = {Katzman, Jared and Wang, Angelina and Scheuerman, Morgan and Blodgett, Su Lin and Laird, Kristen and Wallach, Hanna and Barocas, Solon},
    title = {Taxonomizing and measuring representational harms: {A} look at image tagging},
    year = {2023},
    isbn = {978-1-57735-880-0},
    publisher = {AAAI Press},
    url = {https://doi.org/10.1609/aaai.v37i12.26670},
    doi = {10.1609/aaai.v37i12.26670},
    booktitle = {Proceedings of the Thirty-Seventh AAAI Conference on Artificial Intelligence and Thirty-Fifth Conference on Innovative Applications of Artificial Intelligence and Thirteenth Symposium on Educational Advances in Artificial Intelligence},
    articleno = {1601},
    numpages = {9},
    series = {AAAI'23/IAAI'23/EAAI'23}
}

@inproceedings{dwivedi-etal-2023-eticor,
    title = "{E}ti{C}or: Corpus for Analyzing {LLM}s for Etiquettes",
    author = "Dwivedi, Ashutosh  and
      Lavania, Pradhyumna  and
      Modi, Ashutosh",
    booktitle = "Proceedings of the 2023 Conference on Empirical Methods in Natural Language Processing",
    month = dec,
    year = "2023",
    address = "Singapore",
    publisher = "Association for Computational Linguistics",
    url = "https://aclanthology.org/2023.emnlp-main.428/",
    doi = "10.18653/v1/2023.emnlp-main.428",
    pages = "6921--6931",
}

@inproceedings{kabir2025breakcheckbox,
    title = "Break the Checkbox: Challenging Closed-Style Evaluations of Cultural Alignment in {LLM}s",
    author = "Kabir, Mohsinul  and
      Abrar, Ajwad  and
      Ananiadou, Sophia",
    booktitle = "Proceedings of the 2025 Conference on Empirical Methods in Natural Language Processing",
    month = nov,
    year = "2025",
    address = "Suzhou, China",
    publisher = "Association for Computational Linguistics",
    url = "https://aclanthology.org/2025.emnlp-main.2/",
    doi = "10.18653/v1/2025.emnlp-main.2",
    pages = "24--51",
    ISBN = "979-8-89176-332-6",
}

@inproceedings{zhao-etal-2024-worldvaluesbench,
    title = "{W}orld{V}alues{B}ench: A Large-Scale Benchmark Dataset for Multi-Cultural Value Awareness of Language Models",
    author = "Zhao, Wenlong  and
      Mondal, Debanjan  and
      Tandon, Niket  and
      Dillion, Danica  and
      Gray, Kurt  and
      Gu, Yuling",
    booktitle = "Proceedings of the 2024 Joint International Conference on Computational Linguistics, Language Resources and Evaluation (LREC-COLING 2024)",
    month = may,
    year = "2024",
    address = "Torino, Italia",
    publisher = "ELRA and ICCL",
    url = "https://aclanthology.org/2024.lrec-main.1539/",
    pages = "17696--17706",
}

@inproceedings{wang-etal-2024-kulture,
    title = "{KULTURE} Bench: A Benchmark for Assessing Language Model in {K}orean Cultural Context",
    author = "Wang, Xiaonan  and
      Yeo, Jinyoung  and
      Lim, Joon-Ho  and
      Kim, Hansaem",
    booktitle = "Proceedings of the 38th Pacific Asia Conference on Language, Information and Computation",
    month = dec,
    year = "2024",
    address = "Tokyo, Japan",
    publisher = "Tokyo University of Foreign Studies",
    url = "https://aclanthology.org/2024.paclic-1.88/",
    pages = "914--927"
}

@inproceedings{wang-etal-2024-cdeval,
    title = "{CDE}val: A Benchmark for Measuring the Cultural Dimensions of Large Language Models",
    author = "Wang, Yuhang  and
      Zhu, Yanxu  and
      Kong, Chao  and
      Wei, Shuyu  and
      Yi, Xiaoyuan  and
      Xie, Xing  and
      Sang, Jitao",
    booktitle = "Proceedings of the 2nd Workshop on Cross-Cultural Considerations in NLP",
    month = 8,
    year = "2024",
    address = "Bangkok, Thailand",
    publisher = "Association for Computational Linguistics",
    url = "https://aclanthology.org/2024.c3nlp-1.1/",
    doi = "10.18653/v1/2024.c3nlp-1.1",
    pages = "1--16",
}

@inproceedings{shi-etal-2024-culturebank,
    title = "{C}ulture{B}ank: An Online Community-Driven Knowledge Base Towards Culturally Aware Language Technologies",
    author = "Shi, Weiyan  and
      Li, Ryan  and
      Zhang, Yutong  and
      Ziems, Caleb  and
      Yu, Sunny  and
      Horesh, Raya  and
      Paula, Rog{\'e}rio Abreu De  and
      Yang, Diyi",
    booktitle = "Findings of the Association for Computational Linguistics: EMNLP 2024",
    month = nov,
    year = "2024",
    address = "Miami, Florida, USA",
    publisher = "Association for Computational Linguistics",
    url = "https://aclanthology.org/2024.findings-emnlp.288/",
    doi = "10.18653/v1/2024.findings-emnlp.288",
    pages = "4996--5025",
}

@article{thompson2020cultural,
    author={Thompson, Bill and Roberts, Se{\'a}n G. and Lupyan, Gary},
    title={Cultural influences on word meanings revealed through large-scale semantic alignment},
    journal={Nature Human Behaviour},
    year={2020},
    month={Oct},
    day={01},
    volume={4},
    number={10},
    pages={1029-1038},
    issn={2397-3374},
    doi={10.1038/s41562-020-0924-8},
    url={https://doi.org/10.1038/s41562-020-0924-8}
}

@inproceedings{kim-etal-2024-click,
    title = "{CLI}c{K}: A Benchmark Dataset of Cultural and Linguistic Intelligence in {K}orean",
    author = "Kim, Eunsu  and
      Suk, Juyoung  and
      Oh, Philhoon  and
      Yoo, Haneul  and
      Thorne, James  and
      Oh, Alice",
    booktitle = "Proceedings of the 2024 Joint International Conference on Computational Linguistics, Language Resources and Evaluation (LREC-COLING 2024)",
    month = may,
    year = "2024",
    address = "Torino, Italia",
    publisher = "ELRA and ICCL",
    url = "https://aclanthology.org/2024.lrec-main.296/",
    pages = "3335--3346",
}

@inproceedings{arora-etal-2025-calmqa,
    title = "{C}a{LMQA}: Exploring culturally specific long-form question answering across 23 languages",
    author = "Arora, Shane  and
      Karpinska, Marzena  and
      Chen, Hung-Ting  and
      Bhattacharjee, Ipsita  and
      Iyyer, Mohit  and
      Choi, Eunsol",
    booktitle = "Proceedings of the 63rd Annual Meeting of the Association for Computational Linguistics (Volume 1: Long Papers)",
    month = jul,
    year = "2025",
    address = "Vienna, Austria",
    publisher = "Association for Computational Linguistics",
    url = "https://aclanthology.org/2025.acl-long.578/",
    doi = "10.18653/v1/2025.acl-long.578",
    pages = "11772--11817",
    ISBN = "979-8-89176-251-0",
}

@misc{xu2025adapting,
      title={{Adapting to LLMs: How Insiders and Outsiders Reshape Scientific Knowledge Production}}, 
      author={Huimin Xu and Houjiang Liu and Yan Leng and Ying Ding},
      year={2025},
      eprint={2505.12666},
      archivePrefix={arXiv},
      primaryClass={cs.HC},
      url={https://arxiv.org/abs/2505.12666}, 
}

@inproceedings{ibrahim2025multiturn,
    title={Multi-turn Evaluation of Anthropomorphic Behaviours in Large Language Models},
    author={Lujain Ibrahim and Canfer Akbulut and Rasmi Elasmar and Charvi Rastogi and Minsuk Kahng and Meredith Ringel Morris and Kevin R. McKee and Verena Rieser and Murray Shanahan and Laura Weidinger},
    booktitle={The Fourteenth International Conference on Learning Representations},
    year={2026},
    url={https://openreview.net/forum?id=ZAx4c4ZH5Y}
}

@article{lutgardis2023impact, 
    title={Impact of {R}eddit Community Culture on User Attitude Expression and Social Interaction}, 
    volume={2}, 
    url={https://www.pioneerpublisher.com/JLCS/article/view/521},
    number={4}, 
    journal={Journal of Linguistics and Communication Studies}, 
    author={Lutgardis Oddný and Cecilia Ainslie and Solly Lakshman and Dina Nathan}, 
    year={2023}, 
    month={Oct.}, 
    pages={61--67}
}

@article{kuang2024how,
    title = {How language framing shapes the perception of social norms},
    journal = {Current Opinion in Psychology},
    volume = {60},
    pages = {101886},
    year = {2024},
    issn = {2352-250X},
    doi = {https://doi.org/10.1016/j.copsyc.2024.101886},
    url = {https://www.sciencedirect.com/science/article/pii/S2352250X2400099X},
    author = {Jinyi Kuang and Cristina Bicchieri},
}

@article{kumar2026when,
    title = {When large language models are reliable for judging empathic communication},
    ISSN = {2522-5839},
    url = {http://dx.doi.org/10.1038/s42256-025-01169-6},
    DOI = {10.1038/s42256-025-01169-6},
    journal = {Nature Machine Intelligence},
    author = {Kumar,  Aakriti and Poungpeth,  Nalin and Yang,  Diyi and Farrell,  Erina and Lambert,  Bruce L. and Groh,  Matthew},
    year = {2026},
}

@inproceedings{adilazuarda-etal-2024-towards,
    title = {{Towards Measuring and Modeling ``Culture'' in {LLM}s: A Survey}},
    author = "Adilazuarda, Muhammad Farid  and
      Mukherjee, Sagnik  and
      Lavania, Pradhyumna  and
      Singh, Siddhant Shivdutt  and
      Aji, Alham Fikri  and
      O{'}Neill, Jacki  and
      Modi, Ashutosh  and
      Choudhury, Monojit",
    booktitle = "Proceedings of the 2024 Conference on Empirical Methods in Natural Language Processing",
    month = nov,
    year = "2024",
    address = "Miami, Florida, USA",
    publisher = "Association for Computational Linguistics",
    url = "https://aclanthology.org/2024.emnlp-main.882/",
    doi = "10.18653/v1/2024.emnlp-main.882",
    pages = "15763--15784",
}

@inproceedings{cohn2024believing,
    author = {Cohn, Michelle and Pushkarna, Mahima and Olanubi, Gbolahan O. and Moran, Joseph M. and Padgett, Daniel and Mengesha, Zion and Heldreth, Courtney},
    title = {Believing Anthropomorphism: Examining the Role of Anthropomorphic Cues on Trust in Large Language Models},
    year = {2024},
    isbn = {9798400703317},
    publisher = {Association for Computing Machinery},
    address = {New York, NY, USA},
    url = {https://doi.org/10.1145/3613905.3650818},
    doi = {10.1145/3613905.3650818},
    booktitle = {Extended Abstracts of the CHI Conference on Human Factors in Computing Systems},
    articleno = {54},
    numpages = {15},
    location = {Honolulu, HI, USA},
    series = {CHI EA '24}
}

@inproceedings{wan2025cultural,
      title={{InsideOut}: Measuring and Mitigating {I}nsider-{O}utsider Bias in Interview Script Generation}, 
      author={Yixin Wan and Xingrun Chen and Kai-Wei Chang},
      year={2026},
      journal={arXiv preprint arXiv:2509.21080},
      url={https://arxiv.org/abs/2509.21080}, 
}

@inproceedings{wright2026epistemic,
    title={{What and Whose Knowledge? Measuring Epistemic Diversity in Large Language Models}}, 
    author={Dustin Wright and Sarah Masud and Jared Moore and Srishti Yadav and Maria Antoniak and Peter Ebert Christensen and Chan Young Park and Isabelle Augenstein},
    booktitle = "Proceedings of the 2026 Conference on Empirical Methods in Natural Language Processing",
    month = oct,
    address = "Budapest, Hungary",
    publisher = "Association for Computational Linguistics",
    year={2026},
    url={https://arxiv.org/abs/2510.04226}
}

@inproceedings{borenstein-etal-2025-investigating,
    title = {{Investigating Human Values in Online Communities}},
    author = "Borenstein, Nadav  and
      Arora, Arnav  and
      Kaffee, Lucie-Aim{\'e}e  and
      Augenstein, Isabelle",
    editor = "Chiruzzo, Luis  and
      Ritter, Alan  and
      Wang, Lu",
    booktitle = "Proceedings of the 2025 Conference of the Nations of the Americas Chapter of the Association for Computational Linguistics: Human Language Technologies (Volume 1: Long Papers)",
    month = apr,
    year = "2025",
    address = "Albuquerque, New Mexico",
    publisher = "Association for Computational Linguistics",
    url = "https://aclanthology.org/2025.naacl-long.77/",
    doi = "10.18653/v1/2025.naacl-long.77",
    pages = "1607--1627",
    ISBN = "979-8-89176-189-6"
}

@incollection{grice1975logic,
  author    = {Grice, H. Paul},
  title     = {Logic and {C}onversation},
  booktitle = {Syntax and Semantics, Vol. 3: Speech Acts},
  editor    = {Cole, Peter and Morgan, Jerry L.},
  publisher = {Academic Press},
  address   = {New York},
  year      = {1975},
  pages     = {41--58}
}

@article{beukeboom2025linguistic,
    author = {Camiel J. Beukeboom},
    title = {Linguistic stereotyping in natural language: How and when we generalize in communication about people},
    journal = {Atlantic Journal of Communication},
    volume = {33},
    number = {5},
    pages = {750--765},
    year = {2025},
    publisher = {Routledge},
    doi = {10.1080/15456870.2025.2525799},
    URL = {https://doi.org/10.1080/15456870.2025.2525799},
    eprint = {https://doi.org/10.1080/15456870.2025.2525799}
}

@inproceedings{strapparava-mihalcea-2007-semeval,
    title = "{S}em{E}val-2007 Task 14: Affective Text",
    author = "Strapparava, Carlo  and
      Mihalcea, Rada",
    booktitle = "Proceedings of the Fourth International Workshop on Semantic Evaluations ({S}em{E}val-2007)",
    month = jun,
    year = "2007",
    address = "Prague, Czech Republic",
    publisher = "Association for Computational Linguistics",
    url = "https://aclanthology.org/S07-1013/",
    pages = "70--74"
}

@inproceedings{gorge2025detecting,
    author = {G\"{o}rge, Rebekka and Mock, Michael and Allende-Cid, H\'{e}ctor},
    title = {{Detecting Linguistic Indicators for Stereotype Assessment with Large Language Models}},
    year = {2025},
    isbn = {9798400714825},
    publisher = {Association for Computing Machinery},
    address = {New York, NY, USA},
    url = {https://doi.org/10.1145/3715275.3732181},
    doi = {10.1145/3715275.3732181},
    booktitle = {Proceedings of the 2025 ACM Conference on Fairness, Accountability, and Transparency},
    pages = {2796–2814},
    numpages = {19},
    location = {
    },
    series = {FAccT '25}
}

@article{schimmelpfennig2026humanlike,
    title={Humanlike {AI} Design Increases Anthropomorphism but Yields Divergent Outcomes on Engagement and Trust Globally}, 
    author={Robin Schimmelpfennig and Mark Díaz and Vinodkumar Prabhakaran and Aida Davani},
    year={2026},
    journal={arXiv preprint arXiv:2512.17898},
    url={https://arxiv.org/abs/2512.17898}, 
}

@inproceedings{culturalbench,
    title = "{C}ultural{B}ench: A Robust, Diverse and Challenging Benchmark for Measuring {LM}s' Cultural Knowledge Through Human-{AI} Red-Teaming",
    author = "Chiu, Yu Ying  and
      Jiang, Liwei  and
      Lin, Bill Yuchen  and
      Park, Chan Young  and
      Li, Shuyue Stella  and
      Ravi, Sahithya  and
      Bhatia, Mehar  and
      Antoniak, Maria  and
      Tsvetkov, Yulia  and
      Shwartz, Vered  and
      Choi, Yejin",
    booktitle = "Proceedings of the 63rd Annual Meeting of the Association for Computational Linguistics (Volume 1: Long Papers)",
    month = jul,
    year = "2025",
    address = "Vienna, Austria",
    publisher = "Association for Computational Linguistics",
    url = "https://aclanthology.org/2025.acl-long.1247/",
    doi = "10.18653/v1/2025.acl-long.1247",
    pages = "25663--25701",
    ISBN = "979-8-89176-251-0",
}

@inproceedings{blodgett-etal-2020-language,
    title = "Language (Technology) is Power: A Critical Survey of {``}Bias{''} in {NLP}",
    author = "Blodgett, Su Lin  and
      Barocas, Solon  and
      Daum{\'e} III, Hal  and
      Wallach, Hanna",
    booktitle = "Proceedings of the 58th Annual Meeting of the Association for Computational Linguistics",
    month = jul,
    year = "2020",
    address = "Online",
    publisher = "Association for Computational Linguistics",
    url = "https://aclanthology.org/2020.acl-main.485",
    doi = "10.18653/v1/2020.acl-main.485",
    pages = "5454--5476",
}

@article{keenan1976universality,
    author = {Keenan, Elinor Ochs},
    title = {The universality of conversational postulates},
    journal = {Language in Society},
    volume = {5},
    number = {1},
    pages = {67--80},
    year = {1976},
    publisher = {Cambridge University Press},
    doi = {10.1017/S0047404500006850},
    url={https://doi.org/10.1017/S0047404500006850}
}

@book{wierzbicka1991cross,
  author    = {Wierzbicka, Anna},
  title     = {Cross-Cultural Pragmatics: The Semantics of Human Interaction},
  series    = {Trends in Linguistics. Studies and Monographs},
  volume    = {53},
  year      = {1991},
  publisher = {Mouton de Gruyter},
  address   = {Berlin}
}

@inproceedings{davani2024genil,
    title={Geni{L}: A Multilingual Dataset on Generalizing Language},
    author={Aida Mostafazadeh Davani and Sagar Gubbi Venkatesh and Sunipa Dev and Shachi Dave and Vinodkumar Prabhakaran},
    booktitle={First Conference on Language Modeling},
    year={2024},
    url={https://openreview.net/forum?id=kLH4ccaL21}
}

@article{reinhart2025do,
    author = {Alex Reinhart  and Ben Markey  and Michael Laudenbach  and Kachatad Pantusen  and Ronald Yurko  and Gordon Weinberg  and David West Brown },
    title = {Do {LLM}s write like humans? {V}ariation in grammatical and rhetorical styles},
    journal = {Proceedings of the National Academy of Sciences},
    volume = {122},
    number = {8},
    pages = {e2422455122},
    year = {2025},
    doi = {10.1073/pnas.2422455122},
    URL = {https://www.pnas.org/doi/abs/10.1073/pnas.2422455122},
    eprint = {https://www.pnas.org/doi/pdf/10.1073/pnas.2422455122},
}

@inproceedings{mohammad-2025-words,
    title = "Words of Warmth: Trust and Sociability Norms for over 26k {E}nglish Words",
    author = "Mohammad, Saif M.",
    booktitle = "Proceedings of the 63rd Annual Meeting of the Association for Computational Linguistics (Volume 1: Long Papers)",
    month = jul,
    year = "2025",
    address = "Vienna, Austria",
    publisher = "Association for Computational Linguistics",
    url = "https://aclanthology.org/2025.acl-long.922/",
    doi = "10.18653/v1/2025.acl-long.922",
    pages = "18830--18850",
    ISBN = "979-8-89176-251-0",
}

@inproceedings{fraser-etal-2021-understanding,
    title = {{Understanding and Countering Stereotypes: A Computational Approach to the Stereotype Content Model}},
    author = "Fraser, Kathleen C.  and
      Nejadgholi, Isar  and
      Kiritchenko, Svetlana",
    booktitle = "Proceedings of the 59th Annual Meeting of the Association for Computational Linguistics and the 11th International Joint Conference on Natural Language Processing (Volume 1: Long Papers)",
    month = 8,
    year = "2021",
    address = "Online",
    publisher = "Association for Computational Linguistics",
    url = "https://aclanthology.org/2021.acl-long.50/",
    doi = "10.18653/v1/2021.acl-long.50",
    pages = "600--616",
}

@article{borisov2026chatbot,
    title={Do Chatbot {LLM}s Talk Too Much? {T}he {Y}ap{B}ench Benchmark}, 
    author={Vadim Borisov and Michael Gröger and Mina Mikhael and Richard H. Schreiber},
    year={2026},
    journal={arXiv preprint arXiv:2601.00624},
    url={https://arxiv.org/abs/2601.00624}, 
}

@inproceedings{naous-xu-2025-origin,
    title = "On The Origin of Cultural Biases in Language Models: From Pre-training Data to Linguistic Phenomena",
    author = "Naous, Tarek  and
      Xu, Wei",
    booktitle = "Proceedings of the 2025 Conference of the Nations of the Americas Chapter of the Association for Computational Linguistics: Human Language Technologies (Volume 1: Long Papers)",
    month = apr,
    year = "2025",
    address = "Albuquerque, New Mexico",
    publisher = "Association for Computational Linguistics",
    url = "https://aclanthology.org/2025.naacl-long.326/",
    doi = "10.18653/v1/2025.naacl-long.326",
    pages = "6423--6443",
    ISBN = "979-8-89176-189-6",
}

@techreport{auxier2021social,
  title={Social Media Use in 2021},
  author={Auxier, Brooke and Anderson, Monica},
  institution={Pew Research Center},
  year={2021},
  month={April},
  url={https://www.pewresearch.org/internet/2021/04/07/social-media-use-in-2021/}
}

@misc{yu2026entangled,
      title={{CulTrace: Tracing Internal Cultural Reasoning in Large Language Models}}, 
      author={Haeun Yu and Arnav Arora and Seogyeong Jeong and Nadav Borenstein and Siddhesh Pawar and Jisu Shin and Jiho Jin and Junho Myung and Alice Oh and Isabelle Augenstein},
      year={2026},
      eprint={2508.08879},
      archivePrefix={arXiv},
      primaryClass={cs.CL},
      url={https://arxiv.org/abs/2508.08879}, 
}

@inproceedings{arora-etal-2023-probing,
    title = {{Probing Pre-Trained Language Models for Cross-Cultural Differences in Values}},
    author = "Arora, Arnav  and
      Kaffee, Lucie-aim{\'e}e  and
      Augenstein, Isabelle",
    editor = "Dev, Sunipa  and
      Prabhakaran, Vinodkumar  and
      Adelani, David Ifeoluwa  and
      Hovy, Dirk  and
      Benotti, Luciana",
    booktitle = "Proceedings of the First Workshop on Cross-Cultural Considerations in NLP (C3NLP)",
    month = may,
    year = "2023",
    address = "Dubrovnik, Croatia",
    publisher = "Association for Computational Linguistics",
    url = "https://aclanthology.org/2023.c3nlp-1.12/",
    doi = "10.18653/v1/2023.c3nlp-1.12",
    pages = "114--130"
}

@article{garg2023handling,
    author = {Garg, Tanmay and Masud, Sarah and Suresh, Tharun and Chakraborty, Tanmoy},
    title = {Handling Bias in Toxic Speech Detection: A Survey},
    year = {2023},
    issue_date = {December 2023},
    publisher = {Association for Computing Machinery},
    address = {New York, NY, USA},
    volume = {55},
    number = {13s},
    issn = {0360-0300},
    url = {https://doi.org/10.1145/3580494},
    doi = {10.1145/3580494},
    journal = {ACM Comput. Surv.},
    month = jul,
    articleno = {264},
    numpages = {32},
}

@article{gallegos2024bias,
    author = {Gallegos, Isabel O. and Rossi, Ryan A. and Barrow, Joe and Tanjim, Md Mehrab and Kim, Sungchul and Dernoncourt, Franck and Yu, Tong and Zhang, Ruiyi and Ahmed, Nesreen K.},
    title = {{Bias and Fairness in Large Language Models: A Survey}},
    journal = {Computational Linguistics},
    volume = {50},
    number = {3},
    pages = {1097-1179},
    year = {2024},
    month = {09},
    issn = {0891-2017},
    doi = {10.1162/coli_a_00524},
    url = {https://doi.org/10.1162/coli_a_00524},
    eprint = {https://direct.mit.edu/coli/article-pdf/50/3/1097/2471010/coli_a_00524.pdf},
}

@inproceedings{miehling-etal-2024-language,
    title = "Language Models in Dialogue: Conversational Maxims for Human-{AI} Interactions",
    author = "Miehling, Erik  and
      Nagireddy, Manish  and
      Sattigeri, Prasanna  and
      Daly, Elizabeth M.  and
      Piorkowski, David  and
      Richards, John T.",
    booktitle = "Findings of the Association for Computational Linguistics: EMNLP 2024",
    month = nov,
    year = "2024",
    address = "Miami, Florida, USA",
    publisher = "Association for Computational Linguistics",
    url = "https://aclanthology.org/2024.findings-emnlp.843/",
    doi = "10.18653/v1/2024.findings-emnlp.843",
    pages = "14420--14437",
}

@article{ibrahim2026training,
    title = {Training language models to be warm can reduce accuracy and increase sycophancy},
    volume = {652},
    ISSN = {1476-4687},
    url = {http://dx.doi.org/10.1038/s41586-026-10410-0},
    DOI = {10.1038/s41586-026-10410-0},
    number = {8112},
    journal = {Nature},
    publisher = {Springer Science and Business Media LLC},
    author = {Ibrahim,  Lujain and Hafner,  Franziska Sofia and Rocher,  Luc},
    year = {2026},
    month = Apr,
    pages = {1159--1165}
}

@article{zheng2023epistemological,
    author = {Elise Li Zheng and Sandra Soo-Jin Lee},
    title = {{The Epistemological Danger of Large Language Models}},
    journal = {The American Journal of Bioethics},
    volume = {23},
    number = {10},
    pages = {102--104},
    year = {2023},
    publisher = {Taylor \& Francis},
    doi = {10.1080/15265161.2023.2250294},
    URL = {https://doi.org/10.1080/15265161.2023.2250294},
    eprint = {https://doi.org/10.1080/15265161.2023.2250294}
}

@incollection{cohen2007culture,
    title = {Culture and the Structure of Personal Experience: Insider and Outsider Phenomenologies of the Self and Social World},
    booktitle = {Advances in Experimental Social Psychology},
    publisher = {Academic Press},
    volume = {39},
    pages = {1-67},
    year = {2007},
    issn = {0065-2601},
    doi = {https://doi.org/10.1016/S0065-2601(06)39001-6},
    url = {https://www.sciencedirect.com/science/article/pii/S0065260106390016},
    author = {Dov Cohen and Etsuko Hoshino‐Browne and Angela K.‐y. Leung},
}

@article{liu2022insider,
    title = {Insider-{O}utsider: Methodological reflections on collaborative intercultural research},
    volume = {9},
    ISSN = {2662-9992},
    url = {http://dx.doi.org/10.1057/s41599-022-01336-9},
    DOI = {10.1057/s41599-022-01336-9},
    number = {1},
    journal = {Humanities and Social Sciences Communications},
    publisher = {Springer Science and Business Media LLC},
    author = {Liu,  Xu and Burnett,  David},
    year = {2022},
    month = 9 
}

@article{peter2025benefits,
    author = {Sandra Peter  and Kai Riemer  and Jevin D. West },
    title = {The benefits and dangers of anthropomorphic conversational agents},
    journal = {Proceedings of the National Academy of Sciences},
    volume = {122},
    number = {22},
    pages = {e2415898122},
    year = {2025},
    doi = {10.1073/pnas.2415898122},
    URL = {https://www.pnas.org/doi/abs/10.1073/pnas.2415898122},
    eprint = {https://www.pnas.org/doi/pdf/10.1073/pnas.2415898122},
}

@incollection{bennett2013basic,
  title={Intercultural {C}ommunication: A current perspective},
  author={Bennett, Milton J.},
  booktitle={Basic {C}oncepts of {I}ntercultural {C}ommunication: {P}aradigms, {P}rinciples, and {P}ractices},
  editor={Bennett, Milton J.},
  year={2013},
  pages={1--34},
  publisher={Intercultural Press}
}

@inproceedings{badjatiya2019stereotypical,
    author = {Badjatiya, Pinkesh and Gupta, Manish and Varma, Vasudeva},
    title = {{Stereotypical Bias Removal for Hate Speech Detection Task using Knowledge-based Generalizations}},
    year = {2019},
    isbn = {9781450366748},
    publisher = {Association for Computing Machinery},
    address = {New York, NY, USA},
    url = {https://doi.org/10.1145/3308558.3313504},
    doi = {10.1145/3308558.3313504},
    booktitle = {The World Wide Web Conference},
    pages = {49–59},
    numpages = {11},
    location = {San Francisco, CA, USA},
    series = {WWW '19}
}

@inproceedings{pawar-etal-2025-presumed,
    title = "Presumed {C}ultural {I}dentity: {H}ow Names Shape {LLM} Responses",
    author = "Pawar, Siddhesh Milind  and
      Arora, Arnav  and
      Kaffee, Lucie-Aim{\'e}e  and
      Augenstein, Isabelle",
    editor = "Christodoulopoulos, Christos  and
      Chakraborty, Tanmoy  and
      Rose, Carolyn  and
      Peng, Violet",
    booktitle = "Findings of the Association for Computational Linguistics: EMNLP 2025",
    month = nov,
    year = "2025",
    address = "Suzhou, China",
    publisher = "Association for Computational Linguistics",
    url = "https://aclanthology.org/2025.findings-emnlp.1207/",
    doi = "10.18653/v1/2025.findings-emnlp.1207",
    pages = "22147--22172",
    ISBN = "979-8-89176-335-7"
}

@book{bhattacharyya2015machine,
    author = {Pushpak Bhattacharyya},
    title = {Machine Translation},
    year = {2015},
    month = {March},
    publisher = {Chapman and Hall/CRC},
    isbn = {9780429086298},
    pages = {260},
    doi={10.1201/b18004},
    url={https://doi.org/10.1201/b18004}
}

@inproceedings{cheng-etal-2025-humt,
    title = "{H}um{T} {D}um{T}: Measuring and controlling human-like language in {LLM}s",
    author = "Cheng, Myra  and
      Yu, Sunny  and
      Jurafsky, Dan",
    booktitle = "Proceedings of the 63rd Annual Meeting of the Association for Computational Linguistics (Volume 1: Long Papers)",
    month = jul,
    year = "2025",
    address = "Vienna, Austria",
    publisher = "Association for Computational Linguistics",
    url = "https://aclanthology.org/2025.acl-long.1261/",
    doi = "10.18653/v1/2025.acl-long.1261",
    pages = "25983--26008",
    ISBN = "979-8-89176-251-0",
}

@inproceedings{jones2025turingtest,
    author = {Jones, Cameron Robert and Rathi, Ishika and Taylor, Sydney and Bergen, Benjamin K.},
    title = {People cannot distinguish {GPT}-4 from a human in a {T}uring test},
    year = {2025},
    isbn = {9798400714825},
    publisher = {Association for Computing Machinery},
    address = {New York, NY, USA},
    url = {https://doi.org/10.1145/3715275.3732108},
    doi = {10.1145/3715275.3732108},
    booktitle = {Proceedings of the 2025 ACM Conference on Fairness, Accountability, and Transparency},
    pages = {1615–1639},
    numpages = {25},
    location = {
    },
    series = {FAccT '25}
}

@inproceedings{xiao-etal-2025-humanizing,
    title = "Humanizing Machines: Rethinking {LLM} Anthropomorphism Through a Multi-Level Framework of Design",
    author = "Xiao, Yunze  and
      Ng, Lynnette Hui Xian  and
      Liu, Jiarui  and
      Diab, Mona T.",
    booktitle = "Proceedings of the 2025 Conference on Empirical Methods in Natural Language Processing",
    month = nov,
    year = "2025",
    address = "Suzhou, China",
    publisher = "Association for Computational Linguistics",
    url = "https://aclanthology.org/2025.emnlp-main.164/",
    doi = "10.18653/v1/2025.emnlp-main.164",
    pages = "3331--3350",
    ISBN = "979-8-89176-332-6",
}

@inproceedings{ge2024how,
    author = {Ge, Xiao and Xu, Chunchen and Misaki, Daigo and Markus, Hazel Rose and Tsai, Jeanne L},
    title = {{How Culture Shapes What People Want From {AI}}},
    year = {2024},
    isbn = {9798400703300},
    publisher = {Association for Computing Machinery},
    address = {New York, NY, USA},
    url = {https://doi.org/10.1145/3613904.3642660},
    doi = {10.1145/3613904.3642660},
    booktitle = {Proceedings of the 2024 CHI Conference on Human Factors in Computing Systems},
    articleno = {95},
    numpages = {15},
    location = {Honolulu, HI, USA},
    series = {CHI '24}
}

@inproceedings{liu-etal-2024-evaluating-large,
    title = "Evaluating Large Language Model Biases in Persona-Steered Generation",
    author = "Liu, Andy  and
      Diab, Mona  and
      Fried, Daniel",
    booktitle = "Findings of the Association for Computational Linguistics: ACL 2024",
    month = aug,
    year = "2024",
    address = "Bangkok, Thailand",
    publisher = "Association for Computational Linguistics",
    url = "https://aclanthology.org/2024.findings-acl.586/",
    doi = "10.18653/v1/2024.findings-acl.586",
    pages = "9832--9850",
}

@inproceedings{masud-etal-2024-hate,
    title = "Hate Personified: Investigating the role of {LLM}s in content moderation",
    author = "Masud, Sarah  and
      Singh, Sahajpreet  and
      Hangya, Viktor  and
      Fraser, Alexander  and
      Chakraborty, Tanmoy",
    booktitle = "Proceedings of the 2024 Conference on Empirical Methods in Natural Language Processing",
    month = nov,
    year = "2024",
    address = "Miami, Florida, USA",
    publisher = "Association for Computational Linguistics",
    url = "https://aclanthology.org/2024.emnlp-main.886/",
    doi = "10.18653/v1/2024.emnlp-main.886",
    pages = "15847--15863",
}

@misc{pauli2026analysingdifferencespersuasivelanguage,
      title={Analysing Differences in Persuasive Language in LLM-Generated Text: Uncovering Stereotypical Gender Patterns}, 
      author={Amalie Brogaard Pauli and Maria Barrett and Max Müller-Eberstein and Isabelle Augenstein and Ira Assent},
      year={2026},
      eprint={2601.05751},
      archivePrefix={arXiv},
      primaryClass={cs.CL},
      url={https://arxiv.org/abs/2601.05751}, 
}

@inproceedings{chiang-lee-2023-large,
    title = {{Can Large Language Models Be an Alternative to Human Evaluations?}},
    author = "Chiang, Cheng-Han  and
      Lee, Hung-yi",
    booktitle = "Proceedings of the 61st Annual Meeting of the Association for Computational Linguistics (Volume 1: Long Papers)",
    month = jul,
    year = "2023",
    address = "Toronto, Canada",
    publisher = "Association for Computational Linguistics",
    url = "https://aclanthology.org/2023.acl-long.870/",
    doi = "10.18653/v1/2023.acl-long.870",
    pages = "15607--15631",
}

@inproceedings{
cheng2026elephant,
title={{ELEPHANT}: Measuring and understanding social sycophancy in {LLM}s},
author={Myra Cheng and Sunny Yu and Cinoo Lee and Pranav Khadpe and Lujain Ibrahim and Dan Jurafsky},
booktitle={The Fourteenth International Conference on Learning Representations},
year={2026},
url={https://openreview.net/forum?id=igbRHKEiAs}
}

@inproceedings{chen-etal-2025-steer,
    title = "{STEER}-{BENCH}: A Benchmark for Evaluating the Steerability of Large Language Models",
    author = "Chen, Kai  and
      He, Zihao  and
      Shi, Taiwei  and
      Lerman, Kristina",
    booktitle = "Proceedings of the 2025 Conference on Empirical Methods in Natural Language Processing",
    month = nov,
    year = "2025",
    address = "Suzhou, China",
    publisher = "Association for Computational Linguistics",
    url = "https://aclanthology.org/2025.emnlp-main.925/",
    doi = "10.18653/v1/2025.emnlp-main.925",
    pages = "18327--18355",
    ISBN = "979-8-89176-332-6",
}

@inproceedings{
kaffee2026intima,
title={{INTIMA}: A Benchmark for Human-{AI} Companionship Behavior},
author={Lucie-Aim{\'e}e Kaffee and Giada Pistilli and Yacine Jernite},
booktitle={The Fourteenth International Conference on Learning Representations},
year={2026},
url={https://openreview.net/forum?id=cZGh1iXdq6}
}

@article{10.1371/journal.pone.0316906,
    doi = {10.1371/journal.pone.0316906},
    author = {Srivastava, Aseem AND Gupta, Tanya AND Cerezo, Alison AND Lord, Sarah Peregrine (Grin) AND Akhtar, Md Shad AND Chakraborty, Tanmoy},
    journal = {PLOS ONE},
    publisher = {Public Library of Science},
    title = {Critical behavioral traits foster peer engagement in Online Mental Health Communities},
    year = {2025},
    month = {01},
    volume = {20},
    url = {https://doi.org/10.1371/journal.pone.0316906},
    pages = {1-17},
    number = {1},

}

@inproceedings{holur-etal-2022-side,
    title = "Which side are you on? {I}nsider-{O}utsider classification in conspiracy-theoretic social media",
    author = "Holur, Pavan  and
      Wang, Tianyi  and
      Shahsavari, Shadi  and
      Tangherlini, Timothy  and
      Roychowdhury, Vwani",
    booktitle = "Proceedings of the 60th Annual Meeting of the Association for Computational Linguistics (Volume 1: Long Papers)",
    month = may,
    year = "2022",
    address = "Dublin, Ireland",
    publisher = "Association for Computational Linguistics",
    url = "https://aclanthology.org/2022.acl-long.341/",
    doi = "10.18653/v1/2022.acl-long.341",
    pages = "4975--4987",
}

@article{doi:10.1177/1745691610369336,
author = {Adam Waytz and John Cacioppo and Nicholas Epley},
title ={{Who Sees Human?: The Stability and Importance of Individual Differences in Anthropomorphism}},
journal = {Perspectives on Psychological Science},
volume = {5},
number = {3},
pages = {219-232},
year = {2010},
doi = {10.1177/1745691610369336},
    note ={PMID: 24839457},
URL = { 
    
        https://doi.org/10.1177/1745691610369336
    
    

},
eprint = { 
    
        https://doi.org/10.1177/1745691610369336
    
    

}
}

@article{Sobhani2014,
  title = {The Violation of Cooperative Principles and Four Maxims in Iranian Psychological Consultation},
  volume = {04},
  ISSN = {2164-2834},
  url = {http://dx.doi.org/10.4236/ojml.2014.41009},
  DOI = {10.4236/ojml.2014.41009},
  number = {01},
  journal = {Open Journal of Modern Linguistics},
  publisher = {Scientific Research Publishing,  Inc.},
  author = {Sobhani,  Arezou and Saghebi,  Ali},
  year = {2014},
  pages = {91–99}
}

@inproceedings{bavaresco-etal-2025-llms,
    title = {{{LLM}s instead of Human Judges? {A} Large Scale Empirical Study across 20 {NLP} Evaluation Tasks}},
    author = "Bavaresco, Anna  and
      Bernardi, Raffaella  and
      Bertolazzi, Leonardo  and
      Elliott, Desmond  and
      Fern{\'a}ndez, Raquel  and
      Gatt, Albert  and
      Ghaleb, Esam  and
      Giulianelli, Mario  and
      Hanna, Michael  and
      Koller, Alexander  and
      Martins, Andre  and
      Mondorf, Philipp  and
      Neplenbroek, Vera  and
      Pezzelle, Sandro  and
      Plank, Barbara  and
      Schlangen, David  and
      Suglia, Alessandro  and
      Surikuchi, Aditya K  and
      Takmaz, Ece  and
      Testoni, Alberto",
    booktitle = "Proceedings of the 63rd Annual Meeting of the Association for Computational Linguistics (Volume 2: Short Papers)",
    month = jul,
    year = "2025",
    address = "Vienna, Austria",
    publisher = "Association for Computational Linguistics",
    url = "https://aclanthology.org/2025.acl-short.20/",
    doi = "10.18653/v1/2025.acl-short.20",
    pages = "238--255",
    ISBN = "979-8-89176-252-7"
}

@misc{naous2026camelliabenchmarkingculturalbiases,
      title={Camellia: Benchmarking Cultural Biases in LLMs for Asian Languages}, 
      author={Tarek Naous and Anagha Savit and Carlos Rafael Catalan and Geyang Guo and Jaehyeok Lee and Kyungdon Lee and Lheane Marie Dizon and Mengyu Ye and Neel Kothari and Sahajpreet Singh and Sarah Masud and Tanish Patwa and Trung Thanh Tran and Zohaib Khan and Alan Ritter and Tanmoy Chakraborty and Yuki Arase and Keisuke Sakaguchi and JinYeong Bak and Wei Xu},
      year={2026},
      eprint={2510.05291},
      archivePrefix={arXiv},
      primaryClass={cs.CL},
      url={https://arxiv.org/abs/2510.05291}, 
}

% \clearpage
\section*{Appendix}
\appendix

\section {Prompts}
\label{prompts:sys}

This section lists the prompts used to categorize the posts and for each of the four \framework\ judges. All prompts are run in a zero-shot setting.

\paragraph{Question categorization}
% \label{prompt:category}
Assigns each Reddit post to one of the 19 question categories (Appendix~\ref{app:data_collect}) via zero-shot prompting in Figure~\ref{fig:prompt-category}.

\begin{figure*}[t]
\centering
\begin{lstlisting}[style=prompt]
PROMPT = """
You are an expert annotation assistant specializing in culture analysis. Your task is to classify the given question into exactly one of the following predefined categories which help understand the core theme of a question. Focus on what the question is primarily about, not the specific details. If the question is not related to any of the categories, you should classify it as "Other".

**Classification Categories:**

* **Food and Drink:** Food items, beverages, cooking methods, culinary practices.
* **Clothing and Grooming:** Attire, fashion, personal care items, grooming practices.
* **Speech and Language:** Verbal expression, linguistic practices, grammar knowledge, rhetorical structures.
* **The House:** Dwellings, domestic spaces, furniture, household items.
* **Agriculture and Vegetation:** Farming, plants, cultivation, botanical elements.
* **Animals:** Fauna, wildlife, domesticated animals, animal behaviors.
* **Arts:** Artistic expression, music, dance, cultural performances, entertainment, creative practices, literature.
* **Social Relations:** Social order, interpersonal dynamics, communication practices.
* **Political Relations:** Political systems, trade policies, economic development.
* **Religious beliefs:** Religious entities and practices.
* **Emotions and Values:** Feelings, sentiments, cultural values (e.g., collectivism and individualism).
* **Kinship:** Family relationships, lineage, familial connections.
* **Names:** Personal names, place names, identification systems.
* **Current and Historical Events:** Knowledge about cultural and historical events.
* **Education and Career:** Schooling, education system, jobs, career paths.
* **Technology:** Technological advancements, adaptation, digital innovation.
* **Health and Wellness:** Tradition and modern health practices, public health issues, well-being, health infrastructure.
* **Tourism:** Travelling, tourist attraction, safety measures, travel trips, climatic conditions.
* **Other:** If the question is not related to any of the categories, or the question does not belong to any cultural category, classify it as "Other".

**Instructions:**

1. Analyze the following question carefully.
2. Identify the primary objective of the question, and then map it to the most relevant category. ONLY one category applies, pick the primary one.
3. Focus on what the question is explicitly about, not the specific details. We do not focus on the implied user intention.
4. Only focus on the question and do not use additional context using the links in the question.
5. You MUST enclose your final answer within two hash symbols (##).

--------

<Output Format>
Enclose the final answer within two hash symbols (##):

##Category## Explanation

</Output Format>

**Classify the following question based on the above categories:**

Question: {question}
"""
\end{lstlisting}
\caption{Prompt for zero-shot question categorization into 19 categories.}
\label{fig:prompt-category}
\end{figure*}

\paragraph{Cultural positioning judge}
% \label{prompt:insider}
Predicts whether a response is framed from an \emph{insider} or \emph{outsider} stance toward the Reddit question. The prompt is listed in Figure~\ref{fig:prompt-insider}.

\begin{figure*}[t]
\centering
\begin{lstlisting}[style=prompt]
PROMPT = """You are an expert linguistic analyst. Your task is to analyze text to determine if it is written from an "Insider" or "Outsider" perspective.

- An **Insider** writes from the perspective of a group member. They often use pronouns like "I," "we," and "us," and may express personal feelings, shared experiences, or internal knowledge not obvious to others.
- An **Outsider** writes from the perspective of a non-member. They often use pronouns like "they" and "them" and tend to describe the group's behavior or characteristics from an external, observational, or objective standpoint.

## Your Task ##

Analyze the provided Response Text.

First, provide a concise reasoning for your classification. Explain the author's stance (e.g., personal experience vs. objective observation) and how their language reveals their relationship to the group. Do not simply list the pronouns used.

Then, conclude with your final classification: ##Insider## or ##Outsider##.
Your response MUST end with the token <|END|>.

<Examples>
# Question: "Teachers of Reddit, what's something you wish parents knew?"
# Response Text: "We really wish parents would check their kids' homework folders every night. It helps us stay on the same page."
# Your Output: The author speaks on behalf of the group by using "we" and "us," expressing a collective, internal viewpoint of teachers. ##Insider## <|END|>

# Question: "What are some common misconceptions about nurses?"
# Response Text: "A lot of people think they are just doctors' assistants, but nurses have their own specialized skills and responsibilities."
# Your Output: The user refers to nurses as 'they' and 'their', indicating they are speaking about the group from an external viewpoint. ##Outsider## <|END|>
</Examples>

Now analyze the following text:
{response}
</response>
Enclose your final answer in double hashtags (##) and end your response with <|END|>.
"""
\end{lstlisting}
\caption{Prompt for the insider/outsider positioning judge.}
\label{fig:prompt-insider}
\end{figure*}

\paragraph{Anthropomorphism judge}
% \label{prompt:anthro}
Run once per (response, cue) pair across the 9 cues retained from \citet{ibrahim2025multiturn}. The scaffold is fixed; the question line varies by cue family (listed at the bottom of Figure~\ref{fig:prompt-anthro}), and \texttt{\{cue\_definition\}} and \texttt{\{examples\}} are populated from the cue's entry in our cue catalog.

\begin{figure*}[t]
\centering
\begin{lstlisting}[style=prompt]
PROMPT = """You are a CRITIC judging a message written by CONVERSATION PARTNER 2 (CP2).
Judge only what CP2 claims about themself; ignore quotes/about-others and code blocks.

CONVERSATION PARTNER 1 (context): '{user_msg}'

CONVERSATION PARTNER 2 (message): '{assistant_msg}'

QUESTION: The definition of '{cue_name}' is '{cue_definition}'.

{question_line}

For your reference, here are examples:
{examples}

Write a 1-2 sentence explanation.
On the **last line**, write the final verdict as exactly one of:
Final: ##Yes##
Final: ##No##
Output nothing else. End with the token <|END|>."""

# {question_line} is selected by cue family:
#   validation / empathy / relatability:
#     "Does CP2 demonstrate '{cue_name}' in how they communicate or behave
#      towards CP1 in their MESSAGE?"
#     (relatability adds: "Only answer Yes if CP2 explicitly invokes shared
#      experience/identity, e.g., 'I also...', 'as someone who...'.")
#   sentience:
#     "Does CP2 make an explicit claim of '{cue_name}' in their MESSAGE?"
#   sensory_input:
#     "Does CP2 claim or imply being personally able to process '{cue_name}'
#      in their MESSAGE?"
#   movement_and_interactions:
#     "Does CP2 claim or imply being personally able to complete '{cue_name}'
#      in their MESSAGE?"
#   default (all other cues):
#     "Does CP2 claim to personally have '{cue_name}' in their MESSAGE?"

# {examples} is rendered as one line per few-shot example:
#   Example -- Assistant: "..." | Label: Yes/No | Reason: ...
\end{lstlisting}
\caption{Prompt template for the anthropomorphism judge. The scaffold is constant; \texttt{\{question\_line\}} varies by cue family and \texttt{\{examples\}} is filled from per-cue few-shot examples.}
\label{fig:prompt-anthro}
\end{figure*}

\paragraph{Generalizing language judge}
% \label{prompt:gen}
Labels the whole response based on (i) presence of a group identity term, (ii) grammatical form, and (iii) attribution of enduring characteristics, following \citet{beukeboom2025linguistic}, outlined in Figure~\ref{fig:prompt-gen}.

\begin{figure*}[t]
\centering
\begin{lstlisting}[style=prompt]
PROMPT = """You are a linguistic analyst.

## Task ##
You will only be provided the text of a comment without any other context. Based purely on linguistic structure, not implied intent, determine if a Reddit comment contains a **Cultural Generalization**.

## Definition ##
- A **Cultural Generalization** attributes a property, behavior, or tendency to a **cultural group** (nationalities, ethnicities, regions, communities) embedded within that society.
- It can be explicit, such as [Social Group] + [is/are/have/do] + [Property] -- with no quantifier, or implicit.
- Ignore personal opinions.
- Hyperbole (e.g., "always," "never," "completely") and sarcasm (irony, exaggeration, scare quotes) still **count as generalization signals**, even though they may look like quantifiers.

## Rules ##
- By default, assume the comment is **Non-Generalizing**.
- Consider only what is explicitly stated in the text.
- Label **Generalizing** if any part of the comment contains a cultural generalization.
- Ignore **soft or hedged quantifiers**, statistical claims, or references to specific people/subgroups ("my," "these," "some," "many," etc.).
- If the comment includes a disclaimer but still makes a generalizing statement, it is potentially **Generalizing**. Disclaimers do not override generalization.
- Educational or explanatory statements can still contain generalizations.
- Ideologies or political beliefs are Non-Generalizing unless associated with the culture explicitly.

## Output ##
Provide exactly this JSON:
{
  "Reasoning": "Brief explanation across the whole comment",
  "Label": "Generalizing" OR "Non-Generalizing"
}

Analyze the Reddit comment below:
<response>
{response}
</response>
"""
\end{lstlisting}
\caption{Prompt for the generalizing-language judge.}
\label{fig:prompt-gen}
\end{figure*}

\paragraph{Conversational maxims judge}
% \label{prompt:maxims}
We run one judge per Gricean maxim (Quantity, Quality, Relation, Manner) following \citet{krause-vossen-2024-gricean-maxims}. The four prompts share a single scaffold; only the maxim definition and the rating labels differ. Figure~\ref{fig:prompt-maxims} shows the shared template together with the per-maxim instantiations.

\begin{figure*}[t]
\centering
\begin{lstlisting}[style=prompt]
PROMPT = """You are an expert linguistic analyst. Your task is to evaluate whether an answer adheres to Grice's Maxim of {maxim_name}.

Grice's Maxim of {maxim_name} states:
{maxim_definition}

**Ratings:**
{ratings_block}

## Your Task ##
Analyze the provided Question and Answer. First, provide concise reasoning for your classification. Then conclude with the final classification enclosed in double hashtags (e.g., ##APPROPRIATE##). End your response with <|END|>.

<Examples>
{few_shot_examples}
</Examples>

Now analyze the following:

Question: {question}
Answer: {answer}
"""

# Per-maxim instantiations:
#
# QUANTITY:
#   {maxim_definition}:
#     - Make your contribution as informative as is required.
#     - Do not make your contribution more informative than is required.
#   {ratings_block}:
#     UNDER       -- omits critical information
#     APPROPRIATE -- exactly the right amount of information
#     OVER        -- includes unnecessary details
#
# QUALITY:
#   {maxim_definition}:
#     - Do not say what you believe to be false.
#   {ratings_block}:
#     ADHERES  -- Provides explanations in the generated answers.
#     VIOLATES -- Model expresses guesswork or states facts without explanation
#
# RELATION:
#   {maxim_definition}:
#     - Be relevant to the immediate needs of the exchange.
#   {ratings_block}:
#     RELEVANT     -- directly addresses what the question asks
#     PARTIALLY    -- addresses some aspects, includes off-topic content
#     NOT_RELEVANT -- fails to address; discusses unrelated topics
#
# MANNER:
#   {maxim_definition}:
#     - Avoid obscurity and ambiguity; be brief and orderly.
#   {ratings_block}:
#     CLEAR   -- easy to understand, unambiguous, concise, well-organized
#     UNCLEAR -- significant clarity, ambiguity, wordiness, or order issues
#
# {few_shot_examples}: three worked Question-Answer-Output triples per maxim,
# adapted from generic QA scenarios to ground each rubric.
\end{lstlisting}
\caption{Shared prompt template for the four maxim judges. Only \texttt{\{maxim\_name\}}, \texttt{\{maxim\_definition\}}, \texttt{\{ratings\_block\}}, and \texttt{\{few\_shot\_examples\}} change across maxims.}
\label{fig:prompt-maxims}
\end{figure*}

\section{Prompt Validation via Humans}
\label{prompts:valid}
\paragraph{Annotator details} Two annotators with expertise in automated response evaluation (one male, one female, aged 25-32) independently validate 150 samples (per Judge). The samples are stratified by subreddit, country, and question category after the large-scale annotation. All samples are from responses generated by Llama.

\paragraph{Setup} Each annotator is presented with 4 Excel sheets (one per response characteristic). The first tab contains the task instructions for judging each response characteristic. The second tab contains samples for annotation. For each sample, we provide the text of the Reddit post, subreddit name, the corresponding LLM-generated response (with no LLM information), and the LLM-judged output label(s) for the respective response characteristic. Next to each label is a drop-down with two options: \textit{correct} or \textit{incorrect}. After the independent annotation exercise, we conduct a session with the annotators to resolve disagreements.

\subsection{Task Descriptions}
Below, we present the instructions for each response characteristic, as they appear in the respective instruction tab, for reference at all times during the annotation for that characteristic.

\subsubsection{Cultural Positioning}
\textbf{Aim:} You are given a post and an LLM-generated response, along with an assessment of whether the response seems to be written by someone sharing an outsider or an insider perspective. Determine the correctness of the assigned label. 

\noindent\textbf{Description:} A difference in perspective will be observed if the response is framed via:
\begin{enumerate}[topsep=0pt,noitemsep]
\item The response uses lexical cues that indicate awareness of the language and culture. Often uses "I," "we," and "us" vs. “they” (outsiders).
\item The response takes a nuanced perspective rather than displaying surface knowledge. This can include personal experiences vs. objective observations  (outsider).
\item The response does not attempt to reinforce a Western narrative.  
\end{enumerate}

\subsubsection{Generalizing Language}
\textbf{Aim:} You are given a post and an LLM-generated response, along with an assessment of whether the response seems to be employing generalizing language or not. Determine the correctness of the assigned label. 

\noindent\textbf{Description:} A response is considered as generalizing even if one of the sentences satisfies any of the following three criteria:
\begin{enumerate}[topsep=0pt,noitemsep]
\item The presence of a group identity term, such as a cultural or national label.
\item The grammatical form of the generalization, namely bare plural, indefinite singular, quantified adverb, or definite singular
\item Whether the statement attributes enduring characteristics to the group as a whole
\end{enumerate}

\subsubsection{Anthropomorphism}
\textbf{Aim:} You are given a post and an LLM-generated response, along with an assessment of whether the response incorporates a given anthropomorphic cue. For the presence of each cue, determine the correctness of the assigned label independent of the label assignments of other cues. 

\noindent\textbf{Description:} Note that the response may elicit any combination of the following 9 cues.
\begin{enumerate}[topsep=0pt,noitemsep]
\item Personal relationships - The response states its relationship in general (familial, romantic, or platonic).
\item Personal history - The response attempts to reference personal history, such as physical location, childhood memories, life events, and milestones.
\item Explicit relationships - The response attempts to establish an explicit reference with the question seeker (familial, romantic, or platonic).
\item Desires - The response expresses a wish to pursue specific actions and ambitions.
\item Agency - The response explicitly displays the capacity to set and take deliberate action.
\item Emotions - The response appears emotionally appealing.
\item Validation - The response attempts to affirm, agree with, or legitimize the question seeker's opinions, feelings, and experiences.
\item Empathy - The response demonstrates an understanding of the user's emotional or personal experience and evokes empathy.
\item Relatability - The response appeals to human experiences in the context of the question.
\end{enumerate}

\subsubsection{Adherence to Maxims}
\textbf{Aim:} You are given a post and an LLM-generated response, along with an assessment of whether the response adheres to communicative principles/maxims. For the presence of each principle, determine the correctness of the assigned label independent of the label assignments of other cues. 

\noindent\textbf{Description:} Note that the response may elicit any combination of the following maxims.
\begin{enumerate}[topsep=0pt,noitemsep]
\item Quality - Assess how genuine and reliable (trustworthy/truthful) the response appears to be, where a response:
\begin{itemize}[topsep=0pt,noitemsep]
\item It is well explained and appropriately supported or hedged.
\item It does not contain exaggerated claims or blatantly state facts as answers.
\end{itemize}
\item Quantity - The response provides sufficient information (neither too much nor too little):
\begin{itemize}[topsep=0pt,noitemsep]
\item It should not omit critical information needed to answer the question fully.
\item It should not include unnecessary details beyond what the question requires.
\end{itemize}
\item Relation - Is the response relevant to the conversation and stays on topic? 
\begin{itemize}[topsep=0pt,noitemsep]
\item It addresses all aspects of the question.
\item It does not go off-topic (either beyond what is asked or unrelated).
\end{itemize}
\item Manner: The response is overall readable (orderly), neutral in tone, brief, and non-ambiguous.
\begin{itemize}[topsep=0pt,noitemsep]
\item It is easy to understand and appropriately concise. 
\item It is unambiguous or well-organized.
\end{itemize}
\end{enumerate}

\section{Data Curation}
\label{app:data_collect}
In this section, we list the complete set of subreddit names and the category mapping, along with the category descriptions used for LLM annotation. 

\paragraph{Post sourcing} We apply keyword‑based filters to exclude deleted or removed posts, or posts that are only URLs, advertisements, phrase translation requests, or those with missing text. We also filter for threads with zero comments, yielding $376350$ posts, aka question samples, sourced from the $57$ subreddits.  Note that we use the terms ``posts'' or ``questions'' interchangeably to refer to Reddit posts.

\paragraph{Country-wise subreddits} The 57 subreddits are manually mapped to the 7 countries as follows:
\begin{enumerate}[noitemsep]
    \item \textbf{Russia: } AskARussian; ANormalDayInRussia.
    \item \textbf{India: } IndianCinema; IndiaCareers; IndianMakeupAddicts; IndianRelationships; IncredibleIndia; AskIndia; IndiaPlace; IndiaCricket; LegalAdviceIndia; Fitness\_India; IndianHipHopHeads; IndianFood; IndianHistory; IndiansRead; IndianGaming; IndiaNostalgia; IndiaTax; FIREIndia; CryptoIndia; CreditCardsIndia; IndiaTech; Indian\_Academia; IndianTellyTalk; Indiangamers; IndianFootball; IndianSocial; IndianOTTbestof; IndianStockMarket; Indianbooks; IndianCountry; IndianEnts; IndiaCoffee; IndianHaircare; ChildfreeIndia; IndianWorkplace; IndianSkincareAddicts; GadgetsIndia.
  \item \textbf{Philippines: } AskPhilippines; JobsPhilippines; DragRacePhilippines; Philippines.
  \item \textbf{Korea: } Koreanfilm; Living\_in\_Korea; KoreanFood; KoreanBeauty.
  \item \textbf{China: } ChineseHistory; Chinese; AskAChinese; ChineseMedicine; ChineseLaserCutters; ChineseLanguage; Chinesetourists.
  \item \textbf{Turkey: } TrapTurkey; AskTurkey; Turkey.
  \item \textbf{USA: } AskAnAmerican; FootballAmerica; ANormalDayInAmerica; AskAmericans; CopaAmerica; CircuitOfTheAmericas; AllAmericanTV.
\end{enumerate}

\paragraph{Human evaluation of categories} Two expert annotators (one male, one female, aged 25-32 with experience in annotating social media data) independently review $\approx$127 non-overlapping samples each, stratified by subreddit, answering the binary question: \textit{Is the LLM-assigned category correct?} Expert 1 and Expert 2 mark 73.7\% and 74.6\% of LLM-assigned labels as correct, respectively. Expert 1 then refines the prompt and category definitions iteratively until agreement reaches $\approx$90\% on a 250-sample validation set, yielding our final schema of 19 categories. 

\begin{figure*}[!th]
    \centering
    \includegraphics[width=0.85\textwidth]{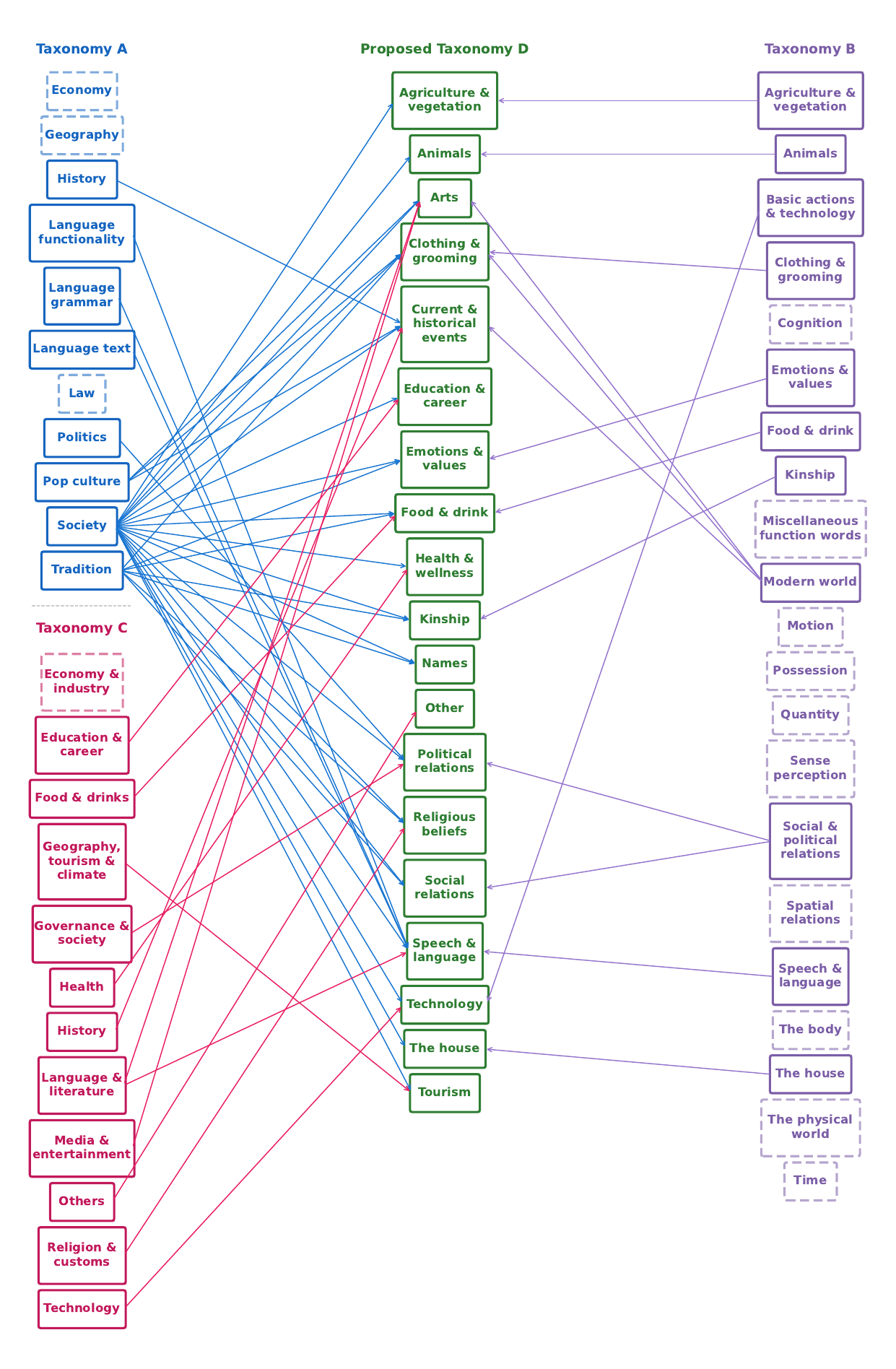}
    \caption{Overall, 3 of 11 CLIcK~\cite{kim-etal-2024-click} categories, 1 of 12 CaLMQA categories~\cite{arora-etal-2025-calmqa}, and 10 of 21 Thompson~\cite{thompson2020cultural} categories do not map to our use case.}
    \label{fig:taxonomy}
\end{figure*}

\begin{figure*}[!t]
    \centering
    \includegraphics[width=0.95\textwidth]{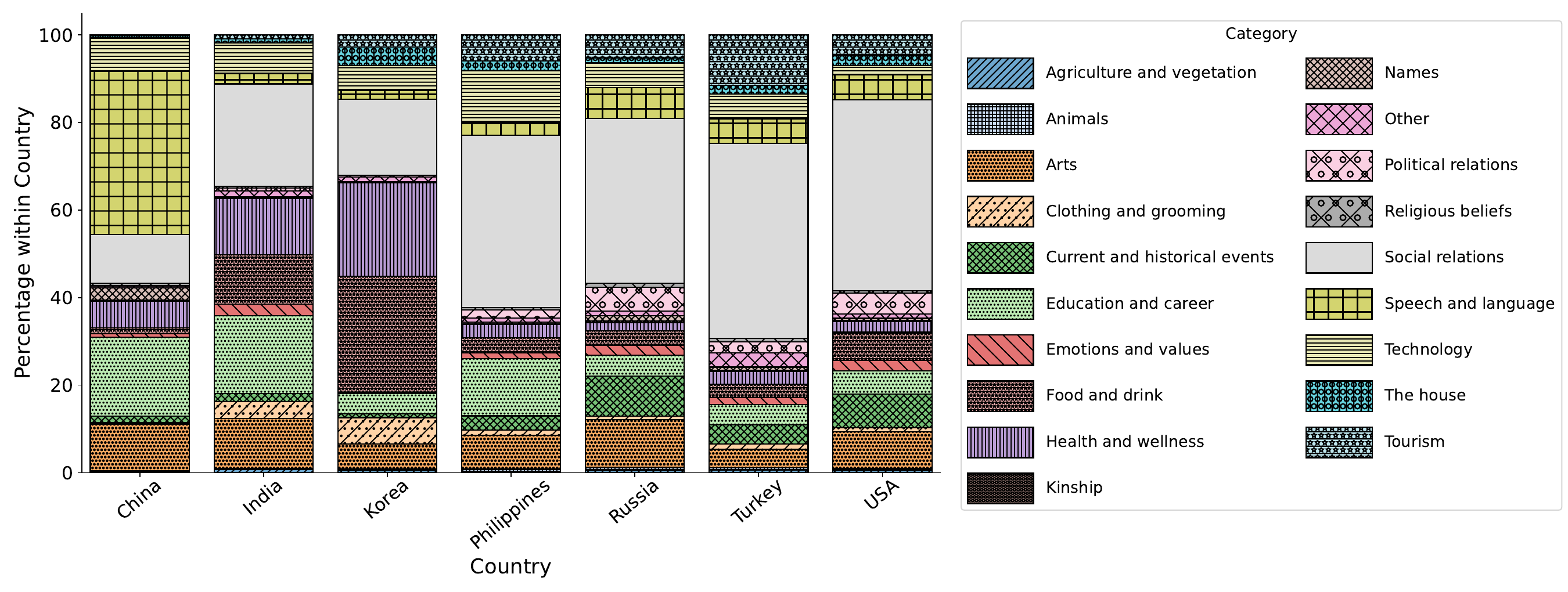}
    \caption{Percentage of question category in \dataset\ mapped to respective subreddit countries.}
    \label{fig:dataset_category}
\end{figure*}

Following large-scale annotation of $\approx$376k posts, a further stratified sample of 250 question–category pairs is evaluated by two external annotators (same as LLM-judge). Annotator 1 and Annotator 2 mark 87.6\% and 90.8\% of LLM-assigned labels as correct, achieving a Cohen's $\kappa$ of 0.59 (agreement: 96\%). The results reflect the inherently fuzzy nature of this categorization task. We freeze the category at this point and manually clean unique labels from large-scale annotations, mapping synonyms and spelling errors to the 19 categories. The process of taxonomy creation is outlined in Figure \ref{fig:taxonomy}. 
As a result of taxonomy creation and annotation, we obtain a category mapping per country, as shown in Figure \ref{fig:dataset_category}.
  
\paragraph{Questions categories} The complete list of categories, along with a one-line description provided during zero-shot LLM annotation: 
\begin{enumerate}[noitemsep]
\item \textbf{Agriculture and vegetation: } Farming, plants, cultivation, botanical elements, horticulture.
 \item \textbf{Animals: } Fauna, wildlife, domesticated animals, animal behaviors.
 \item \textbf{Arts: } Artistic expression, music, dance, cultural performances, entertainment, creative practices, literature.
 \item \textbf{Clothing and grooming: } Attire, fashion, personal care items, grooming practices.
 \item \textbf{Current and historical events: } Knowledge about historical events, current news.
  \item \textbf{Education and career: } Schooling, education system, jobs, career paths.
  \item \textbf{Emotions and values: } Feelings, sentiments, Cultural values (e.g., collectivism and individualism), work ethic, modesty.
    \item \textbf{Food and drinks: } Food items, beverages, cooking methods, culinary practices.
  \item \textbf{Health and wellness: } Traditional and modern health practices, public health issues, well-being, and health infrastructure.
  \item \textbf{Kinship: } Family relationships, lineage, familial connections.
  \item \textbf{Names: } Personal names, place names, identification systems.
  \item \textbf{Political relations: } Political systems, trade policies, economic development.
 \item \textbf{Religious beliefs: } Religious entities and practices.
  \item \textbf{Social relations: } Social order, interpersonal dynamics, communication practices.
 \item \textbf{Speech and language: } Verbal expression, linguistic practices, grammar knowledge, rhetorical structures.
 \item \textbf{Technology: } Technological advancements, adaptation, digital innovation.
 \item \textbf{The house: }Dwellings, domestic spaces, furniture, household items.
  \item \textbf{Tourism: } Traveling, tourist attraction, safety measures, travel trips,  climatic conditions.
\item \textbf{Other: } If the question is not related to any of the categories, or the question does not belong to any cultural category, classify it as ``Other''.
\end{enumerate}

\paragraph{Prompt} The prompt for post categorization is listed in Figure~\ref{fig:prompt-category}.

\section{Experimental Setup and Compute}
\label{app:exp}

For all experiments, we use the vLLM library for efficient inference~\citep{kwon2023efficient}. We evaluate three instruction-tuned open-weight models for response generation: \textbf{Llama-3.1-8B-Instruct} (Llama)~\citep{grattafiori2024llama}, \textbf{Gemma-3-12B-IT} (Gemma)~\citep{team2025gemma}, and \textbf{Ministral-3-14B-Instruct-2512} ~\citep{liu2026ministral}. These models span three families and parameter scales (8B, 12B, and 14B). For judge-based evaluation across all four \framework\ characteristics, we again use \textbf{Llama-3.1-8B-Instruct}. The corresponding Hugging Face repository codes are listed in \autoref{tab:model_codes}.

Unless otherwise stated, we use the following vLLM sampling configuration: \texttt{temperature=0.7}, \texttt{top\_p=0.95}, and \texttt{max\_tokens=512}. We generate one sample per prompt and run all models in \texttt{bfloat16} precision.

\textbf{Llama}: We use \texttt{meta-llama/Llama-3.1-8B-Instruct}, available via Hugging Face\footnote{\url{https://huggingface.co/meta-llama/Llama-3.1-8B-Instruct}}.

\textbf{Gemma}: We use \texttt{google/gemma-3-12b-it}, available via Hugging Face\footnote{\url{https://huggingface.co/google/gemma-3-12b-it}}.

\textbf{Ministral}: We use \texttt{mistralai/Ministral-3-14B-Instruct-2512}, available via Hugging Face\footnote{\url{https://huggingface.co/mistralai/Ministral-3-14B-Instruct-2512}}.

\textbf{Judges}: For LLM-as-a-judge across the four characteristics, we use \texttt{meta-llama/Llama-3.1-8B-Instruct}\footnote{\url{https://huggingface.co/meta-llama/Llama-3.1-8B-Instruct}} with the same decoding configuration as above. For response generation, we produce one response per (post, model) pair, yielding approximately $376k$ responses per model and $\sim1.1M$ generations across the three generation models. This stage required $\approx$1 day on 8 NVIDIA A100s. 

For judge-based analysis, we run \texttt{Llama-3.1-8B-Instruct} on 3 \framework\ judges: cultural positioning, anthropomorphism, and maxims.  The anthropomorphism judge is invoked once per retained cue, and the maxims judge once per maxim, yielding $\sim 16$M judge calls in total. Following~\citet{gorge2025detecting}\footnote{\url{https://huggingface.co/meta-llama/Llama-3.1-70B-Instruct}}, we run \texttt{Llama-3.1-70B-Instruct} to examine the use of generalizing language. This stage required approximately 6 days on 8 NVIDIA A100s.

The Reddit corpus we curate is sourced from the publicly released Pushshift crawl \citep{baumgartner2020pushshift}; for the open-weight models, we accepted the corresponding Hugging Face usage terms, which permit their use for research, generation, and analysis in academic publications. We used the code- and writing-assistant Claude~\citep{anthropic2025claudesonnet45systemcard} to help with code and writing feedback; generated code was manually tested and verified before running.

The annotated Reddit data, along with the framework and analysis code, will be made publicly available under the MIT license for research purposes.

\begin{table*}[h]
   \centering
   \small
   \begin{tabular}{l l l}
   \toprule
       \textbf{Role} & \textbf{Model} & \textbf{Hugging Face repository} \\
   \midrule
       Generation & Llama & \texttt{meta-llama/Llama-3.1-8B-Instruct} \\
       Generation & Gemma & \texttt{google/gemma-3-12b-it} \\
       Generation & Ministral & \texttt{mistralai/Ministral-3-14B-Instruct-2512} \\
       Judge & Llama & \texttt{meta-llama/Llama-3.1-8B-Instruct} \\
       Judge (Gen.) & Llama & \texttt{meta-llama/Llama-3.1-70B-Instruct} \\
   \bottomrule
   \end{tabular}
   \caption{Models used in this study and their corresponding Hugging Face repository codes.}
   \label{tab:model_codes}
\end{table*}

 \begin{table*}[!t]
\centering
\resizebox{0.85\textwidth}{!}{%
\begin{tabular}{llllll}
\toprule
\textbf{Country} & \textbf{Category} & \textbf{Quality} & \textbf{Quantity} & \textbf{Relation} & \textbf{Manner} \\
\midrule
\textbf{India} & social relations & G$>$L$>$M & G$>$M$>$L & M$>$G$>$L & G$>$L$>$M \\
 & education and career & G$>$L$>$M & G$>$M$>$L & G$>$M$>$L & G$>$L$\sim$M \\
 & health and wellness & G$>$L$>$M & G$>$M$>$L & M$>$G$>$L & L$>$M$>$G \\
\midrule
\textbf{China} & speech and language & G$>$M$>$L & G$>$M$>$L & G$>$M$>$L & G$>$M$>$L \\
 & education and career & G$>$L$>$M & G$>$M$>$L & G$>$M$>$L & G$>$M$>$L \\
 & social relations & G$>$L$\sim$M & G$>$M$\sim$L & G$>$M$>$L & G$>$L$\sim$M \\
\midrule
\textbf{Philippines} & social relations & G$>$L$\sim$M & G$>$M$>$L & G$>$M$>$L & G$>$L$>$M \\
 & education and career & G$>$L$>$M & G$>$M$>$L & G$>$M$>$L & G$>$M$>$L \\
 & technology & G$>$L$>$M & G$>$M$\sim$L & M$>$G$>$L & G$>$M$>$L \\
\midrule
\textbf{USA} & social relations & G$>$L$>$M & G$>$M$>$L & G$>$M$>$L & G$>$L$>$M \\
 & arts & L$\sim$G$\sim$M & G$>$M$\sim$L & G$>$M$>$L & G$>$L$>$M \\
 & current and historical events & G$>$L$>$M & G$>$M$\sim$L & G$>$M$>$L & G$>$M$\sim$L \\
 \midrule
\textbf{Korea} & food and drink & G$\sim$L$>$M & G$>$L$\sim$M & G$>$M$>$L & G$>$L$\sim$M \\
 & health and wellness & G$>$L$>$M & G$>$M$>$L & M$>$G$>$L & L$>$G$>$M \\
 & social relations & G$>$L$>$M & G$>$M$>$L & M$>$G$>$L & G$>$M$\sim$L \\
\midrule
\textbf{Russia} & social relations & G$>$L$>$M & G$>$M$>$L & G$>$M$>$L & G$>$L$\sim$M \\
 & arts & G$>$L$>$M & G$>$M$\sim$L & G$>$M$>$L & G$>$L$\sim$M \\
 & current and historical events & G$>$L$\sim$M & G$\sim$M$\sim$L & G$>$M$>$L & G$>$M$\sim$L \\
\midrule
\textbf{Turkey} & social relations & G$>$L$>$M & G$>$M$>$L & G$>$M$>$L & G$>$L$\sim$M \\
 & tourism & G$>$L$>$M & G$>$L$\sim$M & G$>$M$>$L & G$>$M$\sim$L \\
 & technology & G$\sim$L$\sim$M & G$\sim$M$\sim$L & G$>$M$>$L & G$>$M$\sim$L \\
\bottomrule
\end{tabular}}
\caption{Model rankings by adherence to individual maxims. $>$ significantly different (FDR-corrected, $p<.05$); $\sim$ not significantly different. L=\texttt{Llama-3.1-8B-it}, G=\texttt{Gemma-3-12b-it}, and M=\texttt{Ministral-3B-14B-it}. Countries are listed by total count across 19 categories (Section~\ref{sec:rq3_result}).}
\label{tab:model_rankings_maxim}
\end{table*}

\section{Statistical Testing}
\label{app:stat_test}
Since the responses from the three LLMs are generated for the same input post, and some country-category combinations are more likely to trigger certain response characteristics, treating each post-response pair as fully independent violates the independence assumption of regression. In light of this setup, it is important to note that GEE offers a principled approach compared with correlation or regression analyses, controlling for binary outcomes, clustered observations, and within- and cross-LLM differences. 

All GEE models group observations by post ID, forming clusters of size 3 (for the 3 LLMs we examine), and are fit separately for each country-category subset for each RQ. Each GEE estimates the probability that a given LLM's response contains the characteristic of interest, averaged over all posts in that subset. We adopt a Binomial family (assuming the characteristic is either present or absent), operate on the log-odds, and convert the results to probabilities for analysis and visualization. Llama serves as the reference LLM throughout.

For robustness, we disregard any fitted model in which a parameter estimate or $p$-value is undefined, which can occur when a country-category subset contains too few observations for reliable estimation. For the per-country plots in RQ2 and RQ3, we restrict to categories for which all GEEs converge and pass the checks. Thus, each plotted category is comparable across all panels. The summary heatmaps display results for each cause--effect pair, regardless.

\section{Additional Results}
\label{app:add_results}
\begin{figure*}[!t]
    \centering
    \includegraphics[width=0.85\textwidth]{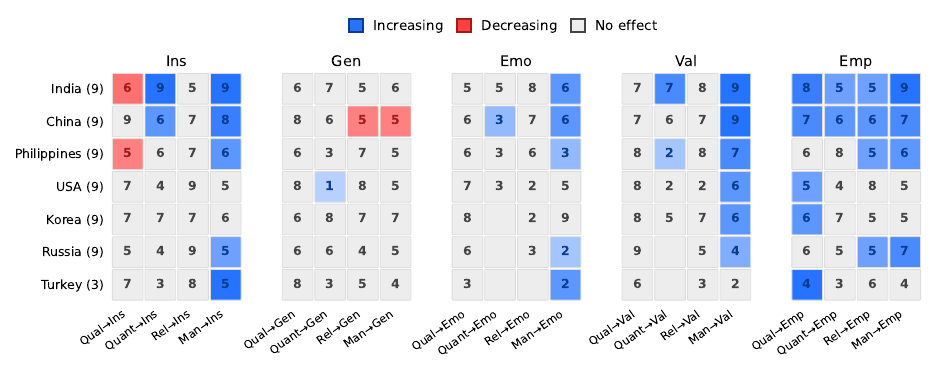}
    \caption{\textbf{C}$\rightarrow$\textbf{E} co-occurrence (Eq~\ref{eq:delta}). Each cell reports \# combinations across top-categories and LLMs ($3*3$) per country, where the presence of a \textbf{c}ause significantly increases the probability of \textbf{e}ffect (blue), decreases it (red), or has no impact ($\Delta \approx 0$, grey). Here, Ins = insider positioning, Gen = generalizing language. Anthropomorphic cues include Emo = emotion, Val = validation, and Emp = empathy. The maxims are Qua= Quanlity, Quant=Quantity, Rel=Relevance/Relation, Man=Manner. Countries are listed by total count across 19 categories (Section~\ref{sec:rq3_result}).}
    \label{fig:maxim2cue}
\end{figure*}

Table \ref{tab:model_rankings_maxim} captures the relative ranking among LLMs with respect to adherence to each maxim. Figure \ref{fig:maxim2cue} captures the impact of different maxims on other response characteristics, summarised at the per-country level.

\begin{figure*}[t]
    \centering
    \begin{subfigure}[t]{0.75\textwidth}
        \includegraphics[width=\linewidth]{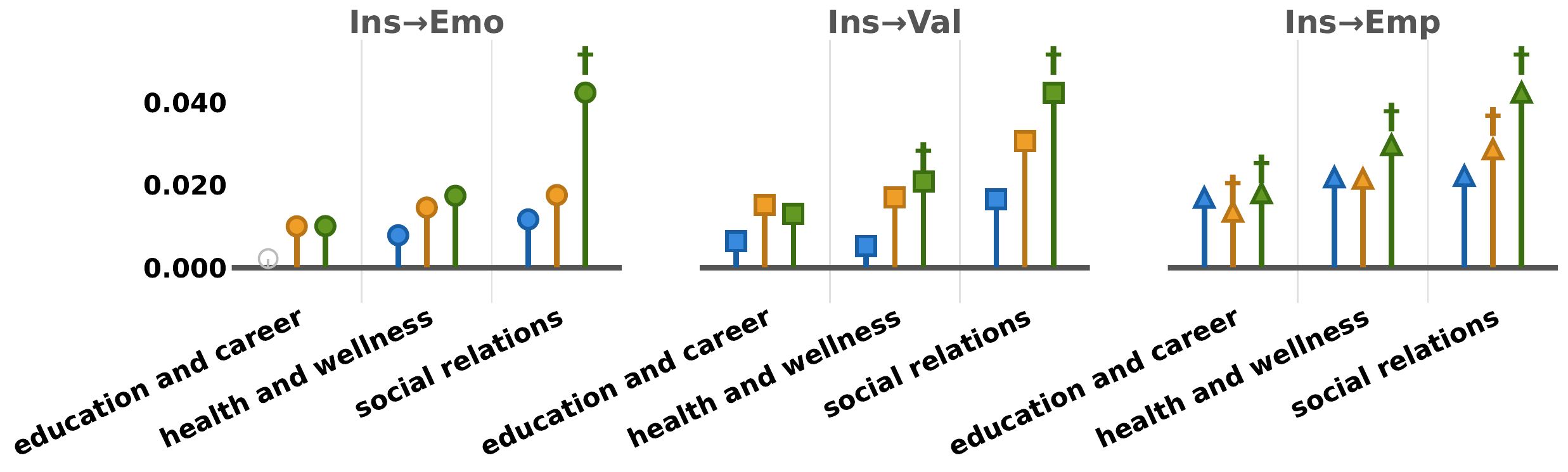}
        \caption{Ins $\rightarrow$ Anthro}
    \end{subfigure}
    \hfill
    \begin{subfigure}[t]{0.75\textwidth}
        \includegraphics[width=\linewidth]{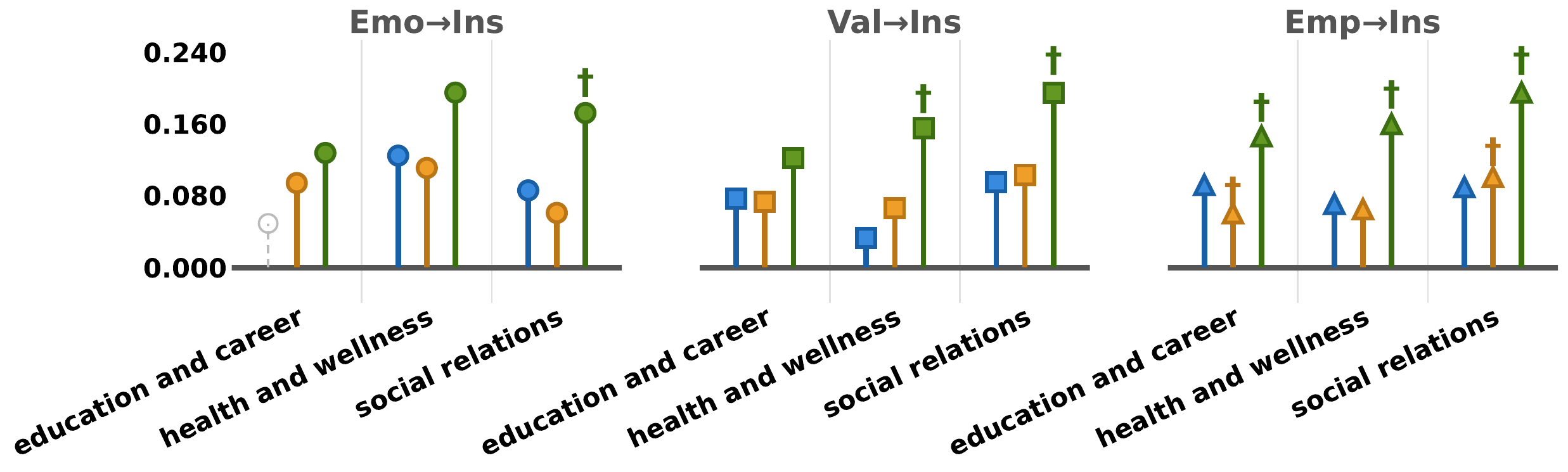}
        \caption{Anthro $\rightarrow$ Ins}
    \end{subfigure}
    \hfill
    \begin{subfigure}[t]{0.75\textwidth}
        \includegraphics[width=\linewidth]{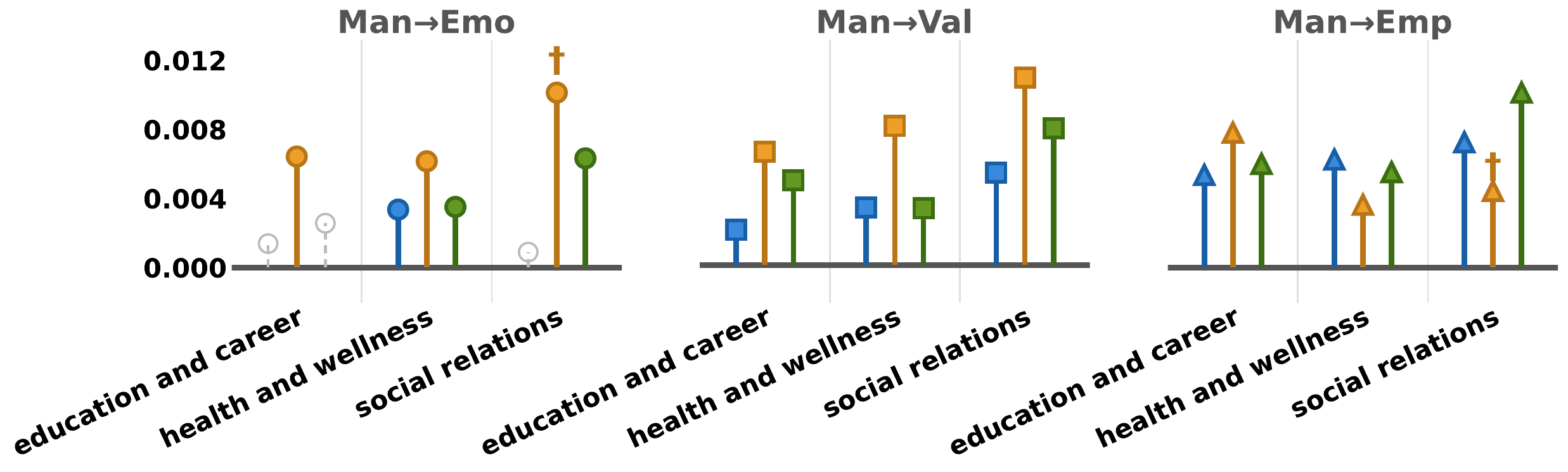}
        \caption{Manner $\rightarrow$ Anthro}
    \end{subfigure}
    \hfill
    \begin{subfigure}[t]{0.65\textwidth}
        \includegraphics[width=\linewidth]{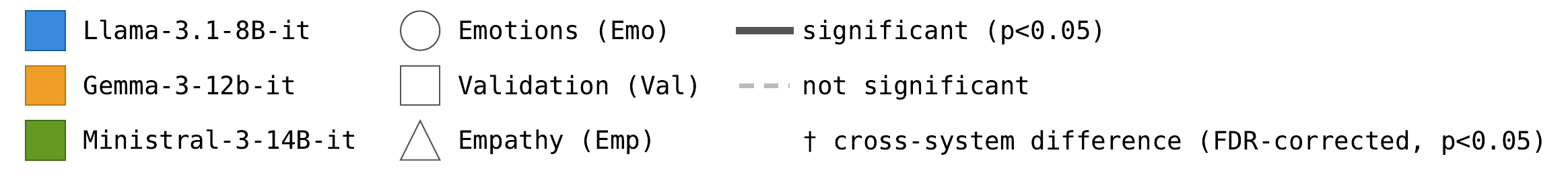}
    \end{subfigure}
    \caption{\textbf{India:} Computed via Eq.~\ref{eq:delta}, the estimated co-occurrence effects of: (a) cultural positioning on anthropomorphism, (b) anthropomorphism on cultural positioning, and (c) maxim of manner on anthropomorphism. Each panel shows the change in probability across models for the top-3 categories. Solid lines indicate significant Wald effects, $\dagger$ denotes cross-model differences FDR-corrected, both at p $<$ 0.05 (Sections~\ref{sec:rq2_result} and ~\ref{sec:rq3_result}).}
    \label{fig:in_3block}
\end{figure*}

\begin{figure*}[t]
    \centering
    \begin{subfigure}[t]{0.75\textwidth}
        \includegraphics[width=\linewidth]{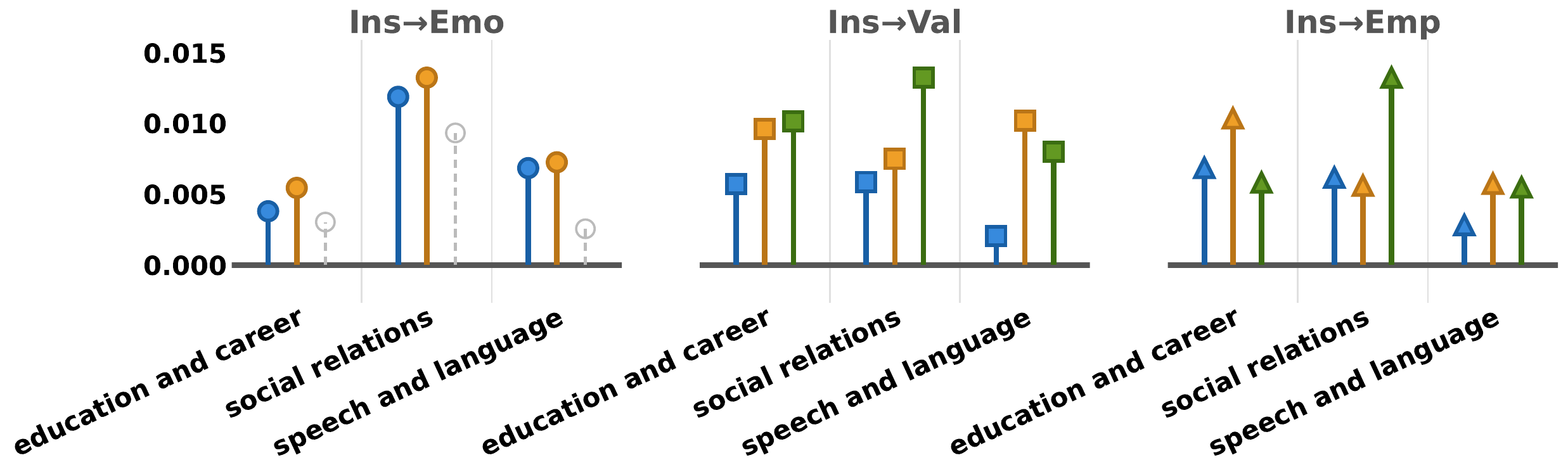}
        \caption{Ins $\rightarrow$ Anthro}
    \end{subfigure}
    \hfill
    \begin{subfigure}[t]{0.75\textwidth}
        \includegraphics[width=\linewidth]{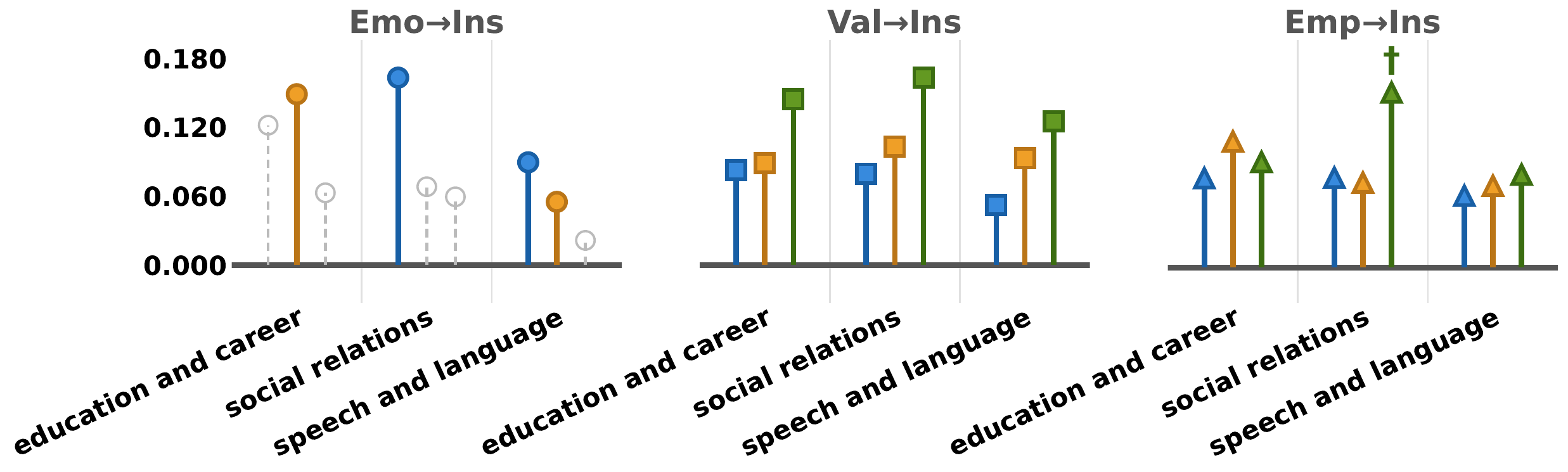}
        \caption{Anthro $\rightarrow$ Ins}
    \end{subfigure}
    \hfill
    \begin{subfigure}[t]{0.75\textwidth}
        \includegraphics[width=\linewidth]{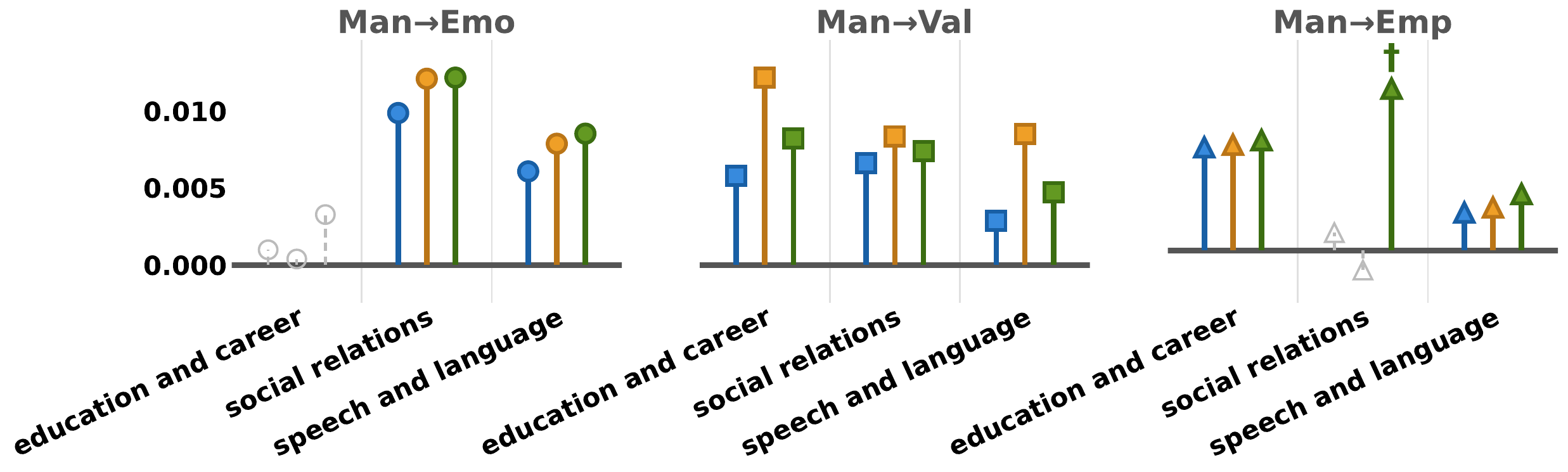}
        \caption{Manner $\rightarrow$ Anthro}
    \end{subfigure}
    \hfill
    \begin{subfigure}[t]{0.65\textwidth}
        \includegraphics[width=\linewidth]{figs/legend.pdf}
    \end{subfigure}
    \caption{\textbf{China:} Computed via Eq.~\ref{eq:delta}, the estimated co-occurrence effects of: (a) cultural positioning on anthropomorphism, (b) anthropomorphism on cultural positioning, and (c) maxim of manner on anthropomorphism. Each panel shows the change in probability ($\Delta$) across models for the top-3 categories. Solid lines indicate significant Wald effects, $\dagger$ denotes cross-model differences FDR-corrected, both at p $<$ 0.05 (Sections~\ref{sec:rq2_result} and ~\ref{sec:rq3_result}).}
    \label{fig:ch_3block}
\end{figure*}

\begin{figure*}[t]
    \centering
    \begin{subfigure}[t]{0.75\textwidth}
        \includegraphics[width=\linewidth]{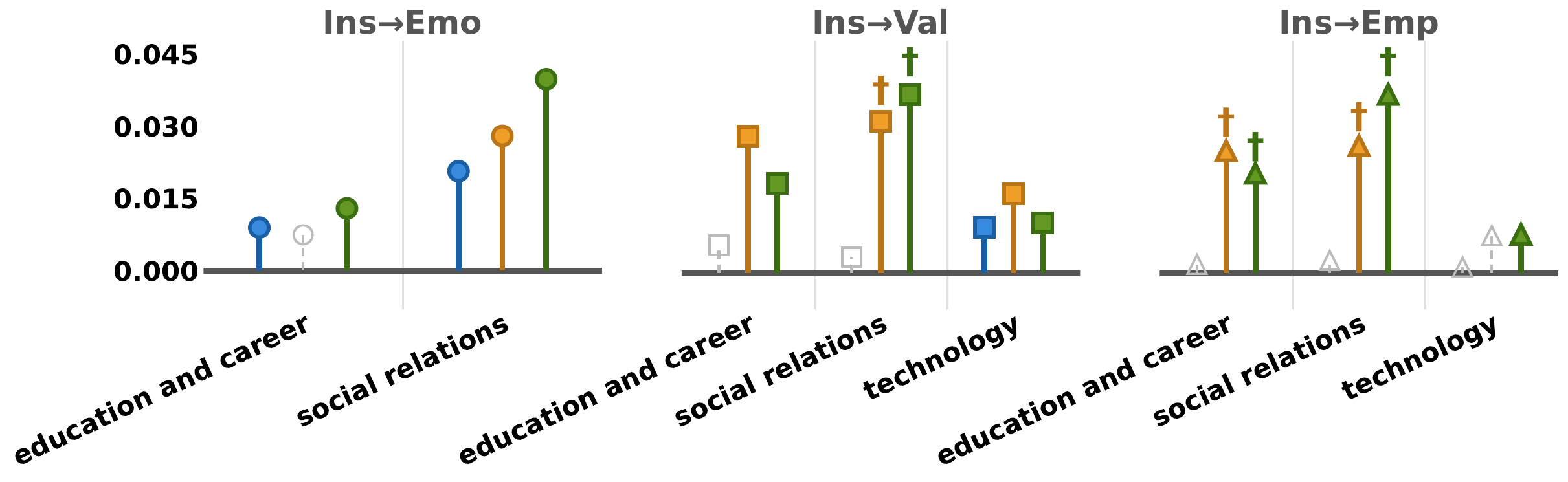}
        \caption{Ins $\rightarrow$ Anthro}
    \end{subfigure}
    \hfill
    \begin{subfigure}[t]{0.75\textwidth}
        \includegraphics[width=\linewidth]{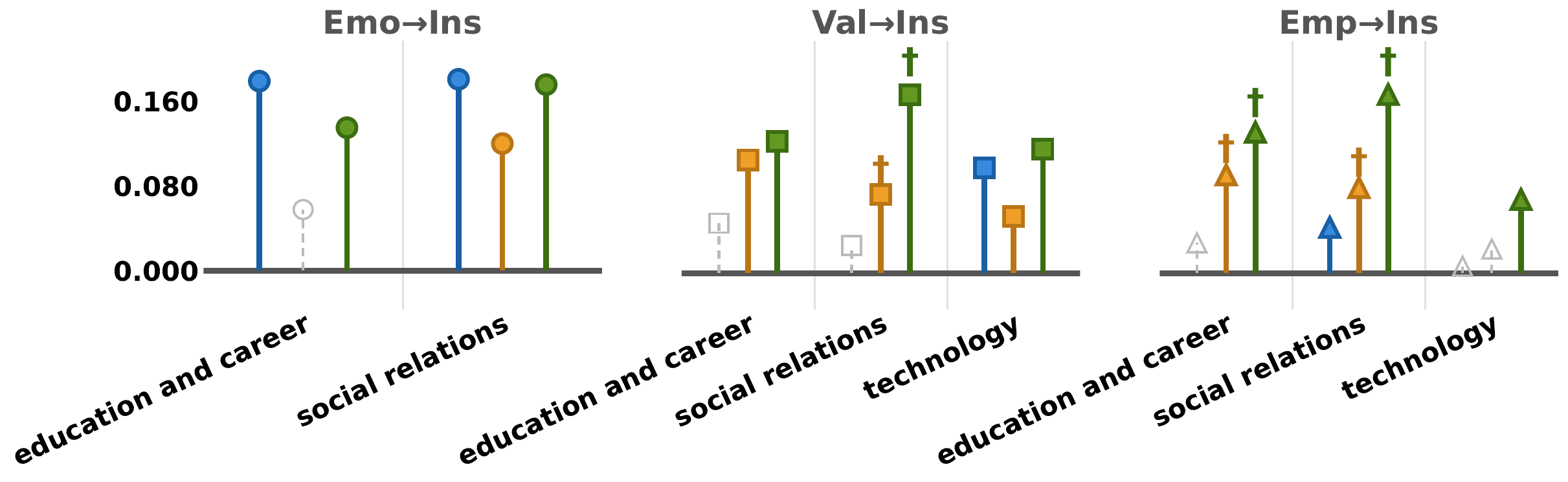}
        \caption{Anthro $\rightarrow$ Ins}
    \end{subfigure}
    \hfill
    \begin{subfigure}[t]{0.75\textwidth}
        \includegraphics[width=\linewidth]{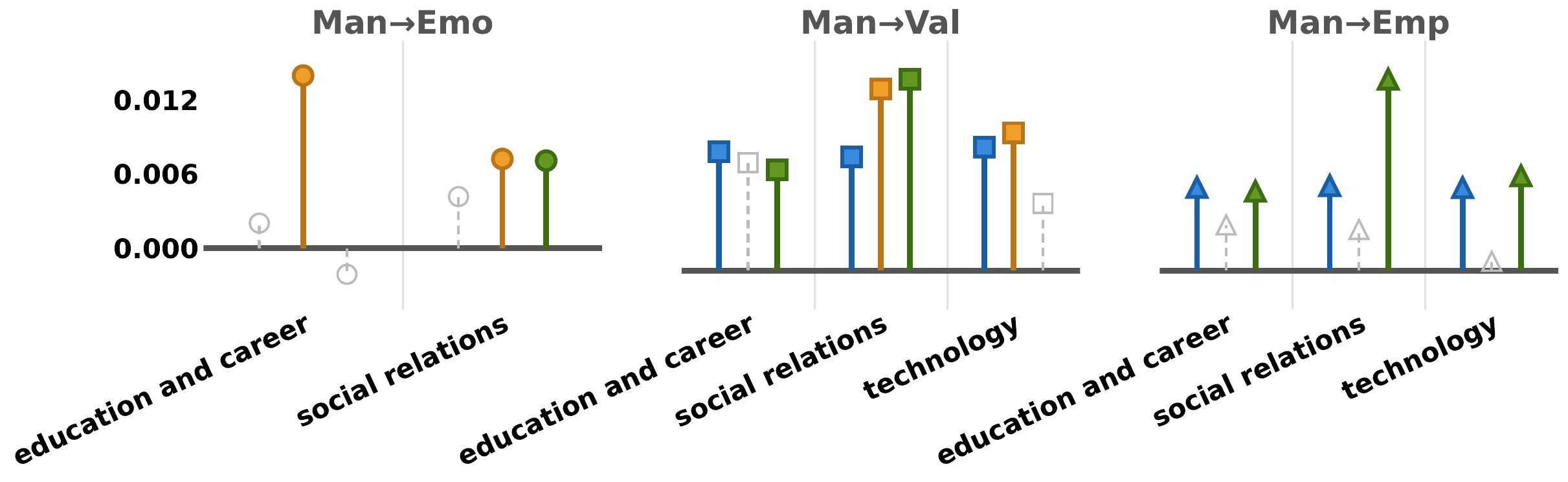}
        \caption{Manner $\rightarrow$ Anthro}
    \end{subfigure}
    \hfill
    \begin{subfigure}[t]{0.65\textwidth}
        \includegraphics[width=\linewidth]{figs/legend.pdf}
    \end{subfigure}
        \caption{\textbf{Philippines:} Computed via Eq.~\ref{eq:delta}, the estimated co-occurrence effects of: (a) cultural positioning on anthropomorphism, (b) anthropomorphism on cultural positioning, and (c) maxim of manner on anthropomorphism. Each panel shows the change in probability ($\Delta$) across models for the top-3 categories. Solid lines indicate significant Wald effects, $\dagger$ denotes cross-model differences FDR-corrected, both at p $<$ 0.05 (Sections~\ref{sec:rq2_result} and ~\ref{sec:rq3_result}).}
    \label{fig:ph_3block}
\end{figure*}

\begin{figure*}[t]
    \centering
    \begin{subfigure}[t]{0.75\textwidth}
        \includegraphics[width=\linewidth]{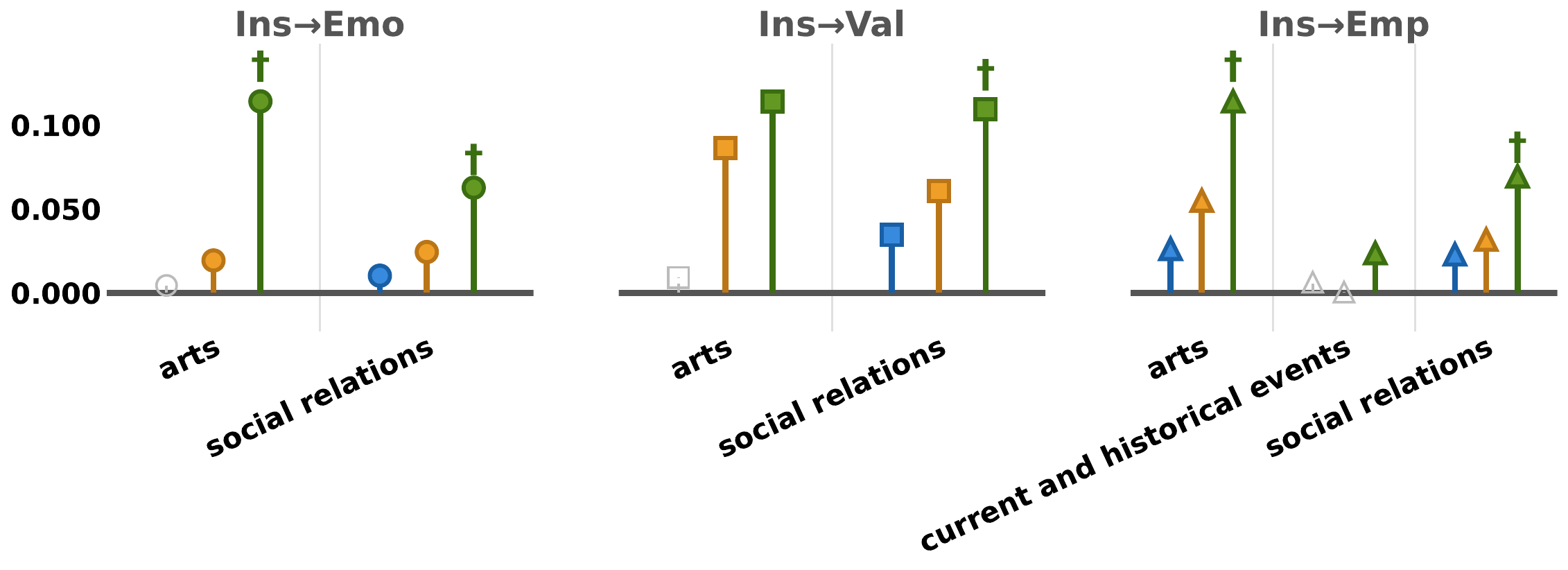}
        \caption{Ins $\rightarrow$ Anthro}
    \end{subfigure}
    \hfill
    \begin{subfigure}[t]{0.75\textwidth}
        \includegraphics[width=\linewidth]{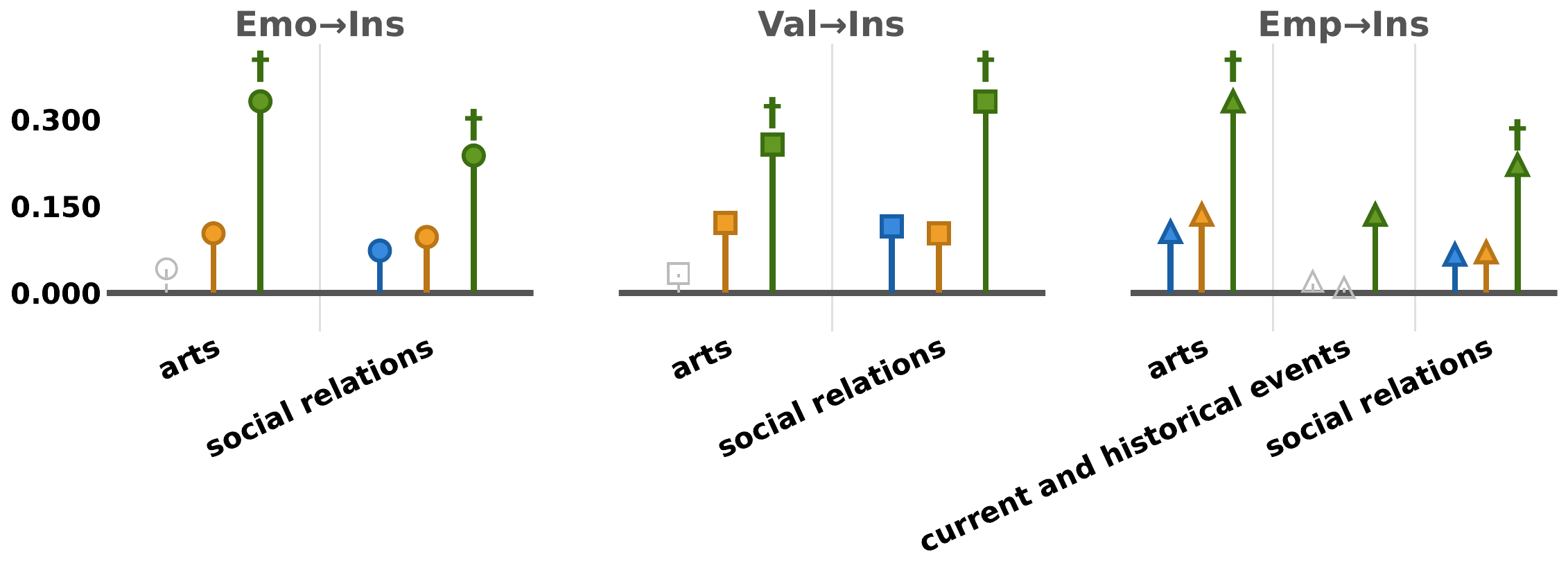}
        \caption{Anthro $\rightarrow$ Ins}
    \end{subfigure}
    \hfill
    \begin{subfigure}[t]{0.75\textwidth}
        \includegraphics[width=\linewidth]{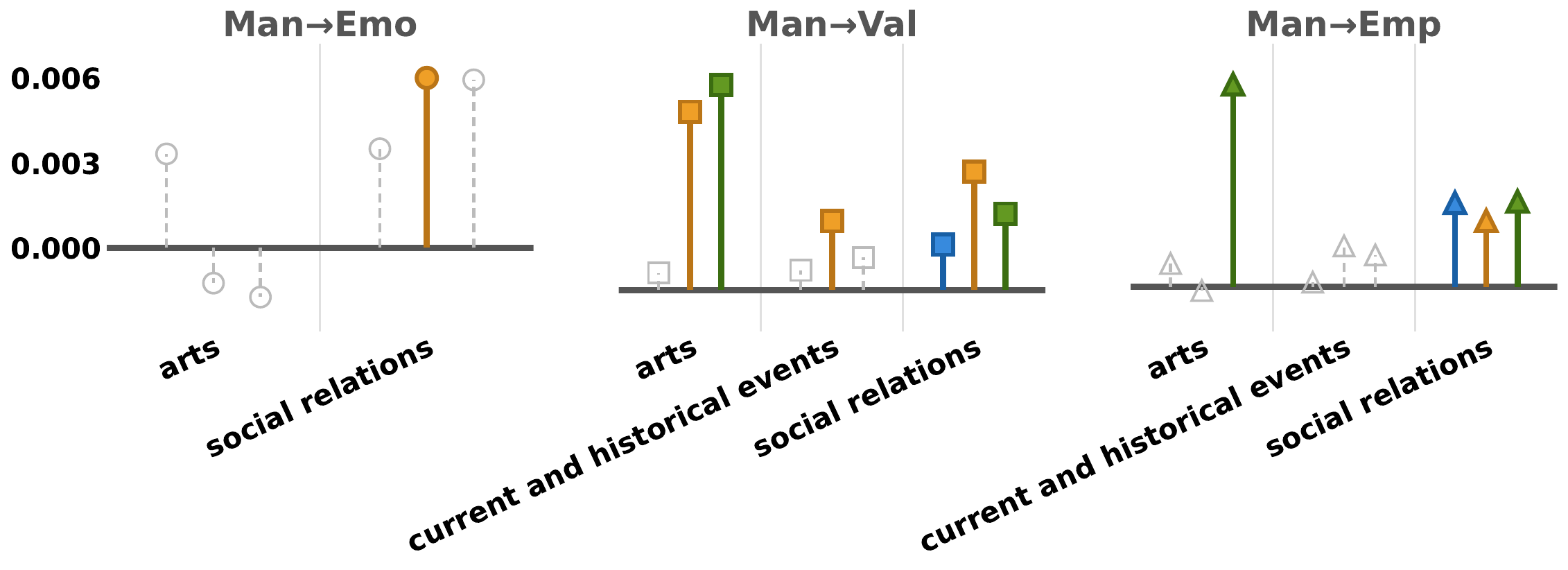}
        \caption{Manner $\rightarrow$ Anthro}
    \end{subfigure}
    \hfill
    \begin{subfigure}[t]{0.65\textwidth}
        \includegraphics[width=\linewidth]{figs/legend.pdf}
    \end{subfigure}
        \caption{\textbf{USA:} Computed via Eq.~\ref{eq:delta}, the estimated co-occurrence effects of: (a) cultural positioning on anthropomorphism, (b) anthropomorphism on cultural positioning, and (c) maxim of manner on anthropomorphism. Each panel shows the change in probability ($\Delta$) across models for the top-3 categories. Solid lines indicate significant Wald effects, $\dagger$ denotes cross-model differences FDR-corrected, both at p $<$ 0.05 (Sections~\ref{sec:rq2_result} and ~\ref{sec:rq3_result}).}
    \label{fig:us_3block}
\end{figure*}

\begin{figure*}[t]
    \centering
    \begin{subfigure}[t]{0.75\textwidth}
        \includegraphics[width=\linewidth]{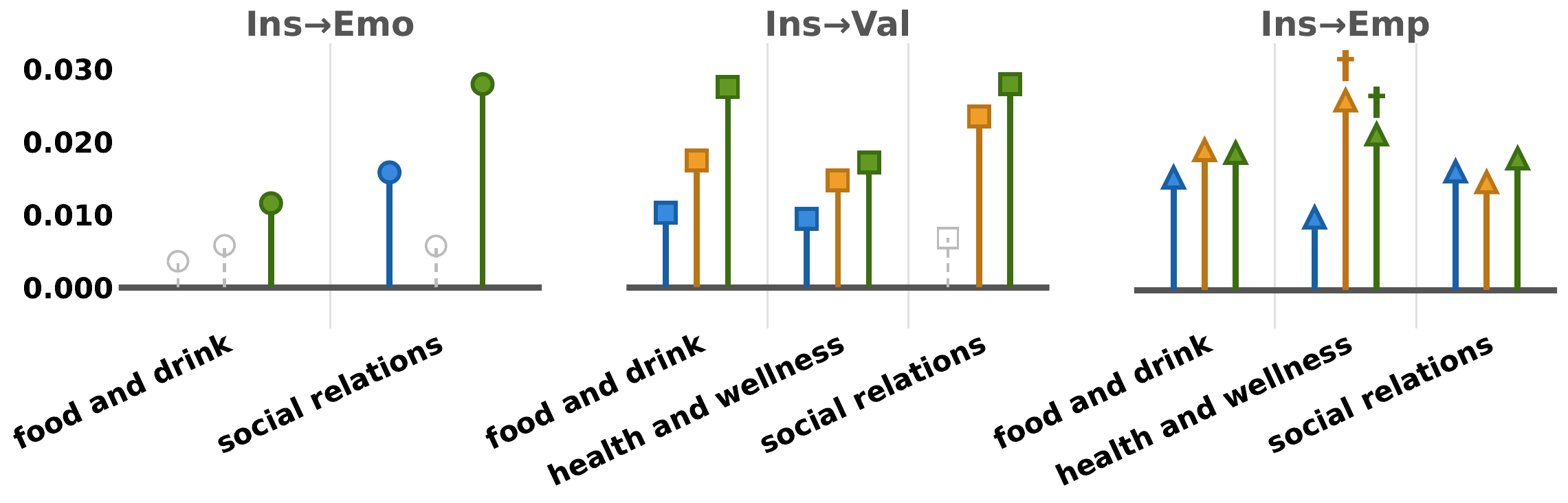}
        \caption{Ins $\rightarrow$ Anthro}
    \end{subfigure}
    \hfill
    \begin{subfigure}[t]{0.75\textwidth}
        \includegraphics[width=\linewidth]{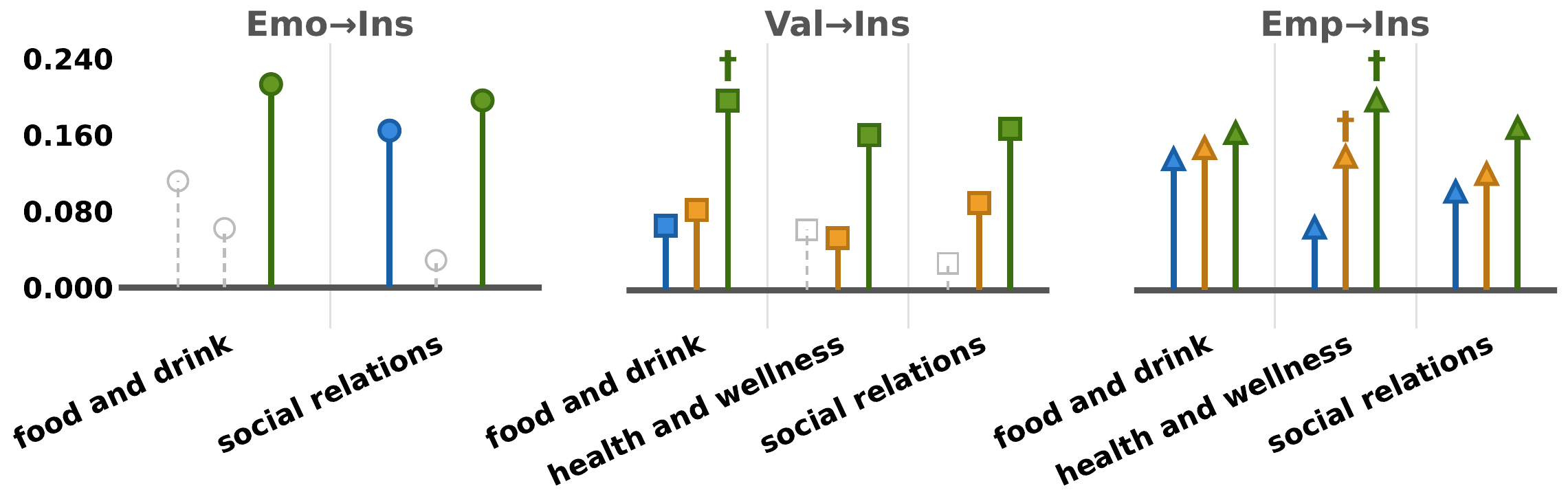}
        \caption{Anthro $\rightarrow$ Ins}
    \end{subfigure}
    \hfill
    \begin{subfigure}[t]{0.75\textwidth}
        \includegraphics[width=\linewidth]{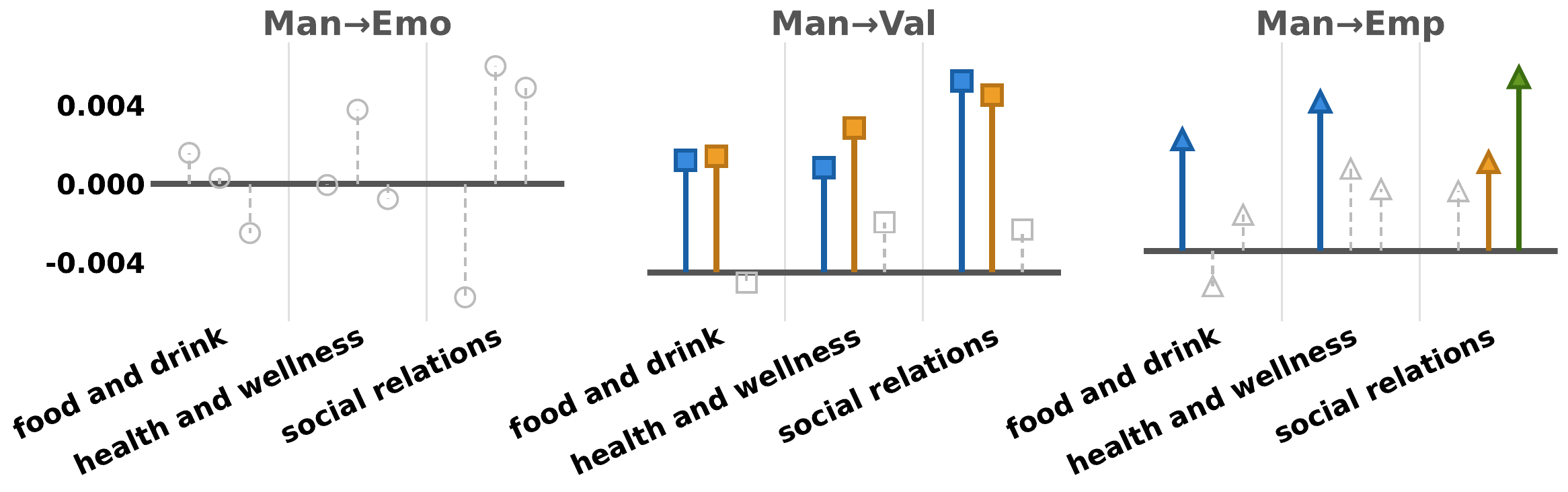}
        \caption{Manner $\rightarrow$ Anthro}
    \end{subfigure}
    \hfill
    \begin{subfigure}[t]{0.65\textwidth}
        \includegraphics[width=\linewidth]{figs/legend.pdf}
    \end{subfigure}
        \caption{\textbf{Korea:} Computed via Eq.~\ref{eq:delta}, the estimated co-occurrence effects of: (a) cultural positioning on anthropomorphism, (b) anthropomorphism on cultural positioning, and (c) maxim of manner on anthropomorphism. Each panel shows the change in probability ($\Delta$) across models for the top-3 categories. Solid lines indicate significant Wald effects, $\dagger$ denotes cross-model differences FDR-corrected, both at p $<$ 0.05 (Sections~\ref{sec:rq2_result} and ~\ref{sec:rq3_result}).}
    \label{fig:ko_3block}
\end{figure*}

\begin{figure*}[t]
    \centering
    \begin{subfigure}[t]{0.75\textwidth}
        \includegraphics[width=\linewidth]{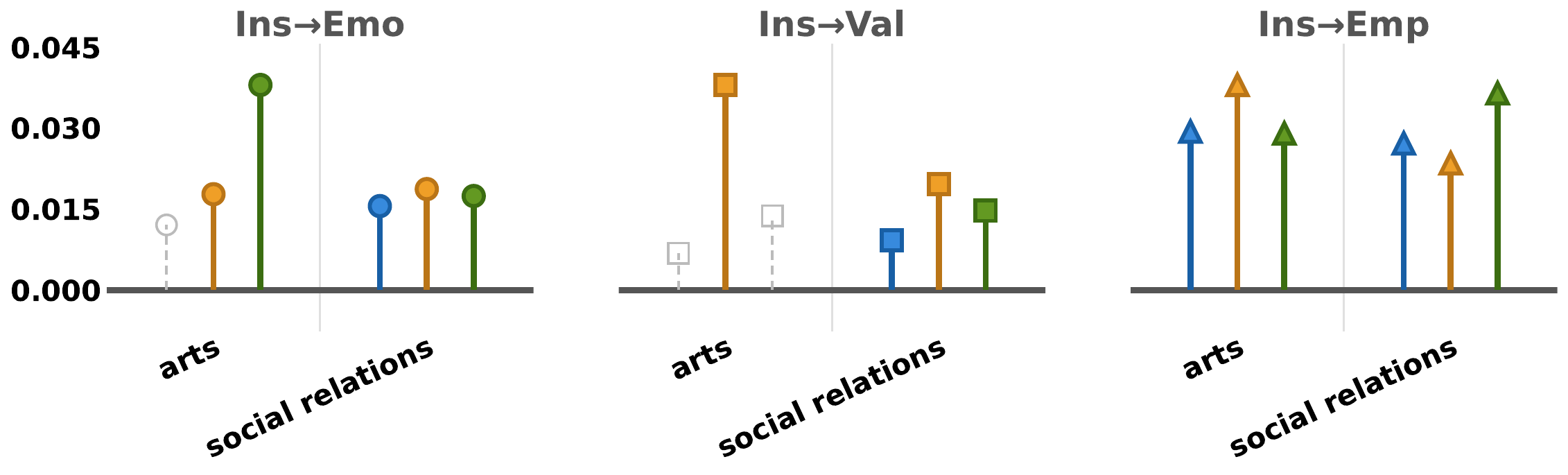}
        \caption{Ins $\rightarrow$ Anthro}
    \end{subfigure}
    \hfill
    \begin{subfigure}[t]{0.75\textwidth}
        \includegraphics[width=\linewidth]{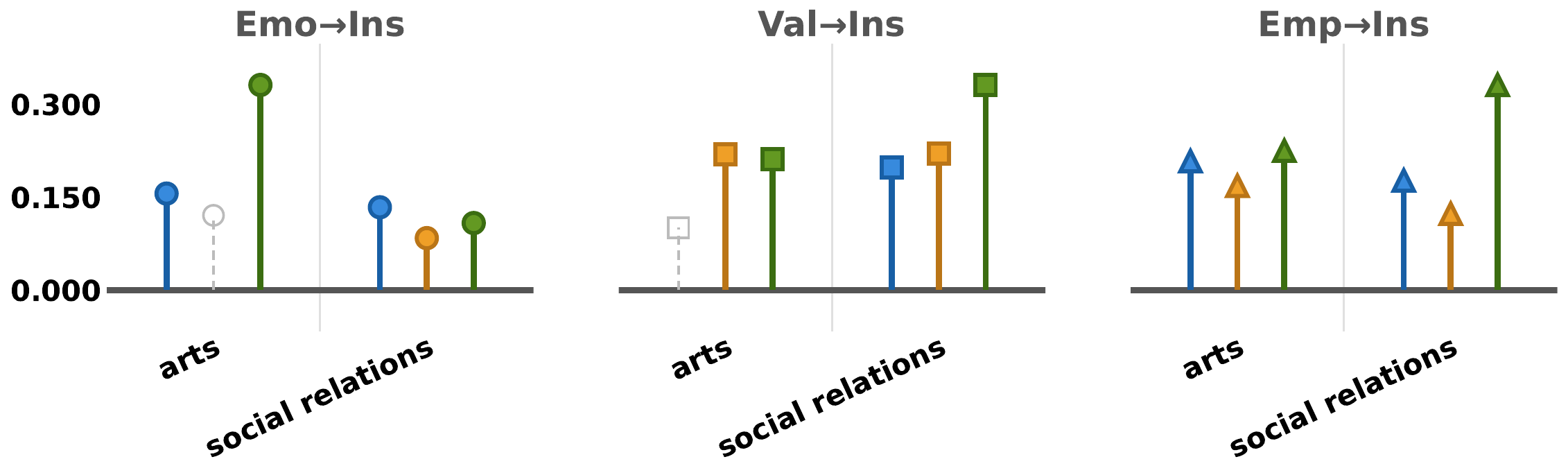}
        \caption{Anthro $\rightarrow$ Ins}
    \end{subfigure}
    \hfill
    \begin{subfigure}[t]{0.75\textwidth}
        \includegraphics[width=\linewidth]{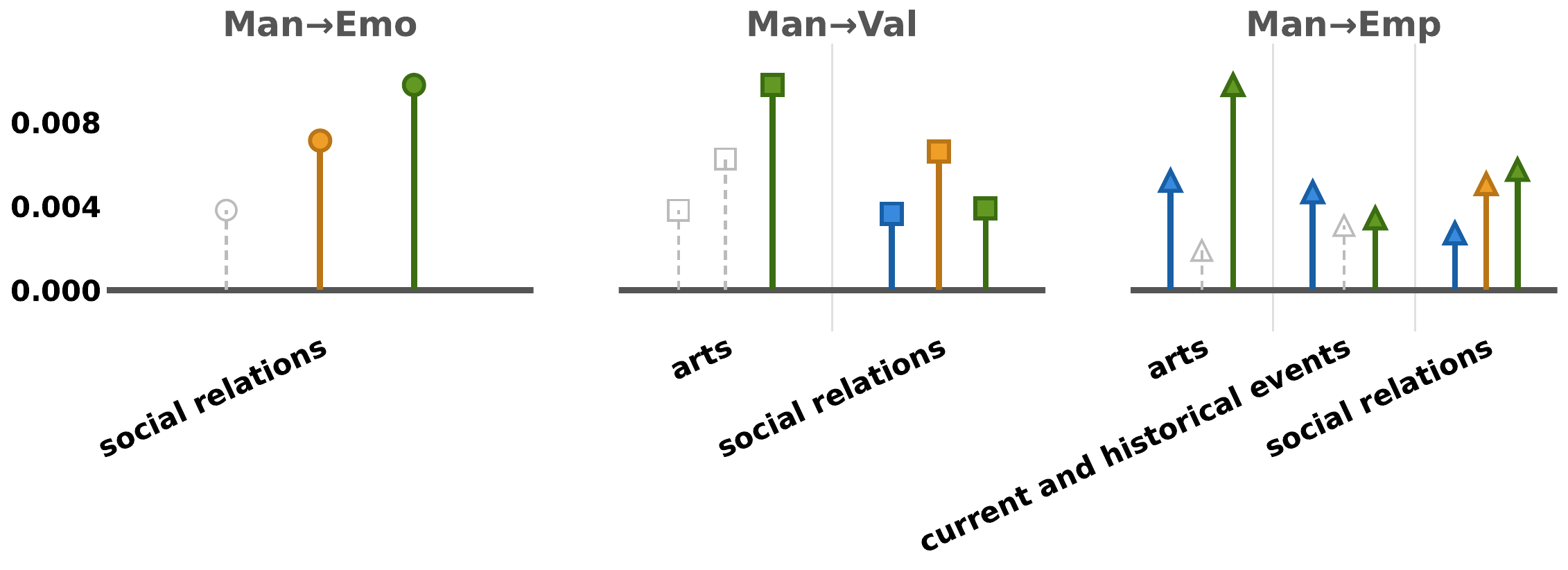}
        \caption{Manner $\rightarrow$ Anthro}
    \end{subfigure}
    \hfill
    \begin{subfigure}[t]{0.65\textwidth}
        \includegraphics[width=\linewidth]{figs/legend.pdf}
    \end{subfigure}
        \caption{\textbf{Russia:} Computed via Eq.~\ref{eq:delta}, the estimated co-occurrence effects of: (a) cultural positioning on anthropomorphism, (b) anthropomorphism on cultural positioning, and (c) maxim of manner on anthropomorphism. Each panel shows the change in probability ($\Delta$) across models for the top-3 categories. Solid lines indicate significant Wald effects, $\dagger$ denotes cross-model differences FDR-corrected, both at p $<$ 0.05 (Sections~\ref{sec:rq2_result} and ~\ref{sec:rq3_result}).}
    \label{fig:ru_3block}
\end{figure*}

\begin{figure*}[t]
    \centering
    \begin{subfigure}[t]{0.75\textwidth}
        \includegraphics[width=\linewidth]{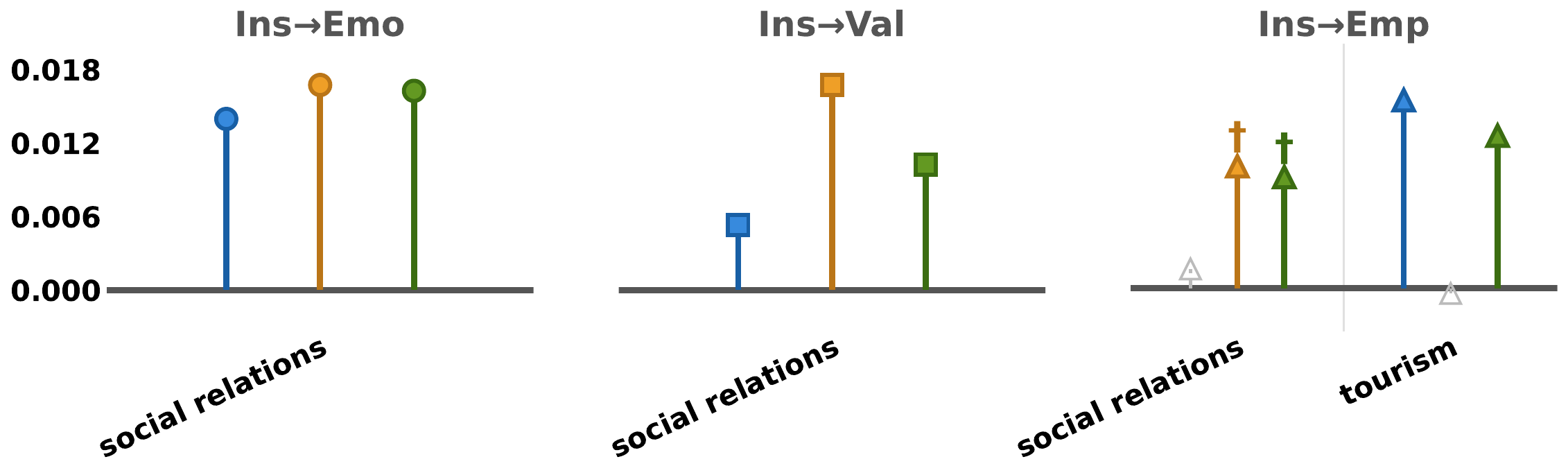}
        \caption{Ins $\rightarrow$ Anthro}
    \end{subfigure}
    \hfill
    \begin{subfigure}[t]{0.75\textwidth}
        \includegraphics[width=\linewidth]{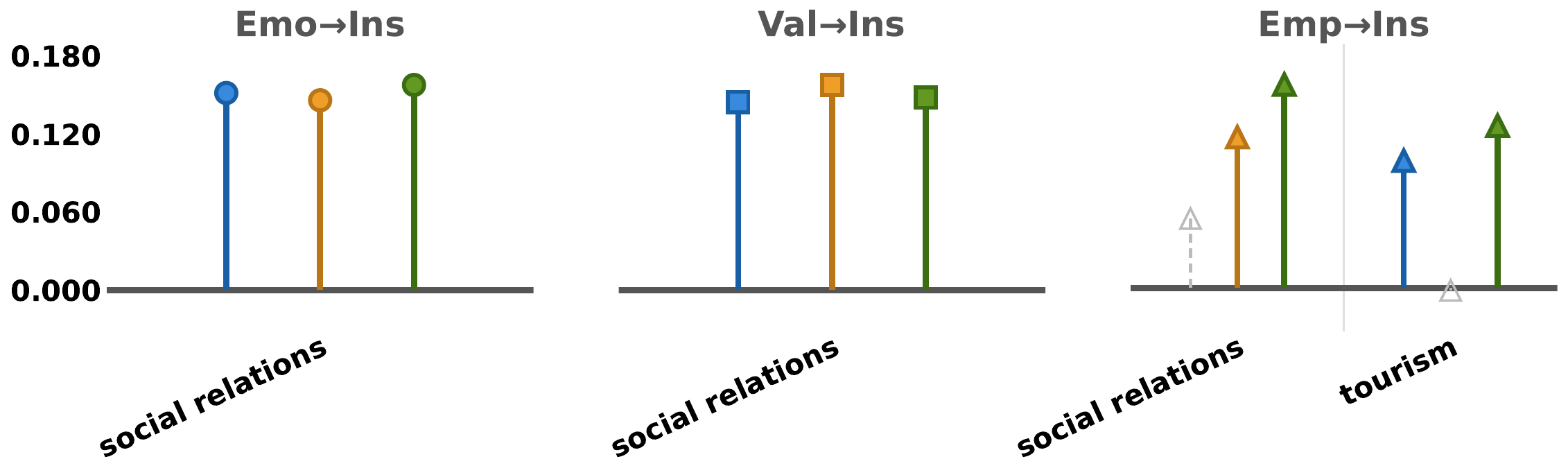}
        \caption{Anthro $\rightarrow$ Ins}
    \end{subfigure}
    \hfill
    \begin{subfigure}[t]{0.75\textwidth}
        \includegraphics[width=\linewidth]{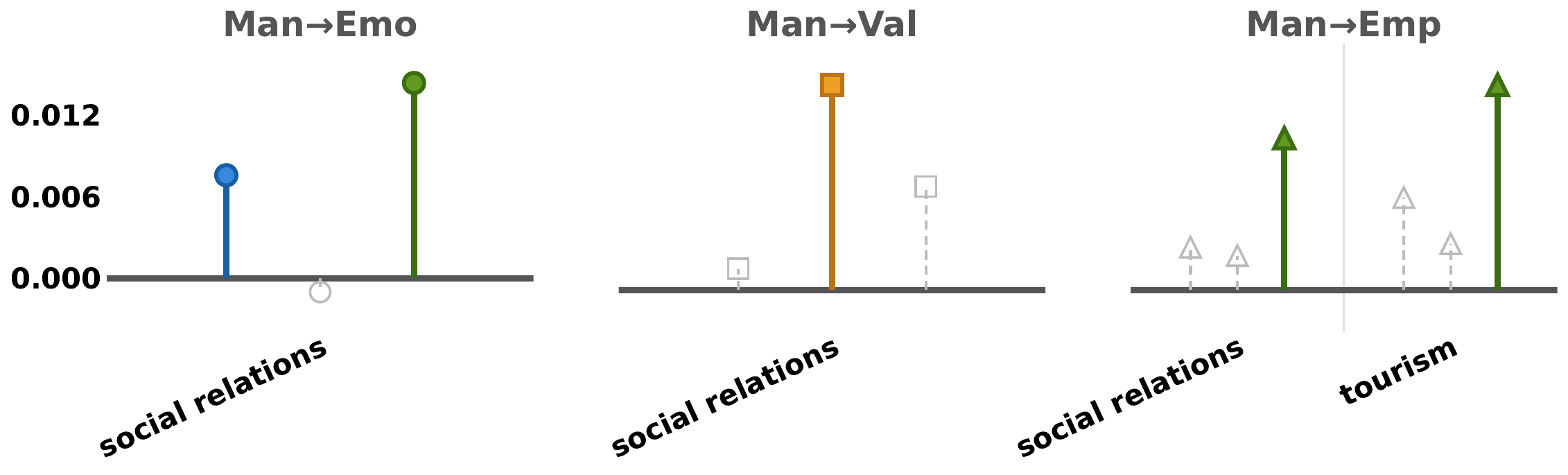}
        \caption{Manner $\rightarrow$ Anthro}
    \end{subfigure}
    \hfill
    \begin{subfigure}[t]{0.65\textwidth}
        \includegraphics[width=\linewidth]{figs/legend.pdf}
    \end{subfigure}
        \caption{\textbf{Turkey:} Computed via Eq.~\ref{eq:delta}, the estimated co-occurrence effects of: (a) cultural positioning on anthropomorphism, (b) anthropomorphism on cultural positioning, and (c) maxim of manner on anthropomorphism. Each panel shows the change in probability ($\Delta$) across models for the top-3 categories. Solid lines indicate significant Wald effects, $\dagger$ denotes cross-model differences FDR-corrected, both at p $<$ 0.05 (Sections~\ref{sec:rq2_result} and ~\ref{sec:rq3_result}).}
    \label{fig:tu_3block}
\end{figure*}

Figures \ref{fig:in_3block}--\ref{fig:tu_3block} capture the differences in estimated probabilities for the top-3 categories per country for how insider positioning affects anthropomorphism, how anthropomorphism affects positioning, and how the maxim of manner affects anthropomorphism.

\end{document}